\documentclass[a4paper,11pt,fleqn]{article}
\usepackage{fullpage}
\usepackage{authblk}
\usepackage[utf8]{inputenc}
\usepackage{amssymb}
\usepackage{amsmath}
\usepackage{graphicx}
\usepackage{hyperref}
\usepackage[subrefformat=parens]{subcaption}

\usepackage{tikz}
\usetikzlibrary{arrows.meta, positioning, calc}
\usetikzlibrary{decorations.pathreplacing}
\usepackage{pgfplots}
\pgfplotsset{compat=1.14}

\usepackage{natbib}

\usepackage{xcolor}
\newcommand{\R}{\ensuremath{\mathbb{R}}}
\newcommand{\dist}{\ensuremath{\text{dist}}}
\newcommand{\Err}[1][(x)]{\ensuremath{\mathcal{E}#1}}
\newcommand{\dose}{\ensuremath{\text{Dose}}}

\newcommand{\orcid}[1]{\href{https://orcid.org/#1}{\includegraphics[width=8pt]{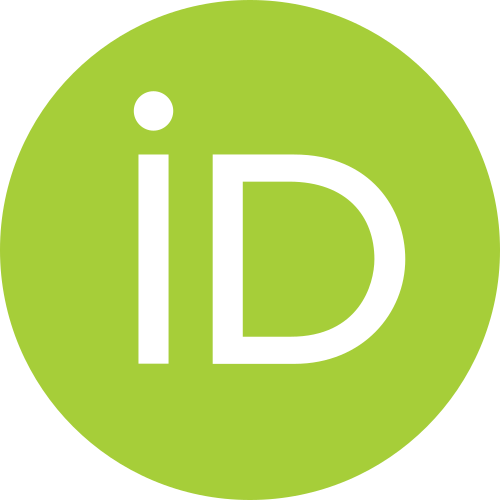}}}
\begin{document} 
\title{A practical probabilistic framework for deformable image registration uncertainty in radiotherapy dose propagation}

\author{
Stefan~Heldmann\textsuperscript{1,*}\orcid{0000-0002-9206-2086}, 
Sven~Kuckertz\textsuperscript{1}\orcid{0000-0002-4374-0880}, 
Nasim~Givehchi\textsuperscript{2}\orcid{0000-0002-2193-7535},
Thomas~Coradi\textsuperscript{2}\orcid{0009-0005-5181-5423},
Mikel~Byrne\textsuperscript{3}\orcid{0000-0003-2105-3964},  
Ben~Archibald-Heeren\textsuperscript{3}\orcid{0000-0001-9013-6048}, 
Nils~Papenberg\textsuperscript{1}\orcid{0009-0000-6385-1919}
}

\affil{
\textsuperscript{1}Fraunhofer Institute for Digital Medicine MEVIS, L\"ubeck, Germany,\\
\textsuperscript{2}Varian, a Siemens Healthineers company, D\"attwil, Switzerland,\\
\textsuperscript{3}Icon Group, South Brisbane, QLD, Australia,\\
\textsuperscript{*}Correspondence: stefan.heldmann@mevis.fraunhofer.de\\
}

\maketitle

\begin{abstract}
Deformable image registration (DIR) is widely used in radiotherapy for dose propagation and accumulation, but uncertainty in the underlying deformation can substantially affect clinically relevant dose estimates. We present a practical probabilistic framework for propagating DIR uncertainty to voxel-wise dose statistics and dose-volume histograms (DVHs). The method models the mapped correspondence at each voxel as a random variable governed by a transparent local certainty map that can be defined by simple safety margins, structure-boundary mismatch, or structure-wise conservative uncertainty values. This yields interpretable quantities such as dose probabilities, expected dose, confidence bounds, and induced DVH envelopes.

The framework is designed to remain lightweight and interpretable: it avoids complex biomechanical or ensemble-based uncertainty models and instead emphasizes simple parameterization, computational feasibility, and transparent dose metrics. We further introduce a structure-guided in/out strategy as an optional refinement that restricts mapping probabilities to anatomically plausible target regions. The approach is demonstrated on a prostate radiotherapy case study and used to compare different certainty-map strategies and probability kernels. The experiments show that the certainty-map design has a stronger effect on resulting dose and DVH uncertainty bounds than the specific kernel choice, while the additional benefit of the in/out strategy is case-dependent and modest in the present example. Overall, the proposed framework provides a transparent way to incorporate DIR uncertainty into radiotherapy dose assessment and to study how modelling choices affect propagated dose metrics.
\\[1em]
\end{abstract}

%% --------------------------------------------------------------------------------
%% --------------------------------------------------------------------------------
%% --------------------------------------------------------------------------------
\section{Introduction} 
\label{sec:introduction}
 Deformable image registration (DIR) is a cornerstone of modern radiation therapy. It enables voxel-by-voxel alignment of images across time and modalities to support adaptive workflows and synthetic CT (sCT) image generation. By capturing anatomical changes during treatment across fractions, DIR enables robust dose mapping and accumulation, improves the accuracy of estimated delivered dose, and provides information for re-optimization when clinically significant deviations occur \cite[]{MurrEtAl2023, ChettyRosuBubulac2019}.   
 3D deformable registration techniques provided the algorithmic basis for image-guided radiation therapy and demonstrated the feasibility of accurate, patient-specific deformation for complex anatomical motions \cite[]{FoskeyEtAl2005}. Current practice emphasizes validation, quantification of uncertainties, and quality assurance to ensure that the dose and contour propagation derived from DIR is reliable for clinical decision-making \cite[]{ChettyRosuBubulac2019, MurrEtAl2023}. Recent work has also addressed practical commissioning and end-user validation of DIR uncertainty tools in radiotherapy workflows \cite[]{BosmaEtAl2024, KipritidisEtAl2025}. 
 
Nevertheless, while DIR is a valuable, well-established tool, it is also a source of uncertainty and prone to error, as there is no dense ground truth for DIR and validation is an insufficiently solved (or perhaps unsolvable) challenge. 

Although there are numerous methods and proposed solutions that offer different potential outcomes for the same problem, the exact solution or deformation is unattainable and practically non-existent. At best, we can verify accuracy by evaluating the alignment of specific anatomical landmarks and regions. However, this leads to systematic or epistemic uncertainty that affects all subsequent processing steps based on registration result.

In this paper, we focus on handling uncertainty associated with dose propagation and accumulation, i.e., transferring fraction doses to a common baseline for therapy control, adaptive planning, and retrospective dose evaluation. Our interest is in quantifying how DIR-induced error alters clinically relevant dose metrics and in providing confidence bounds that are transparent and easy to interpret.

Various methods and approaches have been proposed to address this problem. \cite{NenoffEtAl2023} provide a comprehensive overview and assessment of current uncertainty due to DIR in radiation therapy, discuss the various causes and approaches to quantifying and detecting uncertainty, as well as the clinical implications of DIR uncertainty, and provide practical recommendations for patient-specific use. They conclude that common geometric metrics (e.g., target registration error (TRE), Dice coefficient (DSC), Hausdorff distance (HD), mean surface distance  (MSD)), DVF quality controls (e.g., Jacobian, inverse consistency), and dosimetric measurements (e.g., DVH bands, local uncertainty maps, distance-to-dose difference) can or should be combined to convert geometric uncertainty into voxel-wise dose uncertainty, and emphasize the need for standardized reporting and lightweight tools for routine workflows.
Closely related to the work in this article, two complementary strategies for dose accumulation that account for DIR uncertainties were recently presented. \cite{MechalakosEtAl2025} introduce RADAR, an Eclipse script that incorporates uncertainties via a voxel-wise ellipsoid search: For each imaged voxel, the maximum dose is accumulated within user-defined radii (global or structure-specific). An optional “structure matching” mode restricts the search to the same OAR to avoid shifting the target dose. \cite{MeyerEtAl2025} extend the concept by estimating the DIR uncertainty per voxel via a hyperparameter perturbation ensemble (63 DVFs) and principal component analysis, and constructing anisotropic voxel-specific confidence ellipsoids. During inverse dose mapping, the dose within each ellipsoid is queried to derive upper/lower bounds, resulting in DVH uncertainty bands. This framework was developed for automated adaptive offline radiation therapy and is integrated into an AWARE~\cite[]{AliottaHuZhang2023} pipeline that estimates uncertainties.

Two alternative approaches to uncertainty estimation and probabilistic dose calculation based on geometric deformations, which are very similar to the uncertainties caused by DIR, have recently been presented.

\cite{TillyEtAl2019} model deformations and setup variability using scenario sampling to quantify dose coverage as a probability (percentile dose, PD) and optimize plans that achieve target coverage at a chosen confidence level while reducing OAR (organ at risk) dose. They also introduced a fast fluence-based perturbation engine to generate dose coverage probability maps (DVCMs) for evaluation and planning under patterned geometric deformations.

\cite{SmoldersEtAl2023} propose a deep learning model (3D U-Net) to predict voxel-wise Gaussian uncertainty for a given deformation vector field (DVF), enabling probabilistic contour propagation and dose accumulation without changing the clinic's DIR algorithm. A 3D U-Net takes fixed/moving CTs and the mean DVF as input and outputs standard deviations per voxel; contour confidence maps and accumulated dose distributions are generated by sampling the resulting probabilistic DVF.

All of this motivated us to develop a new lightweight and transparent modelling approach suitable for routine clinical practice. Our overarching goal is to take a practical approach to dealing with uncertainties in order to improve patient safety and therapy outcome. A practical model means it can be calculated in a reasonable amount of time using standard infrastructure, and more importantly, it allows clinicians to develop an intuition and understanding of the calculated metrics and values. 

In brief, for a given registration, we assume a tolerance range for the calculated vector field at each point, which we then use to calculate probabilities and statistics for dose values. This allows us to make statements such as \emph{“Assuming that the registration error does not exceed 1 cm, the expected average dose is 0.7 Gy.”} or \emph{“With a probability of 99\%, the dose is below 0.9 Gy.”}.
For practical application, it is crucial to have a simple and understandable tolerance definition. The simplest strategy is to use a globally uniform tolerance. Secondly, we consider a straightforward strategy assigning smaller tolerances at locations where information about registration accuracy is available, yielding more precise estimates and sharper probability distributions. In addition, we consider a structure-guided "in/out strategy" as a post-processing refinement that suppresses anatomically implausible mappings when matched source and target structures are known.

 In contrast to other work, our  goal is not to provide a most accurate and realistic estimate of uncertainties, and we propose a simple model to account for DIR errors. The basis of our considerations in this context is that we are guided by the understanding of the terms “error” and “uncertainty” in the sense of "unknown possible variations of DIR" and uncertainty modeling in the sense of providing "safety margins". 

The paper is organized as follows: Section~\ref{sec:modelling_of_uncertainties_in_dose} introduces the uncertainty model and its mathematical foundations. Section~\ref{sec:experiments} describes the experimental setup and the results, and Section~\ref{sec:discussion} discusses the findings and clinical implications. Conclusions are drawn in Section~\ref{sec:conclusions}.

%
%
%
%
% --------------------------------------------------------------------------------
% --------------------------------------------------------------------------------
% --------------------------------------------------------------------------------
% --------------------------------------------------------------------------------
% --------------------------------------------------------------------------------
%
%
%
%
%

\section{Method / A probabalistic model form dose propagation  } 
\label{sec:modelling_of_uncertainties_in_dose}

In the following, we consider a model for dose propagation, i.e. the mapping or warping of a given dose distribution onto a reference coordinate system. This is essential, for example, for the summation of dose over fractions and adaptive radiation therapy. Here, the calculated dose distributions for each fraction are defined for the current patient anatomy and coordinates, which requires registration and warping to a uniform reference coordinate system. 

We do not specifically address dose summation or accumulation, and, for clarity, we describe only the mapping of a single dose at any point in time. To simplify the presentation, we therefore omit any reference to or indexing of fraction time or number, assuming that the model is independently applicable to each fraction.  

Thus, let $\dose:\R^3\to\R$ be the dose distribution at a given fraction. That is, $\dose(x)$ is the dose administered at 3D spatial position $x$ at the given fraction, as defined by the current patient coordinates. We then aim to propagate the dose distribution to a fixed baseline reference coordinate system, which is used for planning, accumulation or treatment monitoring, for example. Note that, the absolute position $x$ in the fraction generally has no anatomical correspondence to the same coordinate in the baseline for many reasons: patient position and motion, progressing disease and  therapy, breathing, filling of bladder and stomach etc. 
However, DIR is used to establish correspondence and a so-called deformation vector field (DVF), $y:\R^3\to\R^3$, is calculated to  map the baseline to the fraction, such that baseline coordinate $x$ corresponds to the warped coordinates $y(x)$ in the fraction. That is, $x$ and $y(x)$ localize the same anatomical point in the baseline and the fraction, respectively. This finally results in a propagated dose $\dose(y(x))$ defined on the fixed reference coordinates of the baseline.

%%
%%
%%
%% -----------------------------------------------------------------------------------------------
%%
%%
%%
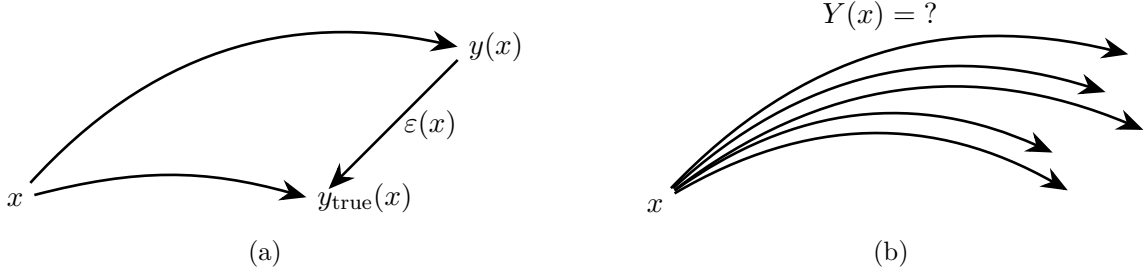
\begin{figure}[!t]
\centering
\begin{subfigure}{0.48\linewidth}
    \centering
    \begin{tikzpicture}[-{Stealth[length=10pt, width=8pt]}, line width=1pt]
    \node (x)     at (0, 0) {$x$};
    \node (y)     at (6, 2) {};
    \node (ytrue) at (4, 0) {};
    \draw[=>] (x) to[bend left] (y) node[right]{$y(x)$}; 
    \draw[=>] (x) to[bend left=15] (ytrue) node[right]{$y_{\text{true}}(x)$};
    \draw[=>] (y) -- (ytrue) node[midway,right]{$\varepsilon(x)$};
\end{tikzpicture}
    \caption{}
    \label{fig:deformation_as_RV:ytrue_y_err}
\end{subfigure}
\hfill
\begin{subfigure}{0.48\linewidth}
    \centering
    \begin{tikzpicture}[-{Stealth[length=10pt, width=8pt]}, line width=1pt]
        \node[anchor=east] (x) at (0, 0) {$x$};
        \draw[=>] (x) to[bend left] node[above=5pt]{$Y(x)=\text{?}$} (6, 2);
        \draw[=>] (x) to[bend left] (5.7, 1.5);
        \draw[=>] (x) to[bend left] (6.2, 1.0);
        \draw[=>] (x) to[bend left] (5.0, 0.7);
        \draw[=>] (x) to[bend left] (5.2, 0.2);
  \end{tikzpicture}
    \caption{}
    \label{fig:deformation_as_RV:Y}
\end{subfigure}
\caption{%
\label{fig:deformation_as_RV}%
Deformation as random variable modeling possible outcomes. 
\subref{fig:deformation_as_RV:ytrue_y_err} 
shows the relationship between the calculated deformation $y(x)$, the unknown true deformation $y_{\text{true}}(x)$, and the unknown error $\varepsilon(x)$. \subref{fig:deformation_as_RV:Y} illustrates the idea of modeling the deformation as a random variable $Y(x)$ that captures possible outcomes and leads to a distribution of propagated dose values per voxel.
}
\end{figure}
%%
%%
%%
%%
%% -----------------------------------------------------------------------------------------------
%%
%%
%%
As mentioned above, there is no fundamental truth for DIR, and a “true” deformation is not accessible. The best we can do is to calculate a reasonable DVF and make local uncertainty estimates for an unknown error. In the following, we will model this in terms of probabilities. To this end, we will consider the pointwise DVF $y(x)$ as a random variable $Y(x)$ that models the possible outcomes and leads to a distribution of the propagated dose values per voxel (cf. Figure~\ref{fig:deformation_as_RV}).

However, the starting point of our model is a map that controls the local (un)certainty and serves as input for the following probability model.

%% ##########################################################################################
%% ##########################################################################################
%% ##########################################################################################
%% 
%% SECTION CERTAINTY MAP
%% 
%% ##########################################################################################
%% ##########################################################################################
%% ##########################################################################################
%%
%%

\subsection{Certainty Map}
\label{certainty-map}
The key input of our uncertainty model is a certainty map $C$, i.e., a spatially varying parameter that controls the assumed registration uncertainty at location $x$. Formally, $C$ maps each spatial position to a local uncertainty descriptor $C(x)$. Depending on the model, $C(x)$ can be interpreted in different ways: as an isotropic radius $r$, as a scalar standard deviation $\sigma$, or as an anisotropic covariance matrix $\Sigma\in\R^{3\times 3}$. In this work, we primarily use the radius interpretation because it is intuitive and easy to communicate in clinical terms (``safety margin around the mapped point'').

\begin{figure}[!htbp]
\centering
\begin{subfigure}{0.48\linewidth}
    \centering
    \begin{tikzpicture}[scale=1.2]
        \coordinate (x0) at (0,-1);
        \coordinate (y0) at (3,0);
        % Ellipse around y0 (tilted)
        \begin{scope}[shift={(y0)}]
            \draw[fill=green!30, fill opacity=0.18, draw=green!30!black, very thick] (0,0) circle (1.4cm);
            \draw[green!30!black, thick] (0,0) -- (1,1);
            \node[green!30!black, anchor=east] at (0.8,0.8) {$C(x)=r$};
        \end{scope}
        \node[green!30!black, anchor=north,text width=6cm,align=center] at (3,-1.5) {Isotropic safety region \linebreak $\{z:\|z-y(x)\| \leq C(x) \}$};
        % Point y and label
        \fill (x0) circle (2pt);
        \node[anchor=east] at (x0) {$x$};

        \fill (y0) circle (2pt);
        \node[anchor=west, right=5pt] at (y0) {$y(x)$};
        
        % Arrow from x0 to y0
        \draw[-{Stealth[length=10pt, width=8pt]}, line width=1pt,shorten >=3pt] (x0) to[bend left] (y0);
    \end{tikzpicture}
    \caption{}
    \label{fig:support_pY:iso}
\end{subfigure}
\hfill
\begin{subfigure}{0.48\linewidth}
    \centering
    \begin{tikzpicture}[scale=1.2]
        % Key points
        \coordinate (x0) at (0,-1);
        \coordinate (y0) at (3,0); 
        % Ellipse around y0 (tilted)
        \begin{scope}[shift={(y0)}, rotate=+25]
            \draw[fill=green!30, fill opacity=0.18, draw=green!30!black, very thick] (0,0) ellipse (2cm and 1cm);
            \draw[dashed,green!30!black] (-2,0) -- (2,0);  
            \draw[dashed,green!30!black] (0,-1) -- (0,1);  
            \node[green!30!black, anchor=east] at (1.5,+0.1) {$C(x)=\Sigma$}; 
        \end{scope}
        \node[green!30!black,anchor=north,text width=6cm,align=center] at (3,-1.5) {Anisotropic safety region $\{z:\|\Sigma^{-\frac12}(z-y(x))\| \leq 1\}$};
        
        % Point y and label
        \fill (x0) circle (2pt);
        \node[anchor=east] at (x0) {$x$};
        \fill (y0) circle (2pt);
        \node[anchor=west, right=5pt] at (y0) {$y(x)$};
        % Arrow from x0 to y0
        \draw[=>,-{Stealth[length=10pt, width=8pt]}, line width=1pt,shorten >=3pt] (x0) to[bend left] (y0);
    \end{tikzpicture}
    \caption{}
    \label{fig:support_pY:aniso}
\end{subfigure}
\caption{%
\label{fig:support_pY}% 
Simple model for a safty margin in the sense of a radius. We consider an example scenario in which we have performed a registration so that the point $x$ is mapped to $y(x)$, but we are not certain and factor in a margin of safety. \subref{fig:support_pY:iso} shows the case of an isotropic radius, i.e., $C(x)=r$ is a positive scalar radius. \subref{fig:support_pY:aniso} shows an example of an anisotropic case in which $C(x)=\Sigma$ is a (symmetric positive definite) matrix that parameterizes an ellipsoidal uncertainty region. 
}
\end{figure}
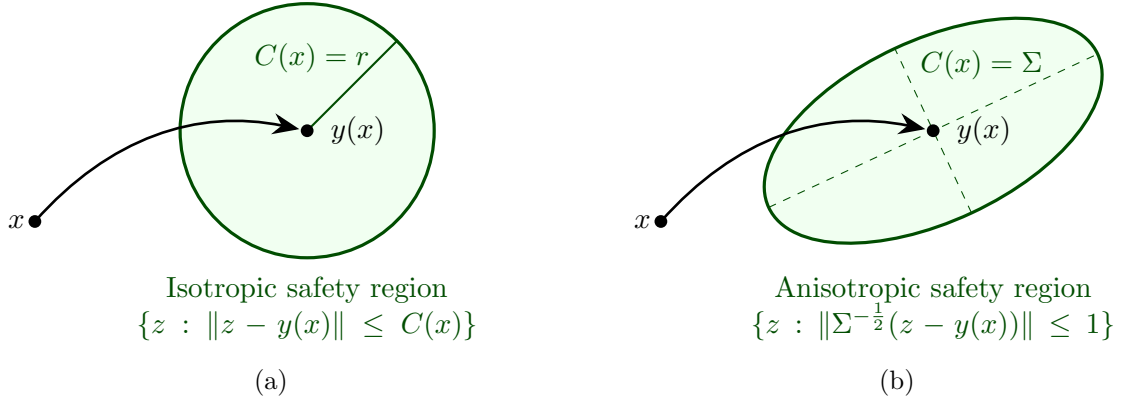

For example, $C(x)=r$ means that the unknown true correspondence of $y(x)$ is assumed to lie within a radius of $r$ around $y(x)$ (cf. Figure~\ref{fig:support_pY:iso}). Analogously, $C(x)=\Sigma$ defines an ellipsoidal region of plausible correspondences (Figure~\ref{fig:support_pY:aniso}). Related concepts can be found in \cite{MechalakosEtAl2025} (user-defined spherical uncertainty regions) and \cite{MeyerEtAl2025} (voxel-wise anisotropic ellipsoids from registration ensembles).

In practice, uncertainty cannot be measured directly, and only sparse evidence for alignment quality is usually available (e.g. of landmarks, contours, structure surfaces). Therefore, we use simple and transparent construction rules for $C$. The goal is not maximal model complexity, but robust and understandable safety margins for routine use.
%%
%%
%%
%% -----------------------------------------------------------------------------------------------
%%
%%
%%
%%
%%
%%
%%---------------------------------------------------------------------------------------
%%
%%
%%
%%

\subsubsection*{Strategy 1: Global constant certainty}
The simplest assumption is a globally constant certainty map,
\begin{equation}
C(x)=C_0 \quad \text{for all } x,
\end{equation}
with user-defined parameter $C_0>0$. This yields a uniform safety margin everywhere and serves as a baseline.

\subsubsection*{Strategy 2: Boundary-based spatially varying certainty}
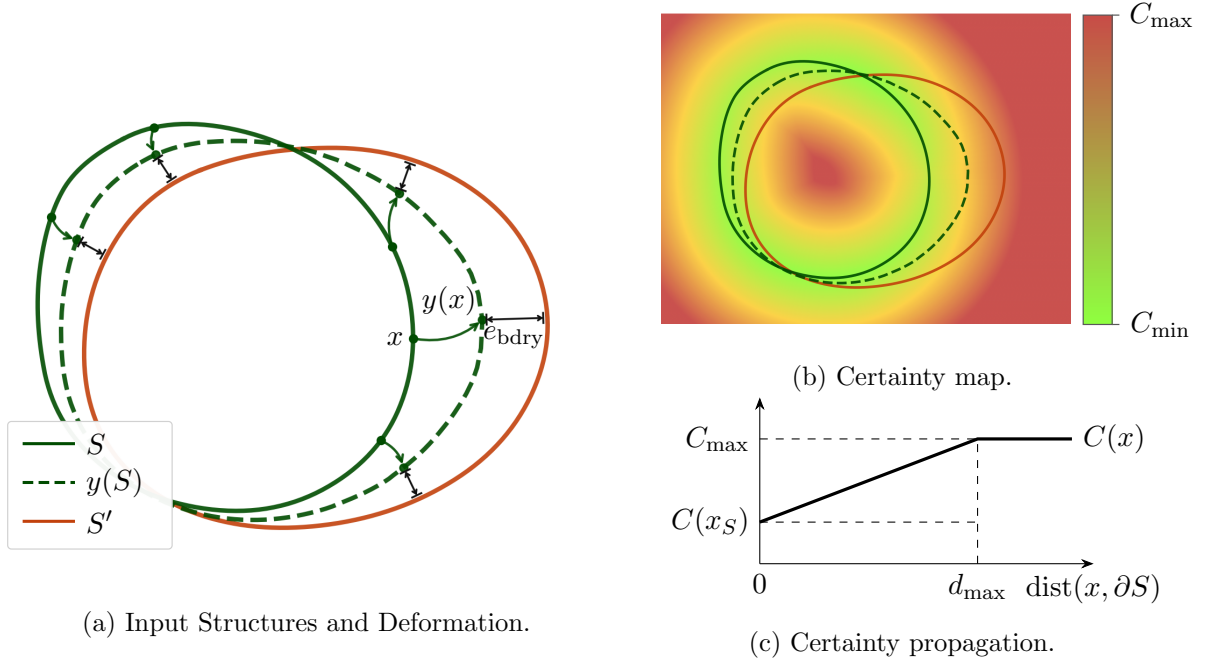
\begin{figure}[!htbp]
    \centering
    \hspace{-20mm}
    \begin{minipage}[c]{0.62\linewidth}
        \centering
        \begin{subfigure}{\linewidth}
            \centering
            \begin{tikzpicture}
                \node[anchor=south west, inner sep=0] (img) at (0,0) {\includegraphics[width=\linewidth]{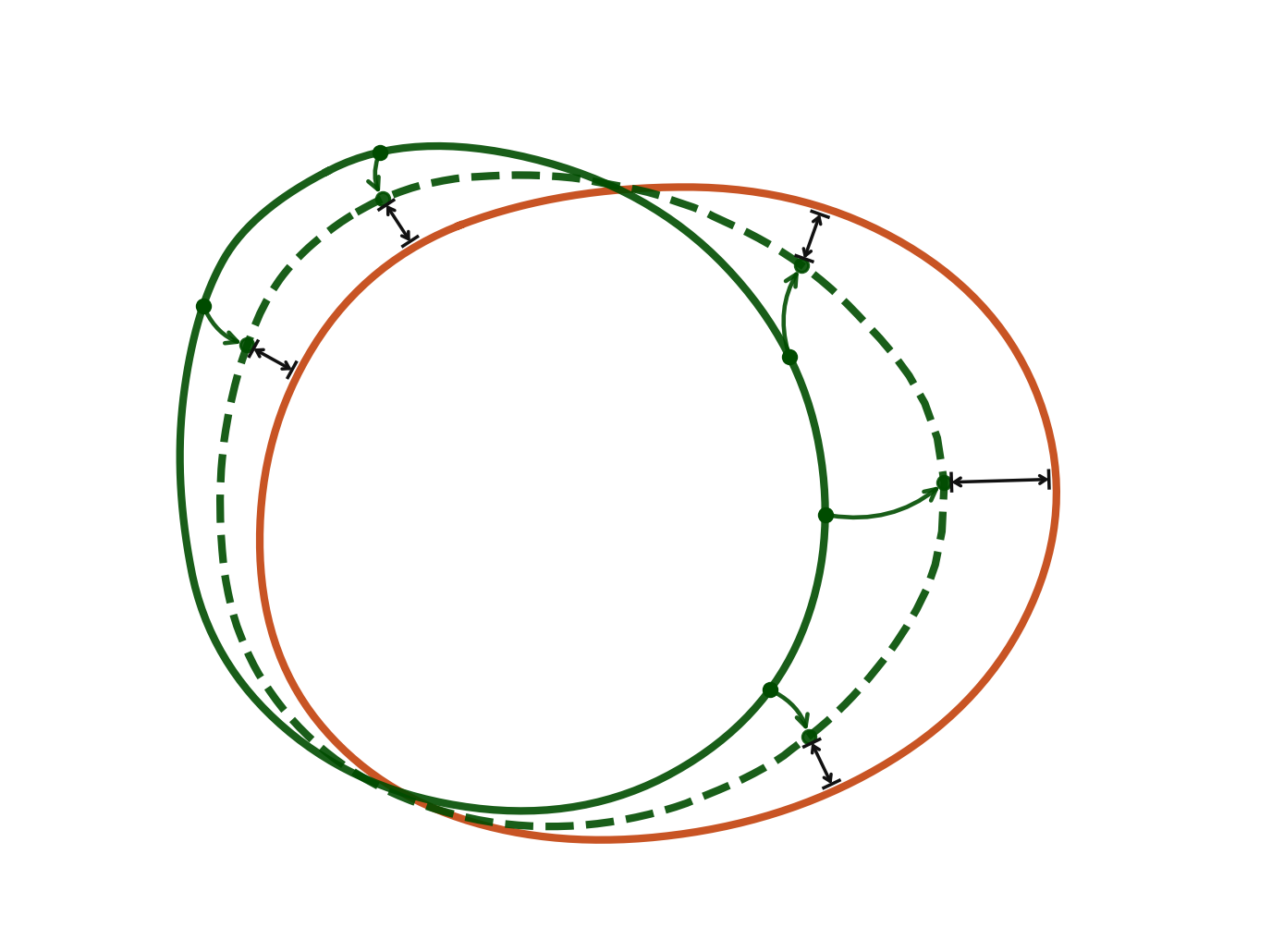}};
                \begin{scope}[x={(img.south east)}, y={(img.north west)}]
                % draw legend box and entries              
                \begin{scope}[shift={({0.1},{0.15})}]
                    \definecolor{FixedGreen}{HTML}{004C00}
                    \definecolor{TargetOrange}{HTML}{C2410C}
                    \tikzset{
                        box/.style={black!18, line width=0.45pt, rounded corners=1.4pt, fill=white, fill opacity=0.95},
                        fixed/.style={draw=FixedGreen, line width=1.15pt},
                        deformed/.style={draw=FixedGreen, line width=1.15pt, dash pattern=on 4pt off 2pt},
                        target/.style={draw=TargetOrange, line width=1.15pt},
                    }               
                    \coordinate (a0) at (0.02,0.12); 
                    \coordinate (a1) at ($(a0)+(0.07,0.0)$); 
                    \coordinate (b0) at ($(a0)-(0.0,0.07)$); 
                    \coordinate (b1) at ($(a1)-(0.0,0.07)$); 
                    \coordinate (c0) at ($(b0)-(0.0,0.07)$); 
                    \coordinate (c1) at ($(b1)-(0.0,0.07)$); 
                    \draw[box     ] (0,-0.07) rectangle (0.22, 0.16);
                    \draw[fixed   ] (a0) -- (a1) node[anchor=west] {$S$}; 
                    \draw[deformed] (b0) -- (b1) node[anchor=west]  {$y(S)$};
                    \draw[target  ] (c0) -- (c1) node[anchor=west] {$S'$};
                \end{scope}

                \coordinate (x) at (0.643,0.46);
                \coordinate (y) at (0.735,0.5);
                \coordinate (e) at (0.82,0.5);
                \node [anchor=east] at (x) {$x$};
                \node [anchor=south east, inner sep=0pt, xshift=-2pt] at (y) {$y(x)$}; 
                \node [anchor=north] at ($(y)!0.5!(e)$) {$e_{\mathrm{bdry}}$}; 
                \end{scope}
            \end{tikzpicture}
            \caption{Input Structures and Deformation.}
            \label{fig:boundary-based-cmap:input}
        \end{subfigure}
    \end{minipage}
    \hspace{-5mm}
    \begin{minipage}[c]{0.40\linewidth}
        \centering
        \begin{subfigure}{\linewidth}
            \centering
            \begin{tikzpicture}
                \node[anchor=south west, inner sep=0] (img) at (0,0) {\includegraphics[width=\linewidth]{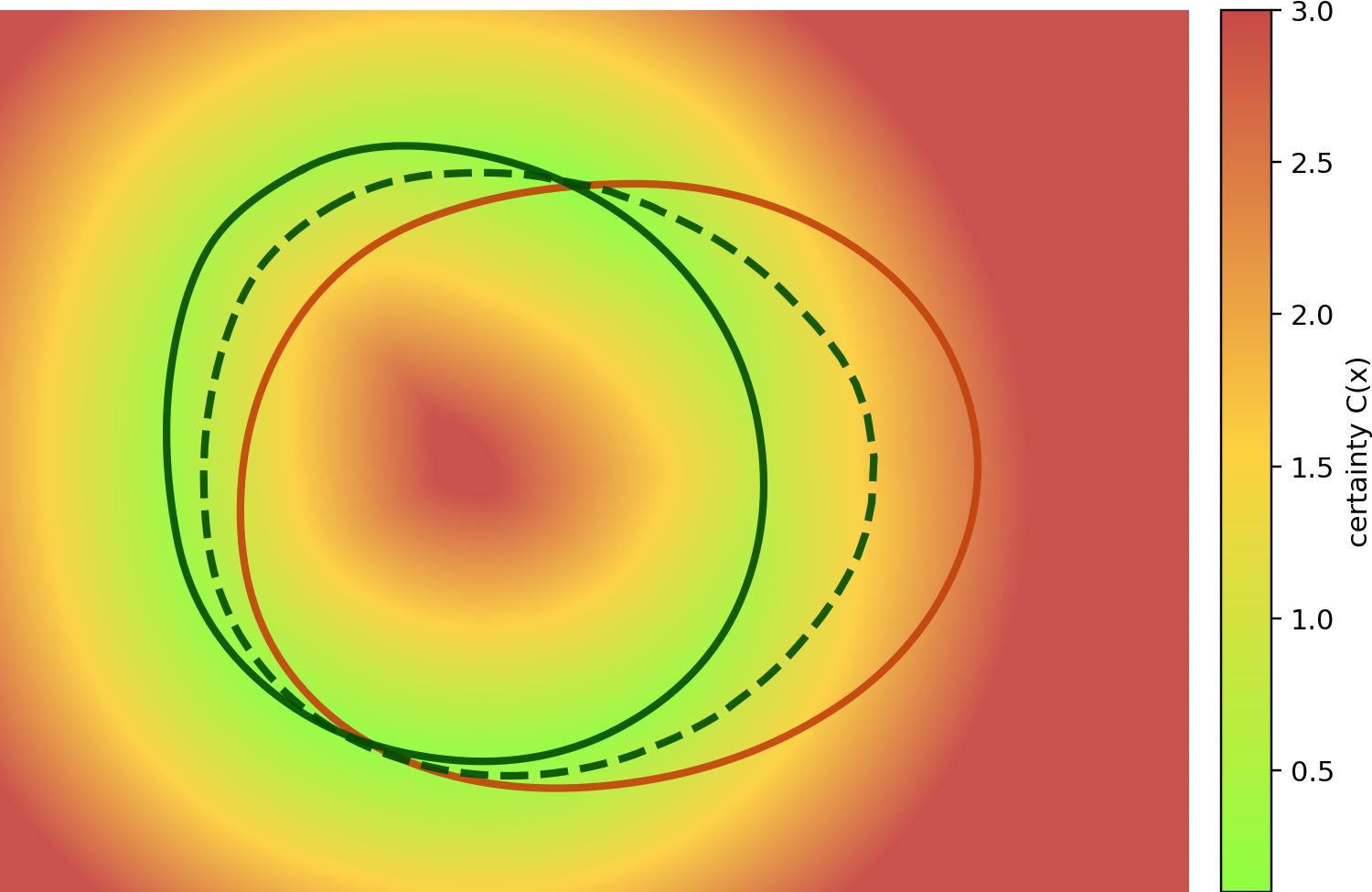}};
                \begin{scope}[x={(img.south east)}, y={(img.north west)}]
                    \definecolor{LowCertainty}{RGB}{143,255,64}
                    \definecolor{MidCertainty}{RGB}{253,209,64}
                    \definecolor{HighCertainty}{RGB}{200,74,71}
                    \fill[white, opacity=1] (0.85,0.00) rectangle (1,1);
                    \shade[bottom color=LowCertainty, middle color=MidCertainty, top color=HighCertainty] (0.875,0.0) rectangle (0.935,0.985);
                    \draw[black!65, line width=0.45pt] (0.875,0.0) rectangle (0.935,0.985);
                    \draw[black!65, line width=0.45pt] (0.935,0.0) -- (0.95,0.00);
                    \draw[black!65, line width=0.45pt] (0.935,0.985) -- (0.95,0.985);
                    \node[anchor=west] at (0.952,0.00) {$C_{\min}$};
                    \node[anchor=west] at (0.950,0.985) {$C_{\max}$};
                \end{scope}
            \end{tikzpicture}
            \caption{Certainty map.}
            \label{fig:boundary-based-cmap:certainty-map}
        \end{subfigure}
        \vspace{0.9em}
        \begin{subfigure}{\linewidth} 
            \centering
            \begin{tikzpicture}[x=1.2cm,y=1.1cm,>=Stealth,xscale=0.8]
                \def\CxS{1.4}
                \def\dmax{3.0}
                \def\CxS{0.5} 
                \def\Cmax{1.5}
                \draw[->] (0,0) -- (4.6,0) node[below] {$\dist(x,\partial S)$};
                \draw[->] (0,0) -- (0,2);
                \draw[dashed] (0,\CxS) -- (\dmax,\CxS);
                \draw[dashed] (\dmax,0) -- (\dmax,\Cmax);
                \draw[dashed] (0,\Cmax) -- (\dmax,\Cmax);
                \draw[very thick] (0,\CxS) -- (\dmax,\Cmax) -- (4.3,\Cmax) node[anchor=west] {$C(x)$};
                \node[left] at (0,\CxS) {$C(x_S)$};
                \node[left] at (0,\Cmax) {$C_{\max}$};
                \node[below] at (0,0) {$0$};
                \node[below] at (\dmax,0) {$d_{\max}$};
            \end{tikzpicture}
            \caption{Certainty propagation.}
            \label{fig:boundary-based-cmap:certainty-propagation}
        \end{subfigure}
    \end{minipage}
    \caption{%
        Illustration of the boundary-based certainty modeling strategy. 
    \subref{fig:boundary-based-cmap:input} Reference structure $S$, its deformation $y(S)$, and the corresponding structure $S'$ annotated on the target image,  with the points $x$ in the reference space mapped to $y(x)$ in the target space as well as the boundary mapping error $e_{\mathrm{bdry}}$ indicated (in fact, the image only shows the structure boundary, but $S$, $y(S)$, and $S'$ are considered as the whole filled area surrounded, not only the boundaries as plotted). 
    \subref{fig:boundary-based-cmap:certainty-map} Certainty map $C$ defined on the reference domain based on the distance of points to the structure boundary as depicted in~\subref{fig:boundary-based-cmap:certainty-propagation}. The color bars indicate the range from minimum to maximum certainty $C_{\min}$ to $C_{\max}$. 
    \subref{fig:boundary-based-cmap:certainty-propagation} Propagation of certainty value $C(x_S)$ from a boundary point $x_S\in\partial S$ in the reference domain to closest other points $x$. Starting from $C(x_S)$ at the boundary, certainty increases linearly with $\dist(x,\partial S)$ and saturates at $C_{\max}$ for distance $d_{\max}$.
    \label{fig:boundary-based-cmap}
    }
\end{figure}

If corresponding segmentations are available, local boundary mismatch can be used as an uncertainty proxy. Let $S\subset\R^3$ be a structure in reference space and $S'$ the corresponding structure in fraction space. For boundary points $x\in\partial S$, we measure the misalignement of the mapped point $y(x)$ to the surface $\partial S'$ as a local error proxy, i.e., the distance from $y(x)$ to the closest point on $\partial S'$:
\begin{equation}
\mathop{e_\text{bdry}}(x):=\dist(y(x),\partial S')=\min_{x'\in\partial S'}\|y(x)-x'\|.
\end{equation}
We then use this boundary error to enforce a local certainty that is at least as large as the observed boundary mismatch, i.e., for boundary points $x\in\partial S$ we assign a minimum safety margin $C_{\min}>0$. For non-boundary points $x\in S\setminus\partial S$, we propagate certainty from the closest boundary point and increase it linearly with distance to the boundary, capped at distance $d_{\max}$ by $C_{\max}$: 
\begin{equation}
C(x)= \left\{\begin{array}{ll}
    \max\big(e_\text{bdry}(x),C_{\min}\big), & x\in\partial S, \\\\
    \min\left(C(x_S)+\frac{\dist(x,\partial S)}{d_{\max}}\big(C_{\max}-C(x_S)\big),\,C_{\max}\right), & x\in S\setminus\partial S
\end{array}\right.
\end{equation}
where $x_S\in\partial S$ is a closest boundary point to $x$. Figure~\ref{fig:boundary-based-cmap} provides a schematic illustration of the boundary-based strategy and its effect on the certainty map.

%%
%%
%%
%% ---------------------------------------------------------------------------------
%% ---------------------------------------------------------------------------------
%%
%%
%%   

%%
%%
%% -----------------------------------------------------------------------------------------
%% -----------------------------------------------------------------------------------------
%%
%%

\paragraph{Practical design rationale.}
For practical clinical use, the certainty-map design must remain transparent and controllable. In our context, interpretability is more important than maximal model complexity: clinicians should be able to understand where local safety margins come from, how they react to measured mismatch, and how parameter changes affect the resulting dose bounds. Therefore, we intentionally use simple construction rules based on measurable geometric quantities (surface mismatch and distance from the structure boundaries) and only a small number of parameters ($C_{\min}$, $C_{\max}$, $d_{\max}$, and optional global/background constants).

\begin{center}
\fbox{\begin{minipage}{0.96\linewidth}
\textbf{Practical computation of the certainty map $C$}\par
\begin{enumerate}
    \item Select one of the construction strategies according to available information: global constant, boundary-based spatially varying.
    \item If segmentations are available, compute boundary mismatch values $\dist(y(x),\partial S')$ on structure boundaries and derive certainty values using the selected rule.
    \item Clamp values to predefined lower/upper limits to avoid unrealistic local certainty values.
    \item Assign a background certainty outside delineated structures (if required).
    \item Compute 3D scalar distribution of $C$ across the image from 2-4.
    \item Use the resulting map $C$ as direct input for the probabilistic deformation model in the following sections.
\end{enumerate}
\end{minipage}}
\end{center}

\subsection{Deformation as a Random Variable}
\label{deformation-as-a-random-variable}
We are pursuing the idea of a probabilistic model for errors and variations of a given DIR, which can then be used to determine statistics, limits, and safety margins. 
To this end, we focus on a given DIR. Let us take a small step back and let $y$ be the calculated DVF, $y_\text{true}$ the unknown true deformation, and $\varepsilon$ the true and also unknown error, such that
\begin{equation}\label{eq:y_true=y+eps}
	y_{\text{true}}(x)=y(x)+\varepsilon(x) 
\end{equation}
Usually, there is little information that can be used to estimate the error. This could be, for example, a few corresponding (anatomical) landmarks or segmented structures/regions. However, the error cannot be measured, and we model $\varepsilon(x)$ as a real-valued random variable, i.e., we assign probabilities to the possible outcomes of the error $\varepsilon(x)$ probabilities.

%% --------------------------------------------------------------------------------
%% MODEL Assumptions
%% --------------------------------------------------------------------------------
%%
For our modeling, we build on following assumptions: 
\begin{enumerate}
    \item Errors are a priori unknown.
    \item Errors are local and depend on the location.
    \item Errors at different locations are independent from each other.
\end{enumerate}
While the first two assumptions might be seen obvious, the third one needs some discussion. 
In general, we expect deformations to be smooth and closely located points to be mapped to new closely located points. This suggests that errors at neighboring locations are also similar, i.e., if we know the error at one location, we can provide a good estimate for the error at the neighboring location, which means that they are (stochastically) dependent on each other.

Contrary to these expectations, we deliberately assume in our model that errors are independent of each other, which means that we make no explicit assumptions about the smoothness of the deformation and actively include the possibility of unknown true deformation with extreme and discontinuous motion. 

Otherwise, assuming coupling requires modeling the smoothness properties of deformations, which is challenging in itself. Common methods for measuring deformation smoothness are either completely artificial, such as considering derivative norms, or mechanically motivated, such as elasticity theory whereby practicability and computational feasibility are crucial aspects. A comprehensive overview of such technologies for "classical" variational approaches can be found e.g., in \cite{SotirasDavatzikosParagios2013} or \cite{XiaEtAl2023} for current deep learning-based approaches. Other approaches focus on complex, detailed biomechanical modeling and simulations. These are usually computationally intensive and difficult to apply on a broad scale. However, research in this area is ongoing, and physically inspired neural networks are one example of current advances; see \cite{BanerjeeEtAll2024} for an overview. 

Stochastic approaches that attempt to generate and integrate such estimates of entire deformations are typically based on a Bayesian registration approach using Gaussian distributions \cite[]{WachingerEtAl2014,WassermannEtAl2014,WangWellsGolland2018, WangWellsGolland2019} and/or Monte Carlo simulations \cite[]{FolgocEtAl2017,RisholmEtAl2013,PursleyEtAl2012} for modeling variations and distributions of entire deformations, respectively. Although this involves spatial coupling through the design, which could be considered advantageous, such models are also not easy to handle—especially since there is no fundamental truth and parameters of the underlying stochastic model, such as standard deviations of Gaussian distributions, are difficult to estimate and are often chosen a priori by hand. 
In any case, developing a practical method for determining parameters that describe coupling is challenging and requires complex parameters whose selection is neither clear nor easily understood or done by non-experts.  

For this reason, we decided against this approach, as the actual benefit seems low given the complexity and comprehensibility.

For clarity, in the following we always denote random variables with capital letters. 
Let $\Err: \Omega\to\R^3$, $\omega\mapsto\Err(\omega)$ be the random variable defined on a sample space $\Omega$ that models the possible outcomes of the error $\varepsilon(x)$ at point $x$. Furthermore, let $\Pr$ be the probability measure on the sample space and let $P_{\mathcal{E}(x)}$ the distribution of $\Err$, such that 
\begin{equation}\label{eq: Error distribution}
 P_{\mathcal{E}(x)}(A):=\Pr[\mathcal{E}(x)\in A] \in [0,1]
\end{equation}
is the probability that the error $\varepsilon(x)$ or $\mathcal{E}(x)$ takes values in some set $A\subset\R^3$. As usual, we use the abbreviated notation for sets $\{\Err\in A\} = \{\omega\in\Omega\,:\,\mathcal{E}(x)(\omega)\in A \}$ and $\Pr[\Err\in A]=\Pr[\{\Err\in A \}]$.
Furthermore, we represent $P_{\mathcal{E}(x)}$ by a density. That is, we consider non-negative integrable functions $p_{\mathcal{E}(x)}:\R^3\to [0,\infty)$ with
\begin{equation}
	P_{\mathcal{E}(x)}(A)=\int_A p_{\mathcal{E}(x)}(z) \, dz.
\end{equation}
Although the following concept is general for arbitrary density models, in our setting we particularly think of the error as local and bounded by some radius for which we have an estimate or rationale given by the certainty map $C(x)$. That is, we set $r \cong C(x)$ so that $p_{\mathcal{E}(x)}=0$ for $\|z\|>r$ and $\Pr[\|\Err\|\leq r]=1$. 
A common approch, is use of kernel densities, where $p_{\Err}$ is constructed as a scaled and truncated version of a one-dimensional kernel $K:\R\to[0,\infty)$, i.e., $p_{\Err}(z)=\frac{1}{r} K(\frac{\|z\|}{r})$ or in case certainty is modelled by an anisotropic covariance matrix $\Sigma$, $p_{\Err}(z):=\frac{1}{|\det\Sigma|} K(|z^\top\Sigma^{-1}\ z|)$. Then, provided the support of $K$ is $[-1,1]$, we have the desired property support of $p_{\Err}$ is $\{z:\|z\|\leq r\}$
 and $\{z: |z^\top\Sigma^{-1}z|\leq 1\}$, respectively, for the anisotropic case. Common choice are B-spline kernels or (truncated) Gaussian kernels for modelling an error whose probability decays away from zero. Examples of such kernels are shown in Figure~\ref{fig:Probability-kernels} (uniform, linear/quadratic/cubic B-splines, and truncated Gaussians).
%%
%%
%%
%%
%% -----------------------------------------------------------------------------------------------
%%
%%
%%
\begin{figure}
    \centering
        \begin{tikzpicture}
            \begin{axis}[
                scale=0.9,
                width=10cm,height=5.5cm,
                xmin=-1.5, xmax=1.5,
                ymin=0,   ymax=1.65,
                axis lines=left,
                axis line style={very thick},
                tick align=outside,
                samples=500,
                legend style={legend cell align=left, draw=none, fill=none, font=\small, at={(1.9,1)}, anchor=north east},
                clip=false
                ]
                % Uniform (Box) kernel: K(x)=1/2 for |x|<=1
                \addplot[blue, very thick, domain=-1.5:1.5] {(abs(x)<=1)*0.5};
                \addlegendentry{0$^\text{th}$ order B-spline $K_\text{uniform}$}
                
                % Linear B-spline (Triangular): max(0, 1-|x|), support [-1,1])
                \addplot[purple, very thick, domain=-1.5:1.5] {max(0, 1-abs(x))};
                \addlegendentry{1$^\text{st}$ order B-spline $K_\text{linear}$}
                
                % Quadratic B-spline, support [-1.5,1.5] scaled to support [-1,1] use x -> 1.5*x
                % M2(x) = 3/4 - x^2          for |x|<=0.5
                %         1/2 (|x|-1.5)^2    for 0.5<|x|<=1.5
                \addplot[green!70!black, very thick, domain=-1.5:1.5]{
                  (abs(x)<=1/3)                 * 9/8 * (1 - 3*x^2) +
                  (abs(x)>1/3 && abs(x)<=1)     * 27/16* ((abs(x)-1)^2)
                };
                \addlegendentry{2$^\text{nd}$ order B-spline $K_\text{quad}$}
                
                % Cubic B-spline, support [-1,1] and still integrate to 1: use x -> 2*x
                % M3(x) = (4 - 6x^2 + 3|x|^3)/6     for |x|<=1
                %         ((2-|x|)^3)/6             for 1<|x|<=2
                \addplot[orange!90!black, very thick, domain=-1.5:1.5]{
                  (abs(x)<=1/2)                       * 8 * (1/6 - x^2 + abs(x)^3) + 
                  (abs(x)>1/2 && abs(x)<=1)           *     (8/3*(1-abs(x))^3)
                };
                \addlegendentry{3$^\text{rd}$ order B-spline $K_\text{cubic}$}
                
                % Gaussian sigma = 1/3
                \addplot[red, very thick, domain=-1.5:1.5] {3/(sqrt(2*pi)) * exp(-9*x^2/2)};
                % \addlegendentry{Gauss sigma 1/3}
                \addlegendentry{(truncated) Gaussian $K_\text{Gauss}$ ($\sigma=\frac13$)}
                
                % Gaussian sigma = 1/4
                \addplot[cyan!60!blue, very thick, domain=-1.5:1.5] {1/(sqrt(2*pi)*0.25) * exp(-x^2/(2*0.25^2))};
                \addlegendentry{(truncated) Gaussian $K_\text{Gauss}$ ($\sigma=\frac14$)}
            \end{axis}
            \begin{scope}[anchor=west,shift={(-1,-0.3)},scale=1.2]
                \node at (0,-0.8) {$K_\text{uniform}(t)=\frac{1}{2} \mathbf{1}_{[-1,1]}(t)$,};
                \node at (6.5,-0.8) {$K_\text{linear}(t)=\max(0,1-|t|)$,};
                \node at (0,-2) {
                    $K_\text{quad}(t)=\begin{cases}
                        \frac{9}{8}(1-3t^2),    &\text{$|t|\leq \frac{1}{3}$}\\ 
                        \frac{27}{16}(|t|-1)^2, &\text{$\frac{1}{3}<|t|<1$}  \\ 0& \text{else}\end{cases}$,
                };                    
                \node at (6.5,-2) {
                    $K_\text{cubic}(t)=\begin{cases} 
                    8(\frac{1}{6} - t^2 + |t|^3),   & \text{$|t|\leq \frac{1}{2}$}\\ 
                    \frac{8}{3}(1-|t|)^3,           & \text{$\frac{1}{2}<|t|<1$}  \\ 0& \text{else} \end{cases}$,
                };               
                \node at (0,-3.2) {
                    $K_\text{Gauss}(t) = \frac{K_\sigma(t)\mathbf{1}_{[-1,1]}(t)}{\int_{[-1,1]} K_\sigma(\theta)\,d\theta}$ 
                    with  $K_\sigma(t)=\frac{1}{\sqrt{2\pi\sigma^2}} e^\frac{-t^2}{2\sigma^2}$.
                };    
            \end{scope}
        \end{tikzpicture}
    % }
    \caption{
    Common Examples for 1D probability kernel functions $K$ with support $[-1,1]$.  
    }
    \label{fig:Probability-kernels}
\end{figure}
%%
%%
%%
%%
%% -----------------------------------------------------------------------------------------------
%%
%%
%%
%%

Inspired by \eqref{eq:y_true=y+eps}, then $y(x)+\mathcal{E}(x)$ is a probabilistic model for the unknown but true deformation $y_\text{true}(x)$ at a point $x$ therefore we define the random variable 
\begin{equation}
	Y(x):=y(x)+\mathcal{E}(x) 
\end{equation}
modeling the possible outcomes of the deformation $y(x)$ of the point $x$ (cf. Figure~\ref{fig:deformation_as_RV}) and let $P_{Y(x)}$ be its distribution such that
\begin{eqnarray}
    P_{Y(x)}(A) &=& \Pr[Y(x)\in A] 
\end{eqnarray} 
the probability that a point $x$ is mapped by the deformation $y_\text{true}$ into some region $A\subset \R^3$. Since the event $Y(x)\in A$ is equivalent to $\Err \in A-y(x)$, we have 
$P_{Y(x)}(A)=P_{\Err}(A-y(x))=\int_{A-y(x)} p_{\Err}(z)$ such that
\begin{eqnarray}
	P_{Y(x)}(A)&=& \int_{A}p_{\Err(x)}(z-y(x)) \, dz
\end{eqnarray}
and $p_{Y(x)}(z) = p_{\Err}(z-y(x)$ is the density of $P_{Y(x)}$.

Once we have settled our probabilistic model for a definition, we use it for dose mapping and apply it to a dose distribtion yielding a new random variable for dose. 
%% ---------------------------------------------------------------------------------
%% ---------------------------------------------------------------------------------
%% ---------------------------------------------------------------------------------

Before we turn to the dose mapping, we would like to turn to an additional intuitive strategy used for setting up mapping probabilities.

\subsubsection{In/Out-Strategy for Structure-Guided Mapping Probabilities}
We again exploit additional prior knowledge by assuming that source and target structures are known and matched during dose mapping. Let $S$ be a source structure, $S'$ the corresponding target structure, and $y(x)$ the predicted image point of $x \in S$. The previous strategies define a baseline certainty radius $C(x)$ and a corresponding search density around $y(x)$. If the correspondence $S \leftrightarrow S'$ is known, then admissible target locations must lie inside $S'$. It is therefore natural to condition the search distribution on the event $z \in S'$ and assign zero probability to points outside the target structure.

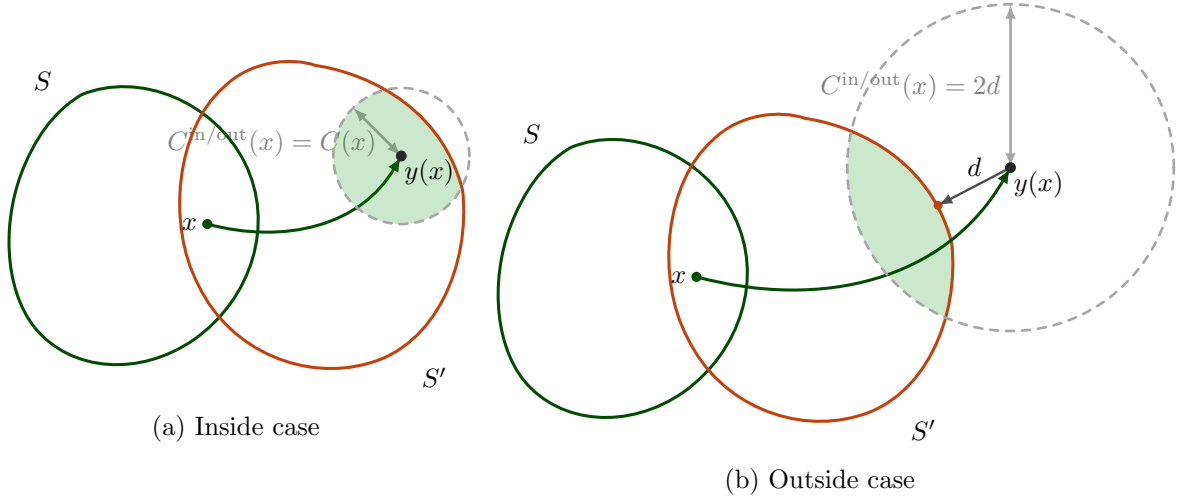
\begin{figure}[!htb]
    \centering
    \definecolor{FixedGreen}{HTML}{004C00}
    \definecolor{TargetOrange}{HTML}{C2410C}
    \def\SourceContour{%
        (1.10,4.35) .. controls (1.95,4.70) and (3.05,4.25) .. (3.50,3.35)
        .. controls (3.95,2.45) and (3.60,1.35) .. (2.75,0.75)
        .. controls (1.95,0.20) and (0.95,0.28) .. (0.42,0.92)
        .. controls (-0.05,1.50) and (-0.08,2.55) .. (0.28,3.35)
        .. controls (0.52,3.90) and (0.82,4.18) .. cycle%
    } 
    \def\TargetContour{%
        (4.54,4.78) .. controls (5.10,4.70) and (6.35,4.30) .. (6.70,2.95)
        .. controls (6.82,1.85) and (6.30,0.65) .. (5.30,0.40)
        .. controls (4.42,0.15) and (3.30,0.55) .. (2.82,1.55)
        .. controls (2.34,2.45) and (2.50,3.85) .. (3.18,4.45)
        .. controls (3.58,4.86) and (4.20,4.90) .. cycle%
    } 

    \begin{minipage}[c]{0.42\linewidth}
        \centering
        \begin{subfigure}{\linewidth}
            \centering
            \begin{tikzpicture}[scale=0.9,x=1cm,y=1cm,>=Latex,line cap=round,line join=round,font=\small,
                source/.style={draw=FixedGreen, line width=1.15pt},
                target/.style={draw=TargetOrange, line width=1.15pt},
                restricted/.style={fill=green!55!black, fill opacity=0.20, draw=none},
                support/.style={draw=gray!70, dashed, line width=1pt},
                dblarrow/.style={{Latex[length=1.9mm]}-{Latex[length=1.9mm]}, line width=1pt, draw=gray!70},
                maparrow/.style={-{Latex[length=2.1mm]}, line width=1.15pt, draw=FixedGreen},
                pointS/.style={circle, fill=FixedGreen, inner sep=1.35pt},
                pointY/.style={circle, fill=black!85, inner sep=1.45pt}
            ]
                \def\InsideRadius{1}
                \coordinate (xA) at (2.95,2.45);
                \coordinate (yA) at (5.8,3.45);
                \coordinate (pA) at ($(yA)+0.707*(-\InsideRadius,\InsideRadius)$);
               
                \draw[maparrow] (xA) to[out=-15,in=240] (yA);
                 
                \draw[dblarrow] (yA) -- (pA);
                \node[gray,anchor=east] at ($(yA)!0.3!(pA)$) {$C^\text{in/out}(x)=C(x)$};
 
                \path[source] \SourceContour;
                \begin{scope}
                    \clip \TargetContour;
                    \fill[restricted] (yA) circle[radius=\InsideRadius];
                \end{scope}
                \path[target] \TargetContour;                
                \draw[support] (yA) circle[radius=\InsideRadius];
                \node[pointS] at (xA) {};
                \node[pointY] at (yA) {};
                \node[anchor=east] at (xA) {$x$};
                \node[anchor=north west,inner sep = 1pt] at (yA) {$y(x)$};
                \node[anchor=west] at (0.25,4.55) {$S$};
                \node[anchor=west] at (5.95,0.20) {$S'$};
            \end{tikzpicture}
            \caption{Inside case}
            \label{fig:structure-guided-tikz-inside}
        \end{subfigure}
    \end{minipage}%\hfill
    \begin{minipage}[c]{0.55\linewidth}
        \centering
        
        \begin{subfigure}{\linewidth}
            \centering
            \begin{tikzpicture}[scale=0.9,x=1cm,y=1cm,>=Latex,line cap=round,line join=round,font=\small,
                source/.style={draw=FixedGreen, line width=1.15pt},
                target/.style={draw=TargetOrange, line width=1.15pt},
                restricted/.style={fill=green!55!black, fill opacity=0.20, draw=none},
                support/.style={draw=gray!70, dashed, line width=1pt},
                supportold/.style={draw=gray!45, dashed, line width=0.9pt},
                dblarrow/.style={{Latex[length=1.9mm]}-{Latex[length=1.9mm]}, line width=1pt, draw=gray!70},
                maparrow/.style={-{Latex[length=2.1mm]}, line width=1.15pt, draw=FixedGreen},
                proj/.style={-{Latex[length=1.9mm]}, line width=0.9pt, draw=black!70},
                pointS/.style={circle, fill=FixedGreen, inner sep=1.35pt},
                pointY/.style={circle, fill=black!85, inner sep=1.45pt},
                pointP/.style={circle, fill=TargetOrange, inner sep=1.2pt}
            ]
                \def\OutsideBaseRadius{0.95}
                \def\OutsideDistance{1.2}
                \def\OutsideRadius{2.4}
                \coordinate (xB) at (2.95,2.45);
                \coordinate (pB) at (6.5,3.5); % boundary point = projection of yB                
                \coordinate (xBhat) at (2.95,1.65); % small helper point, such that pB ist the projection of yBonto the target contour
                \coordinate (yB) at ($(pB)!-\OutsideDistance cm!(xBhat)$);

                \path[source] \SourceContour;
                \begin{scope}
                    \clip \TargetContour;
                    \fill[restricted] (yB) circle[radius=\OutsideRadius];
                \end{scope}
                \path[target] \TargetContour;
                \draw[maparrow] (xB) to[out=-15,in=240] (yB);
                \draw[support] (yB) circle[radius=\OutsideRadius];
                \draw[proj] (yB) -- (pB);
              
                \draw[dblarrow] (yB) -- ($(yB)+(0,\OutsideRadius)$);
                \node [gray,anchor=east] at ($ (yB)+(0,0.5*\OutsideRadius)$) {$C^\text{in/out}(x) = 2d$};
                \node[pointS] at (xB) {};
                \node[pointY] at (yB) {};
                \node[pointP] at (pB) {};
                \node[anchor=east] at (xB) {$x$};
                \node[anchor=north west, inner sep = 1pt] at (yB) {$y(x)$};
                \node[anchor=south] at ($(yB)!0.5!(pB)$) {$d$};
                
                % \node[anchor=north east] at (pB) {$\Pi$};
                \node[anchor=west] at (0.25,4.55) {$S$};
                \node[anchor=west] at (5.95,0.20) {$S'$};
            \end{tikzpicture}
            \caption{Outside case}
            \label{fig:structure-guided-tikz-outside}
        \end{subfigure}
    \end{minipage}
    \caption{
        Structure-aware illustrations of the inside and outside cases. In \subref{fig:structure-guided-tikz-inside}, the mapped point $y(x)$ lies inside $S'$, and the green region shows the admissible support of the conditioned density within the uncertainty radius $C(x)$, i.e., the region where $p_{Y(x)}^\text{in/out}>0$. In \subref{fig:structure-guided-tikz-outside}, $y(x)$ lies outside $S'$, $d$ denotes its distance to the boundary, and the admissible support extends to radius $2d$.  Note that in this case, the mapped (probabilistic) dose will not consider values near the originally mapped location $y(x)$, as these locations are considered implausible with a zero probability. 
    }
    \label{fig:structure-guided-strategy}
\end{figure}

Accordingly, we replace $p_{Y(x)}$ by the conditioned density $p^{\mathrm{in/out}}_{Y(x)}$, which is supported in $S'$. For $z\in S'$ we define
\begin{equation}
    p_{Y(x)}^\text{in/out}(z) := 
    \frac{p_{Y(x)}(z)\,\mathbf{1}_{S'}(z)}{\int_{S'} p_{Y(x)}(\xi)\,\mathrm{d}\xi}.
\end{equation}
This construction is well defined only if the support of $p_{Y(x)}$ intersects $S'$. Otherwise, the denominator vanishes. Here, we suggest to enlarge the radiusuntil a positive overlap is obtained. For the radial kernels considered here, the support is the ball $\{z:\|z-y(x)\|\leq C(x)\}$. If $y(x)\in S'$, intersection is automatic. If $y(x)\notin S'$, however, the support may miss $S'$ entirely. In that case, it suffices to enlarge the radius beyond the distance from $y(x)$ to $S'$. To remain conservative, we use
\begin{equation}
    C^\text{in/out}(x) := \max(C(x), 2\,\dist(y(x),S')).
\end{equation}
Figure \ref{fig:structure-guided-strategy} illustrates both cases. If $y(x)$ lies inside $S'$, the distribution is simply truncated and renormalized within $S'$. If $y(x)$ lies outside $S'$, the radius is first enlarged to guarantee overlap and the same conditioning step is then applied. Thus, the in/out strategy acts as a structure-aware post-processing step on top of the previously defined certainty-map strategies.

%%
%%
%%
%% ---------------------------------------------------------------------------------
%% ---------------------------------------------------------------------------------
%%
%%
%%
%%
%%
%%
%% ---------------------------------------------------------------------------------
%% ---------------------------------------------------------------------------------
%%
%%
%%

%% ---------------------------------------------------------------------------------
%%  Dose Mapping with Uncertainties
%% ---------------------------------------------------------------------------------

\subsection{Dose as random variable / Dose Mapping with Uncertainties}
\label{dose-mapping}
As mentioned above, common dose mapping means we apply a deformation to a given dose map $\dose:\R^d\to\R$ to propagate values from fraction to the baseline.

If the deformation is “only” a vector field, we can simply calculate the dose at the mapped point, and the propagated dose at $x$ is $\dose(y(x))$. In our probabilistic model, we now consider the case that $x$ is mapped to a random variable $Y(x)$. We no longer ask for a specific propagated value of the dose $\dose$ at $y(x)$, but rather for the probabilities with which the propagated dose $\dose(Y(x))$ takes values in a certain range, e.g., the interval $[a,b]\subset\R$ of Gy values. The probability in this case then is given by 
\begin{eqnarray}
\Pr[\dose(Y(x)) \in [a,b]]
& = & 
\Pr[Y(x) \in \dose^{-1}([a,b])]
\\ & = & 
P_{Y(x)}(\dose\in  [a,b])
\end{eqnarray}
with $\{\dose \in [a,b]\}$ as shorthand for the pre-image $\dose^{-1}([a,b])=\{z\,:\,\dose(z)\in [a,b]]\}$ of the interval $[a,b]$ of $\dose$. Thus, the propagated dose is a random variable by itself, given by
\begin{equation}
    D(x) := \dose(Y(x)) 
\end{equation}
with distribution $P_{D(x)}(\Theta) = \Pr[D(x)\in \Theta] = P_{Y(x)}(\dose\in\Theta)$.

\subsubsection{Computing probabilities and moments for dose values}
For calculating probabilities and statistics, an easy-to-calculate density function would be desirable. Unfortunately, this is not as simple as above, where we can derive a shifted version of the error distribution density for the deformation distribution and it turns out the probability is a weighted sum of all dose values that fall in the specific range. This is in general computationally demanding, as the computation for a single pixel requires a summing over all voxels or at least in some neighborhood around $y(x)$ provide the local error and $\{p_{\Err}>0\}$, respectively, is bounded:
\begin{eqnarray}
P_{D(x)}(\Theta) 
&=&
P_{Y(x)}(\dose \in\Theta)
\\
&=&
\int_{\{\dose \in\Theta\}} p_{\Err}(z-y(x))\,dz
\\
&=&
\int_{\R^3} \mathbf{1}_{\Theta}\big(\dose(z+y(x))\big) \, p_{\Err}(z)\,dz
\\
&=&
 \int_{\{p_{\Err}>0\}} \mathbf{1}_{\Theta}\big(\dose(z+y(x))\big) \, p_{\Err}(z)\,dz
\end{eqnarray}
with indicator function $\mathbf{1}_\Theta(t)=1$ iff $t\in \Theta$ else $0$.

In addition to probabilities, the calculation of expected values, such as  mean and standard deviation of dose, is of interest. Generally, for a measurable function $f:\mathbb{R}\to\mathbb{R}$, the expected value of $f(D(x))$ is given by
\begin{eqnarray*}
    \mathrm{E}[f(D(x))] &=& \int_\R f(t)\,dP_{D(x)}(t) 
    \\
    &=&
    \int_{\R^3} f(\dose(z))\,dP_{Y(x)}(z) 
    \\
    &=&
    \int_{\R^3} f(\dose(z+y(x)))\,p_{\Err}(z) \,dz,
\end{eqnarray*}
where the transformation rule/variable change is used (see, for example, \cite[]{Hoffmann1994a} or \cite[§19]{Bauer2001}). The mean and standard deviation are therefore given by
\begin{eqnarray}   \label{eq:mean_std:mean}
\mu(x) &=& \int_{\{p_{\Err}>0\}} \dose(z+y(x)) \,p_{\Err}(z)\,dz
\\
\label{eq:mean_std:std}    
\sigma^2(x) &=& \int_{\{p_{\Err}>0\}} \big|\dose(z+y(x))-\mu(x)\big|^2 \,p_{\Err}(z)\,dz.
\end{eqnarray}

%%%%%%%%%%%%%%%%%%%%%%%%%%%%%%%%%%%%%%%%%%%%%%%%%%%%%%%%%%%%%%%%%%%%%%%%%%%%%%%%%%%%%%%%%%%%%%%%%%%%
%%%%%%%%%%%%%%%%%%%%%%%%%%%%%%%%%%%%%%%%%%%%%%%%%%%%%%%%%%%%%%%%%%%%%%%%%%%%%%%%%%%%%%%%%%%%%%%%%%%%
%%%%%%%%%%%%%%%%%%%%%%%%%%%%%%%%%%%%%%%%%%%%%%%%%%%%%%%%%%%%%%%%%%%%%%%%%%%%%%%%%%%%%%%%%%%%%%%%%%%%
%%
\subsubsection{Confidence Bounds}\label{confidence-bounds}

In the case where the dose $D(x)$ is a random variable, we are interested in bounds for a given confidence level $\alpha \in [0,1]$. Specifically, we want the largest lower bound $a$ and the smallest upper bound $b$ such that 
\begin{equation}
 \Pr[D(x)\geq a]\geq\alpha \quad \text { and } \quad \Pr[D(x)\leq b]\geq \alpha
\end{equation}
Here, we define
\begin{eqnarray}
    \label{eq:confidence-bounds:min}
	D^{\min}_\alpha(x) &:=& \text{lower confidence bound} = \sup\{a \in \mathbb{R} : \Pr[D(x) \ge a] \ge \alpha\}, \\
	\label{eq:confidence-bounds:max}
	D^{\max}_\alpha(x) &:=& \text{upper confidence bound}  = \inf\{b \in \mathbb{R} : \Pr[D(x) \le b] \ge \alpha\}.
\end{eqnarray}
such that by construction, $\Pr[D(x) \ge D^{\min}_{\alpha}(x)] \ge \alpha$ and $\Pr[D(x) \le D^{\max}_{\alpha}(x)] \ge \alpha$. 
Note, that these bounds are the common $\alpha$-percentiles of the distribution of $D(x)$ for the general case of distribution functions that are not strictly monotonic or have discontinuities. Otherwise, the commulative distribution function $F_{D(x)}(t):=\Pr[D(x)\leq t]$ is continuous and invertible, such that
\begin{equation}
    D_\alpha^{\max}(x)= F_{D(x)}^{-1}(\alpha) \quad \text{ and } \quad D_\alpha^{\min}(x)= F_{D(x)}^{-1}(1-\alpha).
\end{equation}

For the special case of 100\%-percentiles, these bounds reduce to the min- and  maximum achievable dose values over the range of possible deformation $Y(x)$. To this end, we define the set of all possible outcomes of $Y(x)$ as
\begin{equation*}
S(x):=\{p_{Y(x)}>0\},
\end{equation*}
such that $\Pr[Y(x)\in S(x)]=P_{Y(x)}(S(x))=1$. 
Now, assume  $a\in\R$ is a lower bound for dose satisfying $\Pr[D(x)\geq a]=P_{Y(x)}(\dose\geq a)=1$ such that 
\begin{equation}
    1= \int_{\{\dose\geq a\}} p_{Y(x)}(z)\,dz = \int_{\R^3} \mathbf{1}_{\{\dose\geq a\}}(z)\, p_{Y(x)}(z)\,dz
\end{equation}
Since the indicator function and the density take only values greater or equal to zero this can only hold if and only if $\mathbf{1}_{\{\dose\geq a\}}\,p_{Y(x)}=p_{Y(x)}$ almost everywhere. In particular, looking at the subset $S(x)=\{p_{Y(x)}>0\}$ we must have $\mathbf{1}_{\{\dose\geq a\}}=1$ almost everywhere, or equivalently $\dose(z)\geq a$ for almost all $z\in S(x)$. Thus, (under mild regularity assumptions on $\dose$, e.g., $\dose$ is piecewise continuous) $a \leq \inf_{z\in S(x)}\dose(z)$ and the largest such $a$ is $a=\inf_{z\in S(x)}\dose(z)$. Accordingly, the smallest upper bound $b$ with $\Pr[D(x)\leq b]=1$ is given by the supremum of the dose on $S(x)$ such that we may state 
\begin{equation}
    \label{eq:min_max_dose}
    D_{100\%}^{\min}(x) =\inf_{z\in S(x)}\dose(z) 
    \quad\text{ and }\quad
    D_{100\%}^{\max}(x) =\sup_{z\in S(x)} \dose(z).
\end{equation} 
Note that the 100\% bounds depend only on the support $S(x)$ of distribution $P_{Y(x)}$ (i.e., where $p_{Y(x)}>0$). They do not depend on how probability is distributed within that support. As long as the support does not change, the actual distribution and density are irrelevant. Consequently, when modeling uncertainties and the distribution of $Y(x)$, if we are only interested in 100\% confidence bounds, we only need to consider the size of the support.

%% ---------------------------------------------------------------------------------
%% ---------------------------------------------------------------------------------
%% ---------------------------------------------------------------------------------
%%  Dose Volume Histograms with Uncertainties
%% ---------------------------------------------------------------------------------
%% ---------------------------------------------------------------------------------
%% ---------------------------------------------------------------------------------

\subsection{Dose Volume Histograms}\label{dose-volume-histograms}

The dose volume histogram (DVH) of a structure $\Sigma\subset\R^3$ and a scalar dose map $\dose:\R^3\to\R$ is the relative volume fraction of $\Sigma$ that receives at least dose level $t$, i.e.,
\begin{equation}
\operatorname{DVH}(\dose,\Sigma,t)=\frac{|\{x\in\Sigma\,:\,\dose(x)\geq t\}|}{|\Sigma|}
=
\frac{1}{|\Sigma|}\int_\Sigma \mathbf{1}_{\{\dose\ge t\}}(x)\,dx.
\end{equation}
This formulation makes explicit that the DVH is completely determined by the underlying dose map: it counts how much of the structure lies above the threshold $t$. An important property, which we make use of later, is the that the DVH is monotone with respect to the dose map. That is, let $d:\R^3\to\R$ be a dose map that is point wise bounded by  maps $d_\text{lower}, d_\text{upper}$ such that $d_\text{lower}(x)\leq d(x) \leq d_\text{upper}(x)$ for all $x\in\Sigma$. Then, this implies, that for any fixed value of $t$  we must have 
$\{d_\text{lower}\geq t\} \subseteq \{ d \geq t\} \subseteq \{ d_\text{upper} \geq t\}$
and hence.
\begin{equation}
    \label{eq:dvh-monotonicity}
\operatorname{DVH}(d_\text{lower},\Sigma,t)\le \operatorname{DVH}(d,\Sigma,t) \le \operatorname{DVH}(d_\text{upper},\Sigma,t) \qquad \text{for all } t\in\R.
\end{equation}
In our setting, however, the mapped dose is not a scalar-valued function but the random field $D(x)=\dose(Y(x))$, and the DVH itself therefore becomes a random quantity. Unlike voxel-wise dose values, the DVH is not a pointwise quantity: it depends on the complete dose map on the structure $\Sigma$. Consequently, uncertainty in DIR induces uncertainty in the full DVH curve, not only in individual voxel doses. Strictly speaking, a particular realization $D_\omega$ of the propagated dose for some element $\omega$ from an underlying sample space $\Omega$ must be viewed as a complete function $x\mapsto D_\omega(x)$ on the spatial domain, not merely as a collection of unrelated voxel values.
The associated realized DVH is then
\begin{equation}
    \operatorname{DVH}(D_\omega,\Sigma,t)=\frac{1}{|\Sigma|}\int_\Sigma \mathbf{1}_{\{D_\omega(x)\ge t\}}\,dx.
\end{equation}
A fully probabilistic treatment of DVH uncertainty would therefore require a probability model on an appropriate space of dose functions or deformation fields and, in particular, probabilities for sets of complete realizations. Such a model is substantially more difficult to specify and calibrate in practice, because it requires assumptions on spatial dependence, regularity, and admissible joint variations over the whole domain.

To circumvent this difficulty, we deliberately work with the pointwise uncertainty model introduced above. That is, we model uncertainty voxel by voxel through the random variables $Y(x)$ and $D(x)=\dose(Y(x))$ and derive summary maps such as means, standard deviations, and confidence bounds pointwise in $x$. When these pointwise quantities are transferred to DVH level, the resulting DVH bands should be understood as induced by pointwise bounds as in \eqref{eq:dvh-monotonicity} rather than by a full stochastic model on the set of entire dose maps. This is computationally simple and clinically interpretable, but also tends to be conservative: because spatial coupling is not enforced, the resulting bounds are typically larger than what one would obtain from a realistic joint model of complete dose realizations.

However, the quantities introduced above provide two natural ways to summarize this uncertainty on DVH level. First, the moments of $D(x)$ yield central or error-margin type summaries. Using the mean map $\bar{D}(x)$~\eqref{eq:mean_std:mean} and standard deviation map $S(x)$~\eqref{eq:mean_std:std}, one may define deterministic surrogate dose maps  such as $\bar{D}(x)\pm c S(x)$ for some margin of error $c>0$,
and evaluate their DVHs,
\begin{equation}
\operatorname{DVH}(\bar D\pm c S,\Sigma,t)
\end{equation}
These curves provide an intuitive moment-based error margin around the DVH of the mean dose map $\operatorname{DVH}(\bar{D},\Sigma,t)$. They are easy to compute and useful for visualization, but they should be interpreted as descriptive or approximate uncertainty bands rather than rigorous confidence statements, unless additional distributional assumptions are imposed.

Second, the pointwise confidence bounds from Section~\ref{confidence-bounds} directly induce confidence-derived DVH bounds. For a given confidence level $\alpha$, we defined the lower and upper confidence-bound dose maps $D_\alpha^{\min}$~\eqref{eq:confidence-bounds:min} and $D_\alpha^{\max}$~\eqref{eq:confidence-bounds:max} and then compute their DVHs:
\begin{equation}\label{eq:dvh-confidence-bounds}
\operatorname{DVH}_\alpha^{\min}(\Sigma,t):=\operatorname{DVH}(D_\alpha^{\min},\Sigma,t),
\qquad
\operatorname{DVH}_\alpha^{\max}(\Sigma,t):=\operatorname{DVH}(D_\alpha^{\max},\Sigma,t).
\end{equation}
Because of the monotonicity~\eqref{eq:dvh-monotonicity} of the DVH with respect to the dose map, pointwise dose bounds immediately translate into DVH bounds: whenever a realized dose map $D_\omega$ satisfies
\begin{equation}
    D_\alpha^{\min}(x)\le D_\omega(x)\le D_\alpha^{\max}(x)
    \qquad \text{for all } x\in\Sigma,
\end{equation}
we obtain for every threshold $t$ that
\begin{equation}
\operatorname{DVH}_\alpha^{\min}(\Sigma,t)
\le
\operatorname{DVH}(D_\omega,\Sigma,t)
\le
\operatorname{DVH}_\alpha^{\max}(\Sigma,t).
\end{equation}
Thus, $D_\alpha^{\min}$ and $D_\alpha^{\max}$ define lower and upper DVH envelopes in exactly the same way as they define lower and upper voxel-wise dose envelopes.

For the special case $\alpha=100\%$, these bounds are deterministic and rigorous because $D_{100\%}^{\min}(x)$ and $D_{100\%}^{\max}(x)$ bound all admissible dose realizations at every voxel. Consequently,
\begin{equation}
\operatorname{DVH}(D_{100\%}^{\min},\Sigma,t)
\le
\operatorname{DVH}(D_\omega,\Sigma,t)
\le
\operatorname{DVH}(D_{100\%}^{\max},\Sigma,t)
\qquad \text{for all } t
\end{equation}
holds for every admissible realization $D_\omega$. For smaller confidence levels $\alpha<100\%$, the curves $\operatorname{DVH}_\alpha^{\min}$ and $\operatorname{DVH}_\alpha^{\max}$ should be interpreted as pointwise confidence-derived DVH margins. They are still highly informative in practice, but they are not automatically simultaneous confidence bands for the entire DVH curve unless additional assumptions are made.

%% ---------------------------------------------------------------------------------
%% ---------------------------------------------------------------------------------
%% ---------------------------------------------------------------------------------

\section{Experiments and Results}
\label{sec:experiments}
In the subsequent section, the concepts delineated above are illustrated using a clinical dataset. The objective of this study is not to provide a clinical evaluation of the concepts presented, but rather to emphasise their interrelationships and practical effects.

A CT dataset (434 x 262 x 163 voxels, with 0.79mm x 0.79mm x 2mm voxel size) was obtained from The Cancer Imaging Archive (TCIA) Prostate Anatomical Edge Cases repository~\cite{ThompsonEtAl2023data}. The CT dataset was annotated with segmentations and a clinically realistic prostate treatment plan dose was added with a 60Gy prescription. Image manipulation software was then used to modify the original images and structures to create a new anatomical scenario with a larger bladder contour, to simulate differences in bladder filling. The following experiments will utilise the segmentations of the bladder, rectum and prostate for certainty computation.

\begin{figure}[!htpb]
  \centering
  \begin{subfigure}[t]{0.475\textwidth}
    \centering
    \includegraphics[width=\textwidth, trim=0mm 28mm 0mm 55mm ,clip]{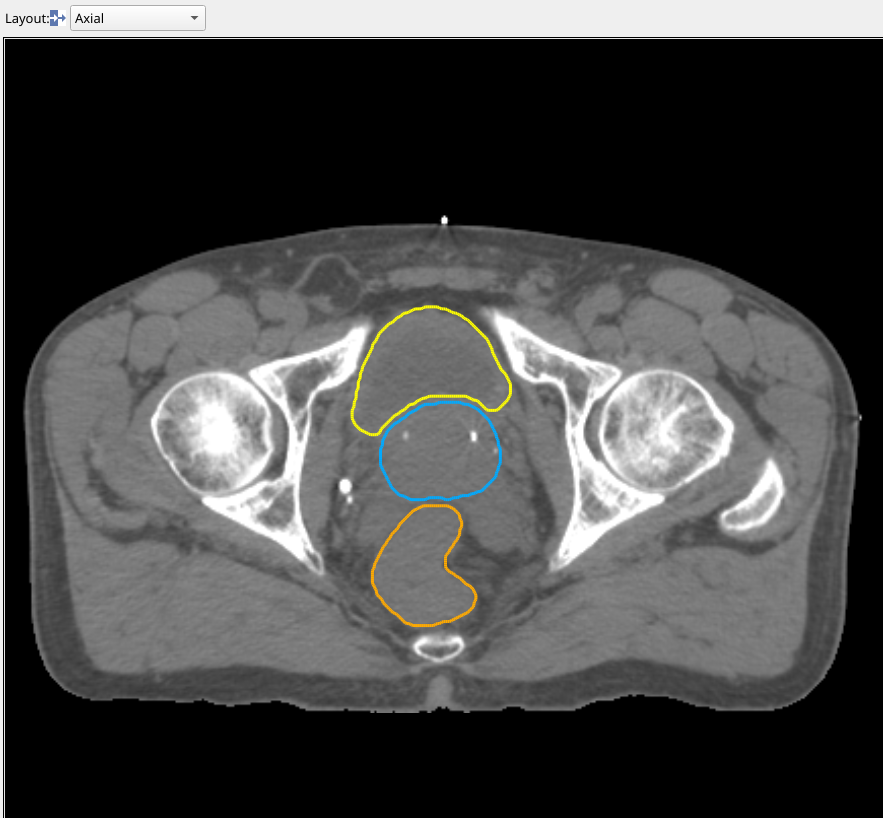}
    \caption{Baseline CT image with annotated anatomical structures (bladder, rectum, and prostate) defining the reference frame for registration and serving as the fixed image to which all other data are mapped.}
    \label{fig:baseline+fraction+dose:baseline}
  \end{subfigure}
\hfill
  \begin{subfigure}[t]{0.475\textwidth}
    \centering
    \includegraphics[width=\textwidth, trim=0mm 28mm 0mm 55mm ,clip]{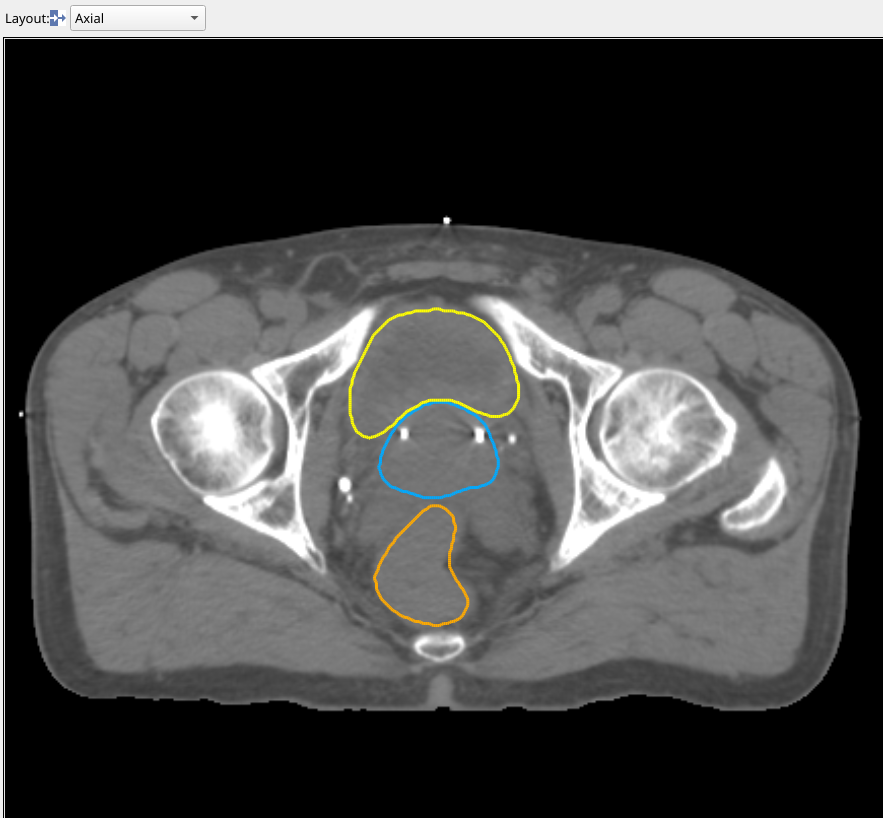}
    \caption{Fraction CT image of the treatment fraction with corresponding anatomical structures (bladder, rectum, and prostate).}
    \label{fig:baseline+fraction+dose:fraction}
  \end{subfigure}
  \begin{subfigure}[t]{0.475\textwidth}
    \centering
    \includegraphics[width=\textwidth, trim=0mm 28mm 0mm 55mm ,clip]{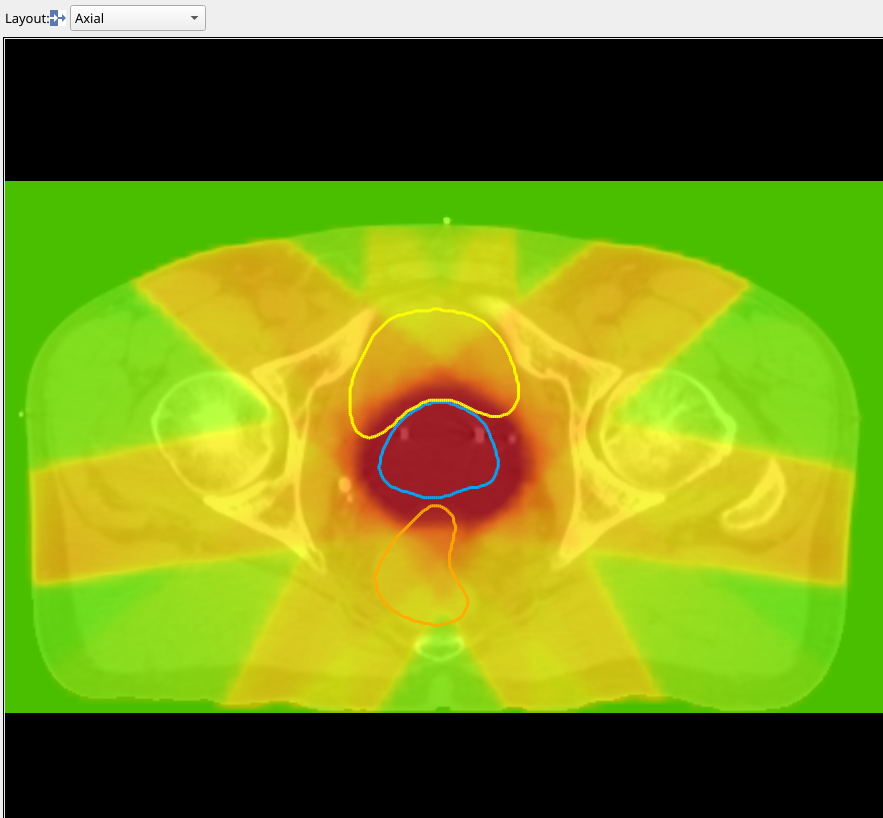} 
    \caption{Dose distribution of the current treatment fraction displayed on the fraction CT image and structures (bladder, rectum, and prostate). This dose is subsequently mapped to the baseline CT reference space for analysis.}
    \label{fig:baseline+fraction+dose:dose}
  \end{subfigure}
\hfill
  \begin{subfigure}[t]{0.475\textwidth}
    \centering
    \includegraphics[width=\textwidth,trim=39mm 48mm 39mm 80mm ,clip]{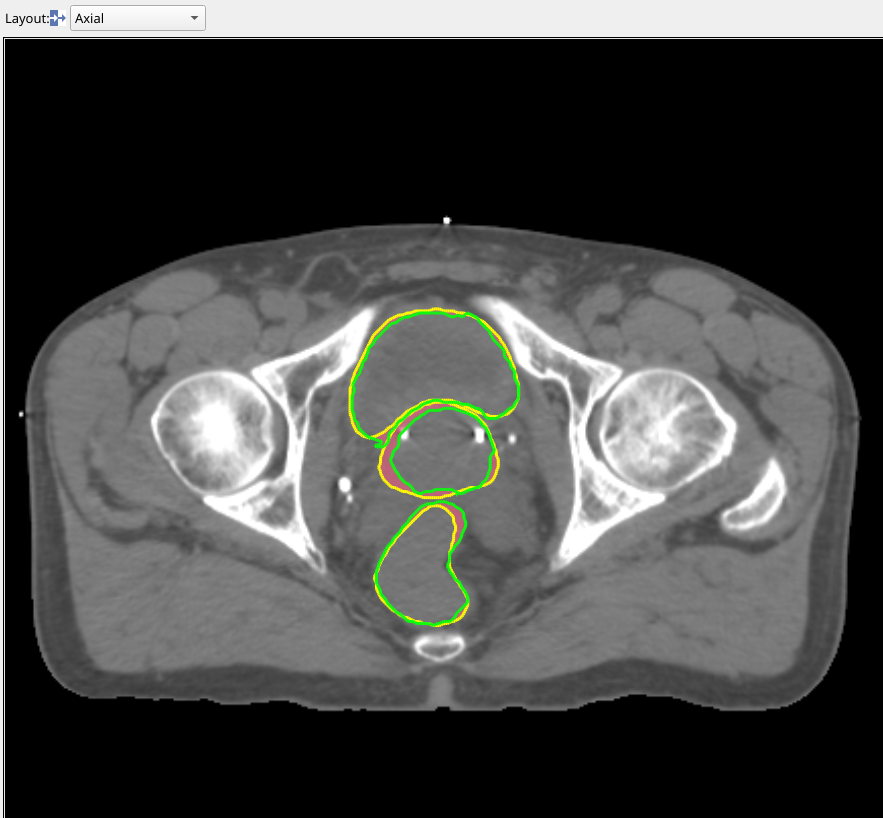}
    \caption{Magnified view of the fraction CT showing local misalignment (red area) between original fraction CT structures (yellow)  and forward-warped baseline CT structures (green).}
    \label{fig:baseline+fraction+dose:fraction_zoom}
  \end{subfigure}
  \caption{Overview of the experimental setup. The figure illustrates the main components of the study, including the baseline CT image, the registered fraction CT with associated dose distribution, and the relevant anatomical structures. This setup forms the basis for the subsequent analysis of dose propagation and uncertainty quantification.}
  \label{fig:baseline+fraction+dose} 
\end{figure}

In the experiments the original scan is the \emph{baseline CT} that defines the reference frame and the modified image is the \emph{fraction CT} that has been registered to the baseline using a standard non-linear registration method and also incorporates a dose image, which is employed for the purpose of calculations. Both images are shown in Figure~\ref{fig:baseline+fraction+dose}, 
%%
%%
%%  ------------------------------------------------------------------------------------
%%
%%
%%
%%
%%  ------------------------------------------------------------------------------------
%%
%%

%% ----------------------------------------------------------------------------------
%% ----------------------------------------------------------------------------------
%% ----------------------------------------------------------------------------------
%% ----------------------------------------------------------------------------------

\subsection{Comparing different strategies for modeling uncertainty}
In the first experiment, we compare three uncertainty settings:
\begin{itemize}
  \item Strategy 1: Globally constant uncertainty, as commonly used in the literature.
  \item Strategy 2: Distance-based uncertainty, where the uncertainty radius depends on the distance to the nearest structure boundary.
  \item Strategy 2 + in/out: The same distance-based certainty map as in Strategy 2, combined with the in/out conditioning from Section~2 so that mappings are restricted to the matched target structure.
\end{itemize}   
The comparison between Strategy 2 and Strategy 2 + in/out isolates the effect of the structure-guided post-processing, because both settings use the same local uncertainty radii and the same uniform kernel.

\begin{figure}[!htpb]
  \centering
   \setlength{\tabcolsep}{3pt}
   \def\imgheight{0.13\textheight}
   \def\imgwidth{0.21\textheight} % ratio of UncertaintyRadiusMap is width : height =  10:6 ~ 1.66 or 
  \scriptsize
  \begin{tabular}{rccc} 
    &  \normalsize Strategy 1  &  \normalsize Strategy 2   &  \normalsize Strategy 2 + in/out      \\
    &  \normalsize (globally constant) &  \normalsize (distance-based) & \normalsize (distance-based) \\
      \rotatebox{90}{\parbox{\imgheight}{\centering Certainity map}} &
    \includegraphics[height=\imgheight,width=\imgwidth]{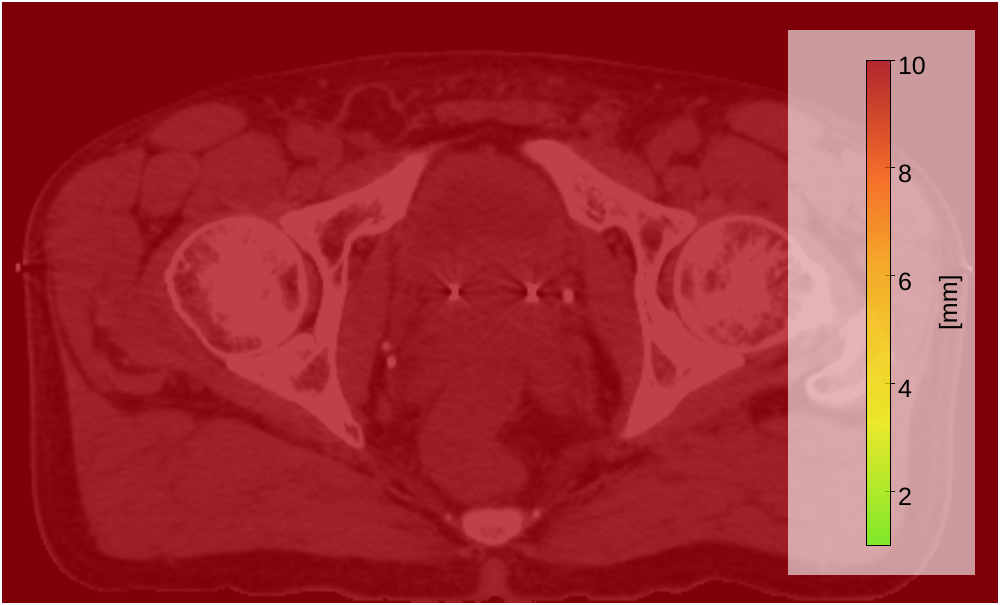} &
    \includegraphics[height=\imgheight,width=\imgwidth]{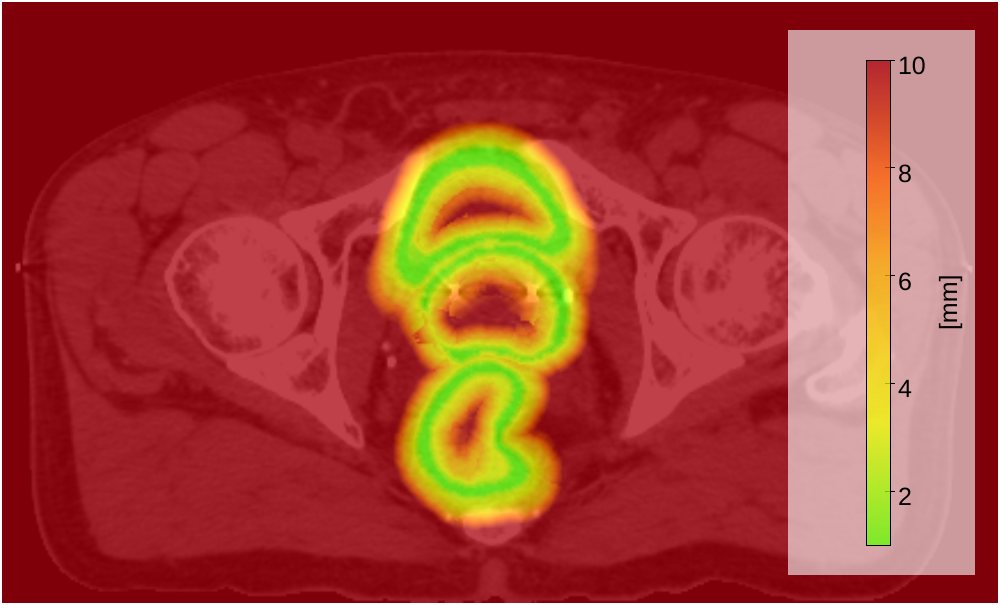} &
    \includegraphics[height=\imgheight,width=\imgwidth]{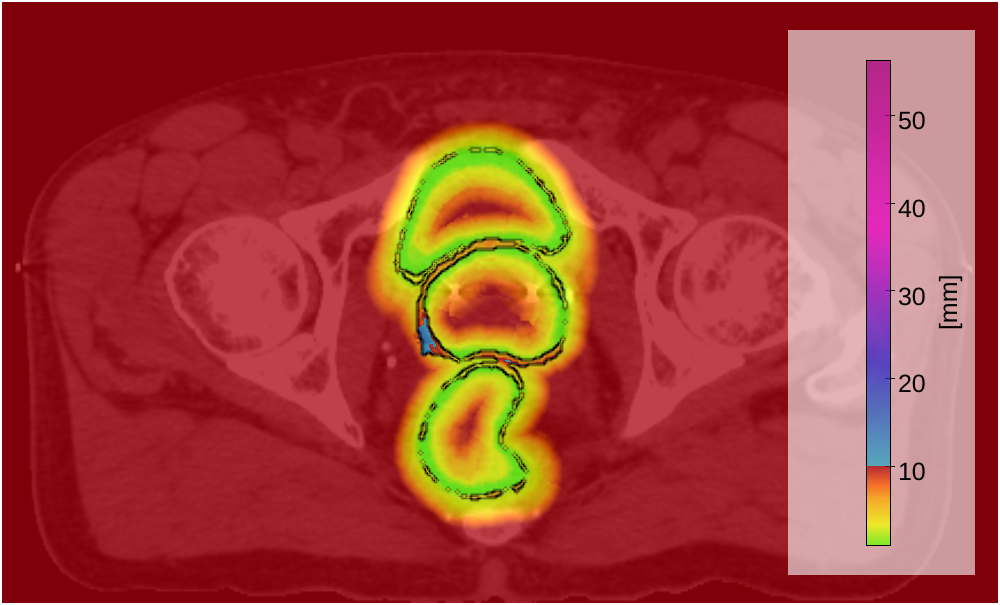} 
      \\
      \rotatebox{90}{\parbox{\imgheight}{\centering $\Pr[\dose(Y) \geq 60\,\mathrm{Gy}]$}} &
    \includegraphics[height=\imgheight,width=\imgwidth]{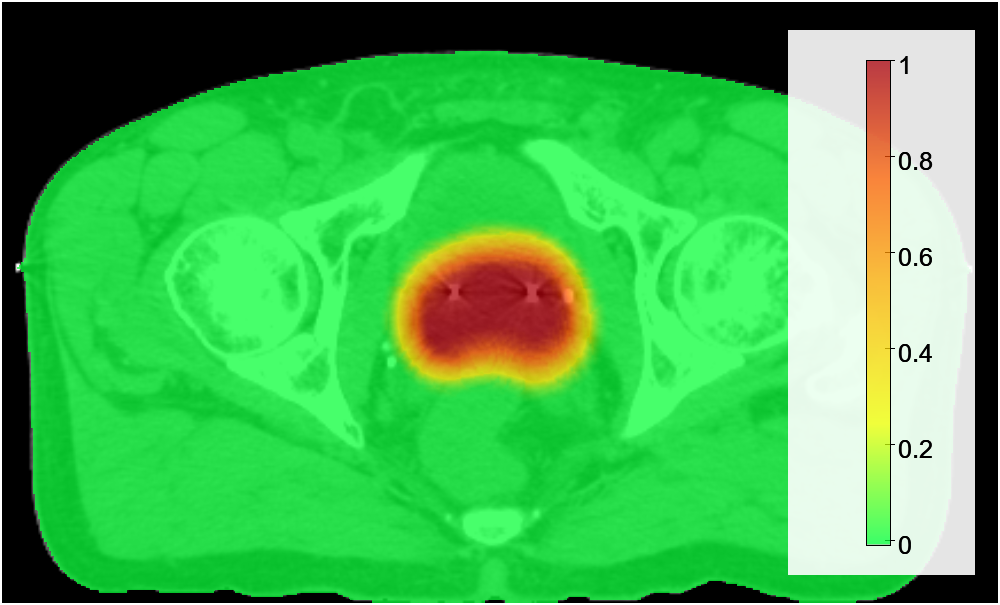} &
    \includegraphics[height=\imgheight,width=\imgwidth]{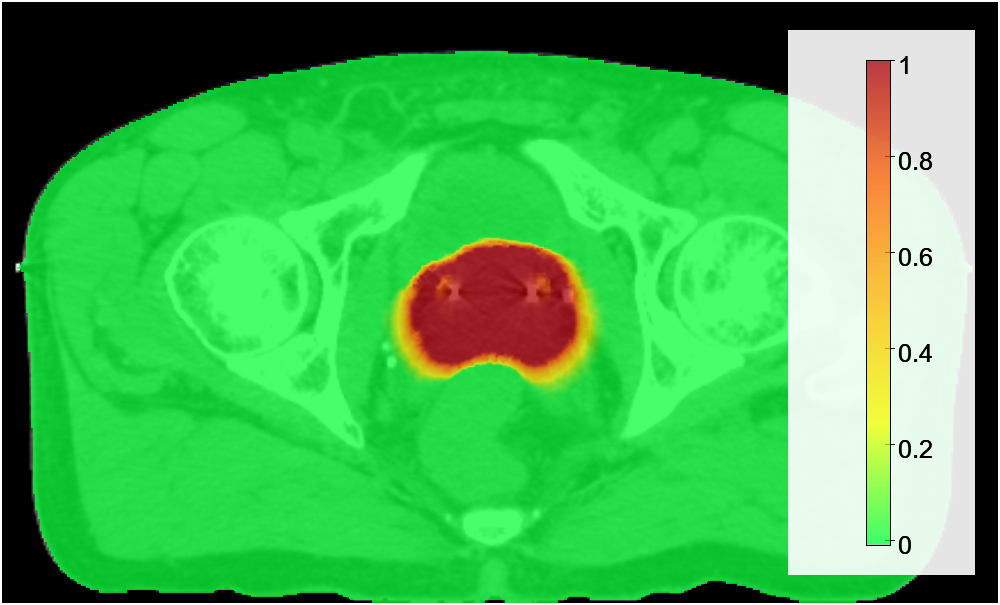} &
    \includegraphics[height=\imgheight,width=\imgwidth]{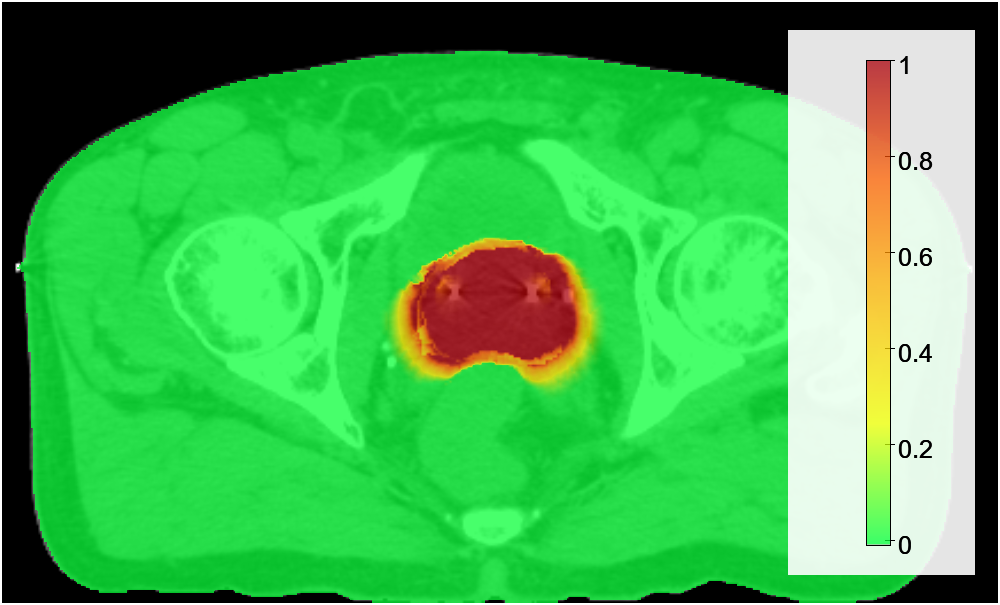} 
      \\
      \rotatebox{90}{\parbox{\imgheight}{\centering Upper 95\% Percentile}} &
    \includegraphics[height=\imgheight,width=\imgwidth]{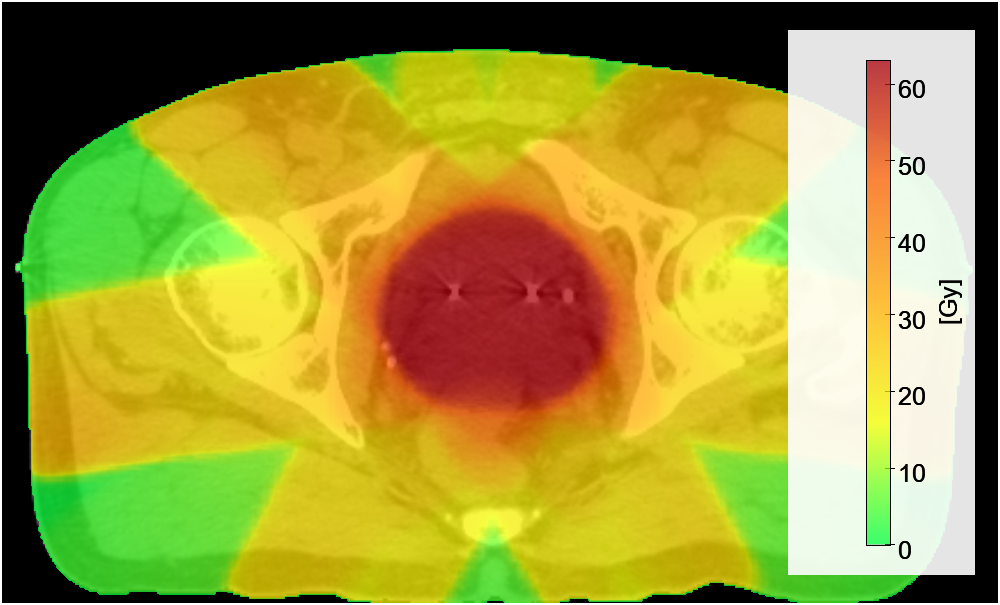} &
    \includegraphics[height=\imgheight,width=\imgwidth]{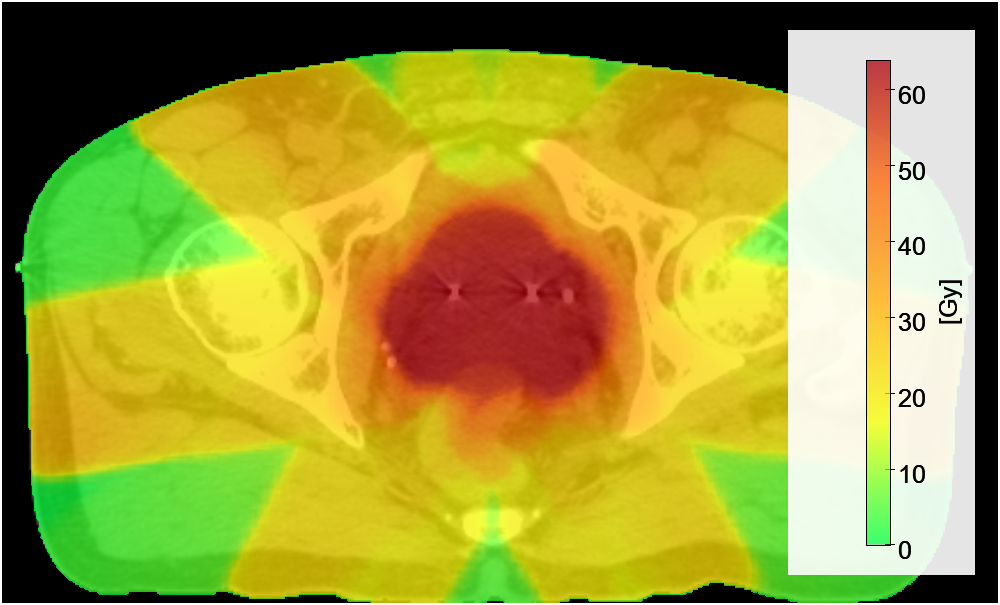} &
    \includegraphics[height=\imgheight,width=\imgwidth]{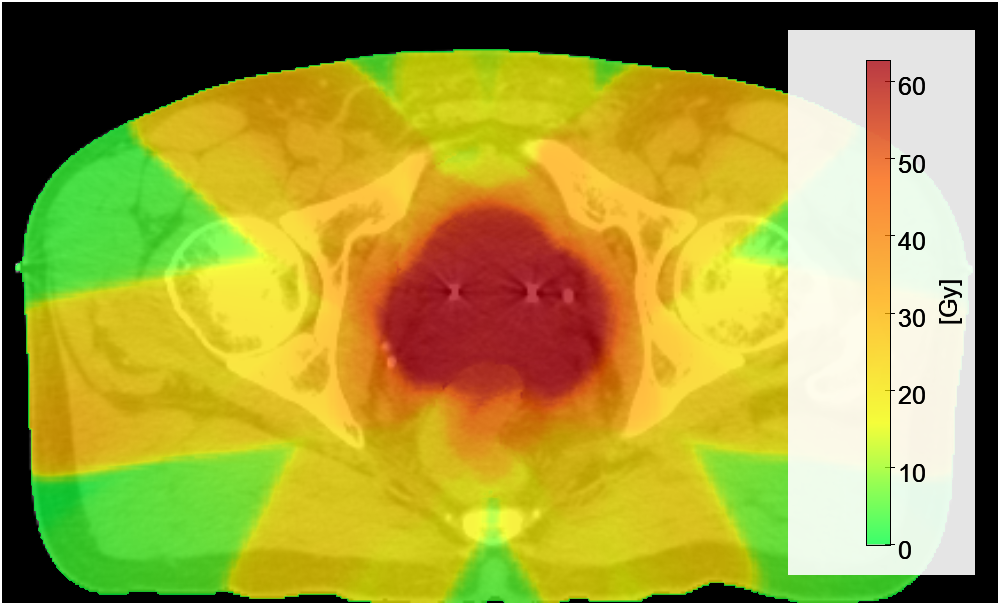} 
      \\
      \rotatebox{90}{\parbox{\imgheight}{\centering DVH Bladder}} &
    \includegraphics[height=\imgheight,width=\imgwidth,trim=10mm 5mm 40mm 30mm, clip]{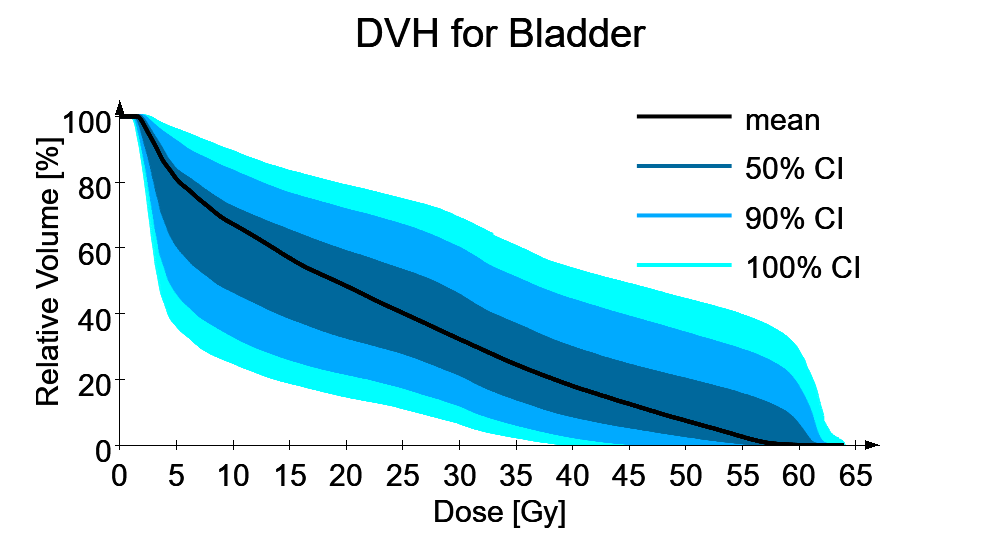} &
    \includegraphics[height=\imgheight,width=\imgwidth,trim=10mm 5mm 40mm 30mm, clip]{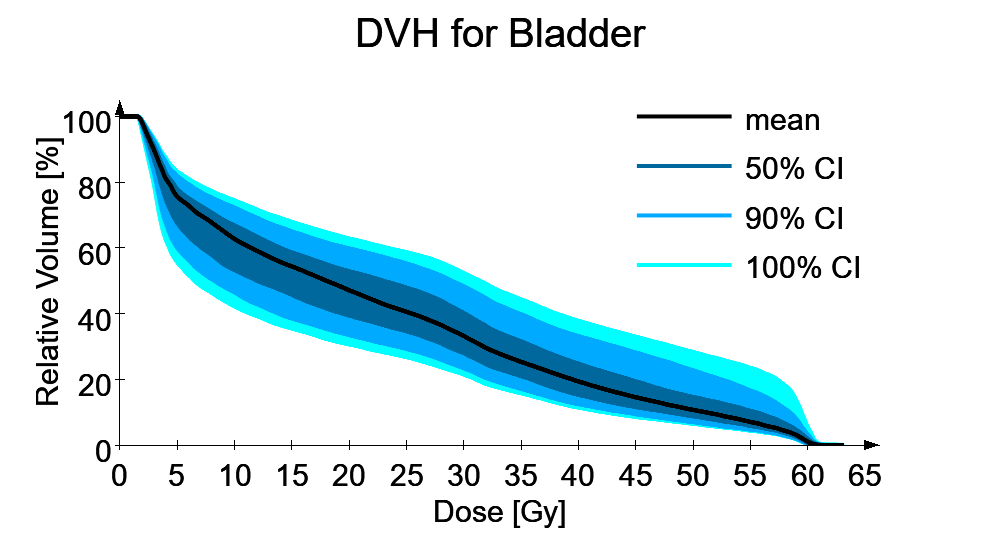} &
    \includegraphics[height=\imgheight,width=\imgwidth,trim=10mm 5mm 35mm 30mm, clip]{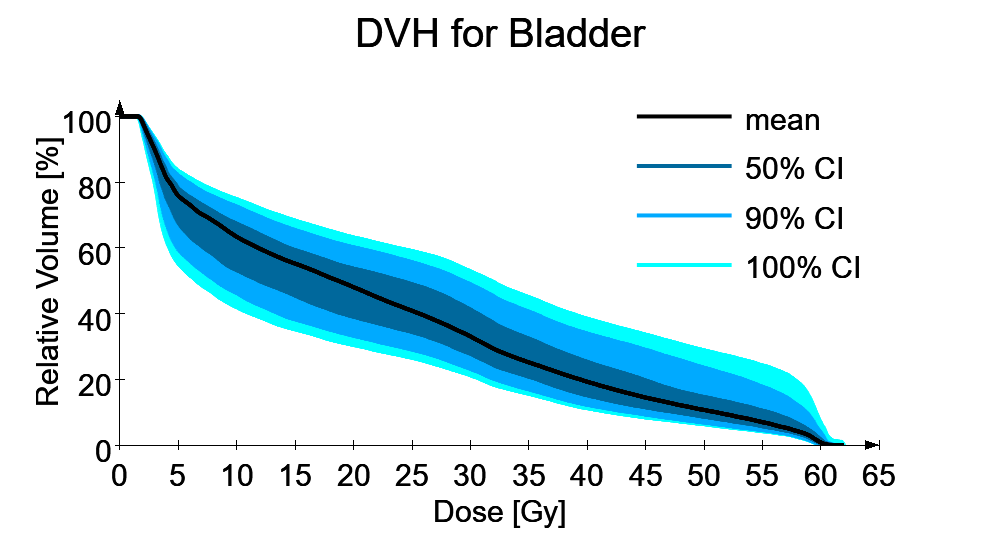} 
      \\
      \rotatebox{90}{\parbox{\imgheight}{\centering DVH Prostate}} &
    \includegraphics[height=\imgheight,width=\imgwidth,trim=10mm 5mm 40mm 30mm, clip]{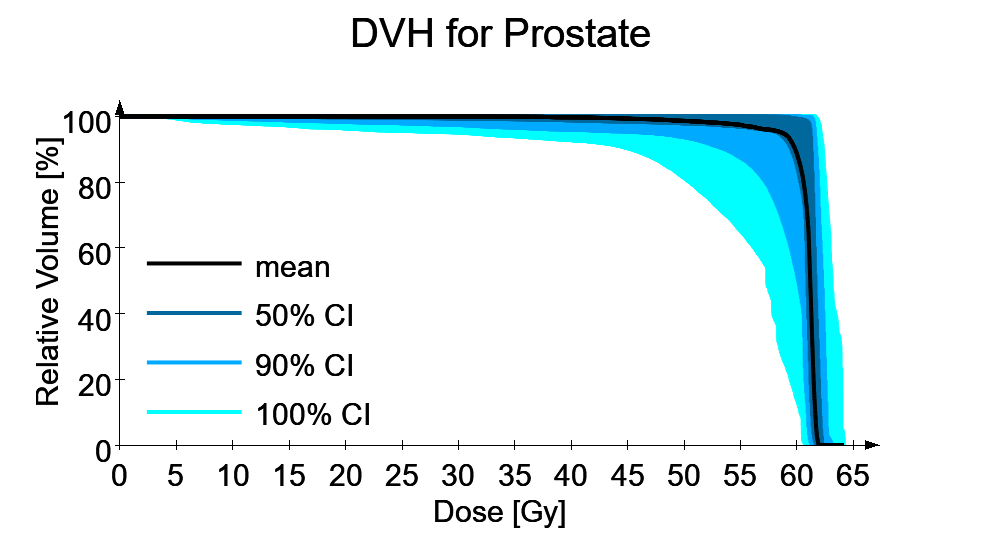} &
    \includegraphics[height=\imgheight,width=\imgwidth,trim=10mm 5mm 40mm 30mm, clip]{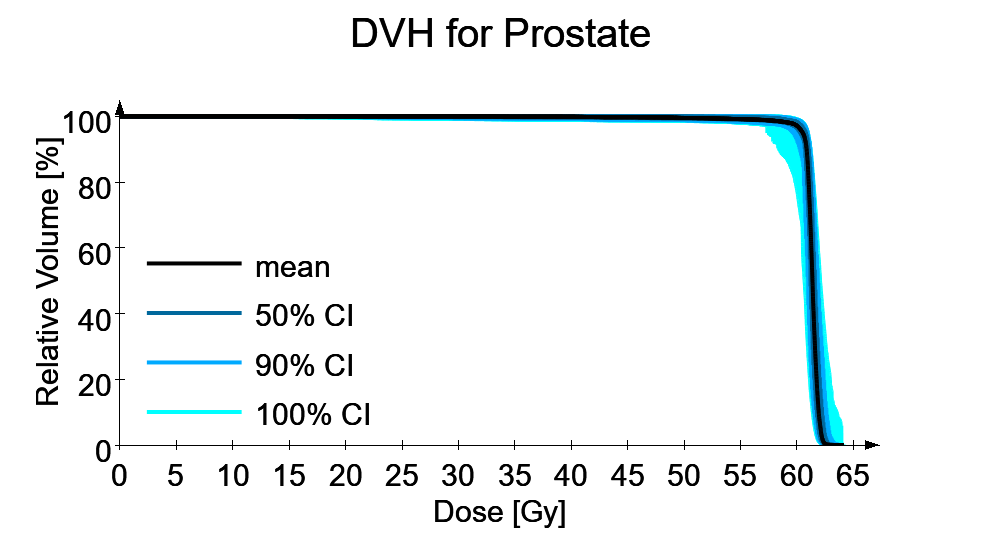} &
    \includegraphics[height=\imgheight,width=\imgwidth,trim=10mm 5mm 40mm 30mm, clip]{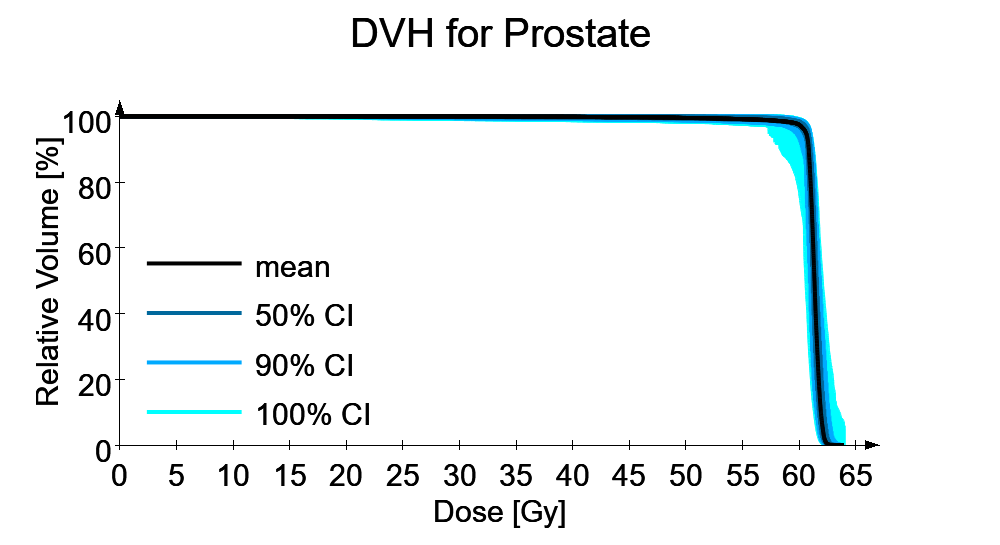} 
      \\
      \rotatebox{90}{\parbox{\imgheight}{\centering DVH Rectum}} &
    \includegraphics[height=\imgheight,width=\imgwidth,trim=10mm 5mm 40mm 30mm, clip]{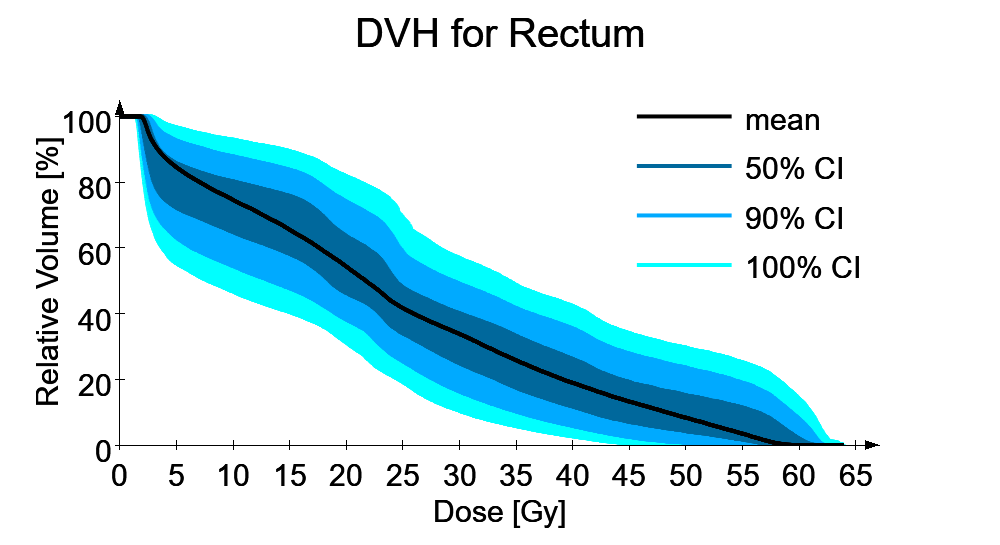} &
    \includegraphics[height=\imgheight,width=\imgwidth,trim=10mm 5mm 40mm 30mm, clip]{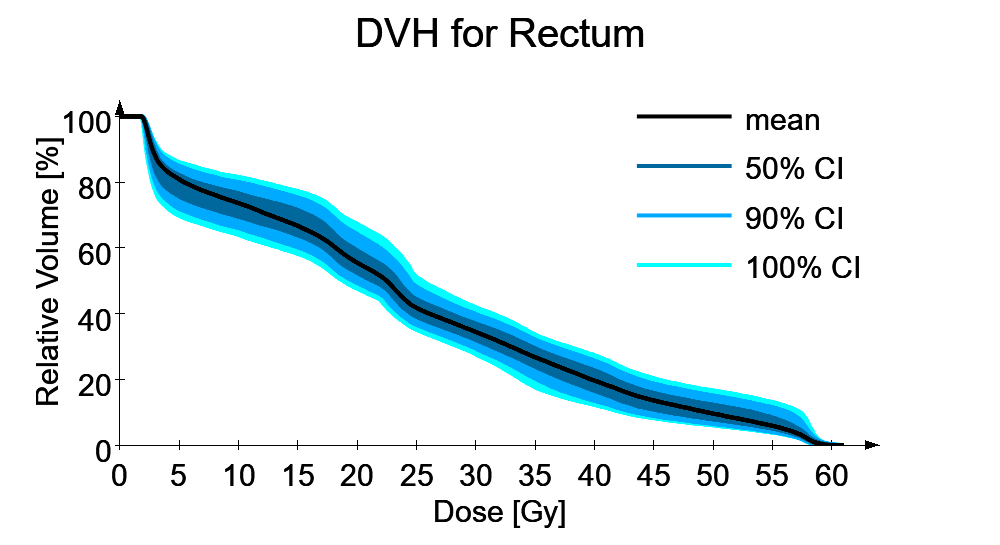} &
    \includegraphics[height=\imgheight,width=\imgwidth,trim=10mm 5mm 40mm 30mm, clip]{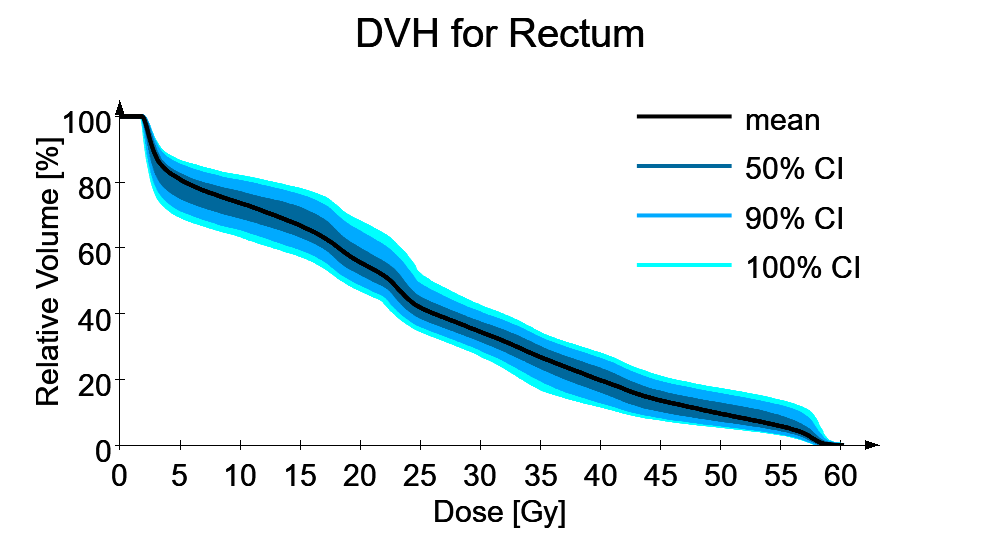} 
  \end{tabular}
  \caption{Comparison of uncertainty strategies. First column: strategy 1 with constant uncertainty radius; second column: strategy 2 with distance-based uncertainty radius; third column: same as strategy 2 in second column enhanced but with in/out consideration. In all examples the kernel is uniform and the max radius is 10mm.}  
  \label{fig:compare_strategies_10mm}
\end{figure}

\begin{figure}[!htpb]
  \centering
   \setlength{\tabcolsep}{3pt}
   \def\imgheight{0.13\textheight}
   \def\imgwidth{0.21\textheight} % ration of UncertaintyRadiusMap is width : height =  10:6 ~ 1.66 or 
  \scriptsize
  \begin{tabular}{rccc} 
    &  \normalsize Strategy 1  &  \normalsize Strategy 2   &  \normalsize Strategy 2 + in/out      \\
    &  \normalsize (globally constant) &  \normalsize (distance-based) & \normalsize (distance-based) \\
      \rotatebox{90}{\parbox{\imgheight}{\centering Certainity map}} &
    \includegraphics[height=\imgheight,width=\imgwidth]{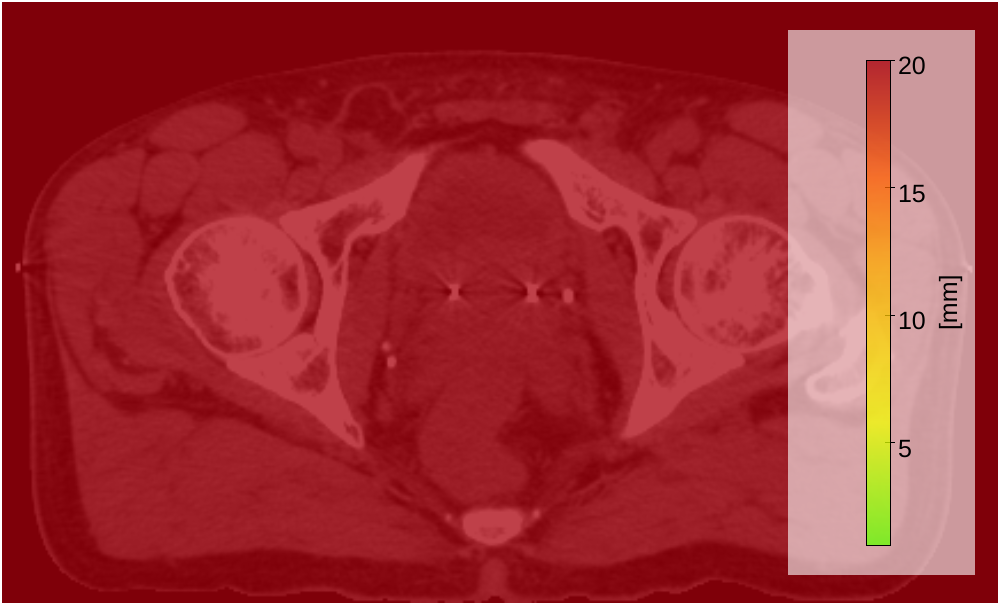} &
    \includegraphics[height=\imgheight,width=\imgwidth]{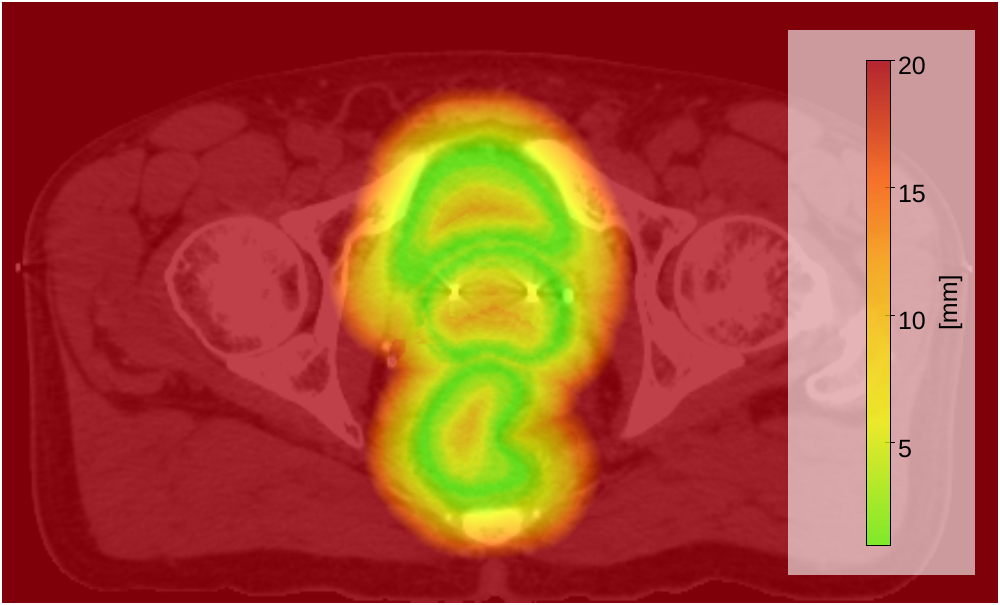} &
    \includegraphics[height=\imgheight,width=\imgwidth]{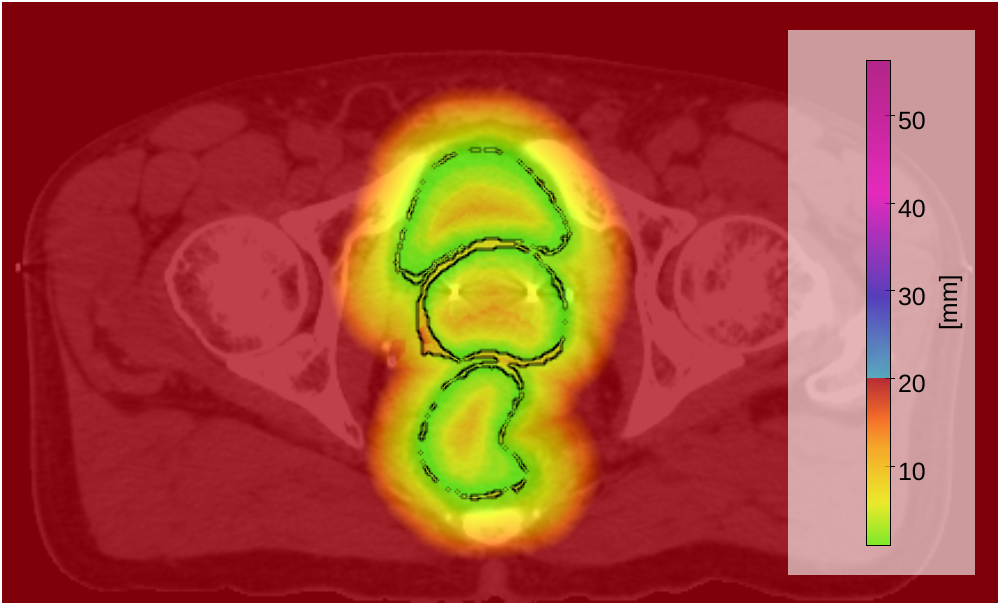} 
      \\
      \rotatebox{90}{\parbox{\imgheight}{\centering $\Pr[\dose(Y) \geq 60\,\mathrm{Gy}]$}} &
    \includegraphics[height=\imgheight,width=\imgwidth]{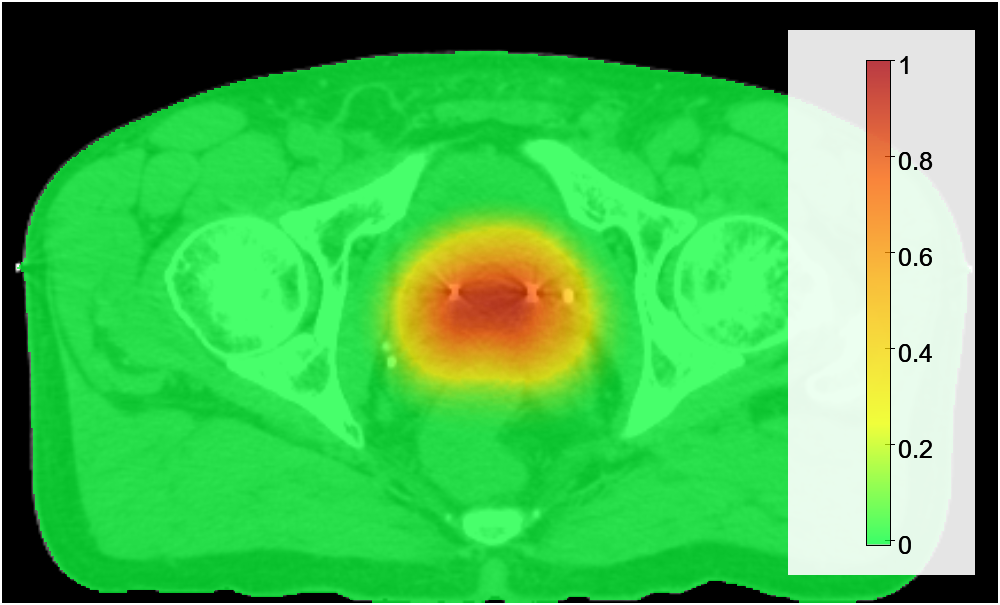} &
    \includegraphics[height=\imgheight,width=\imgwidth]{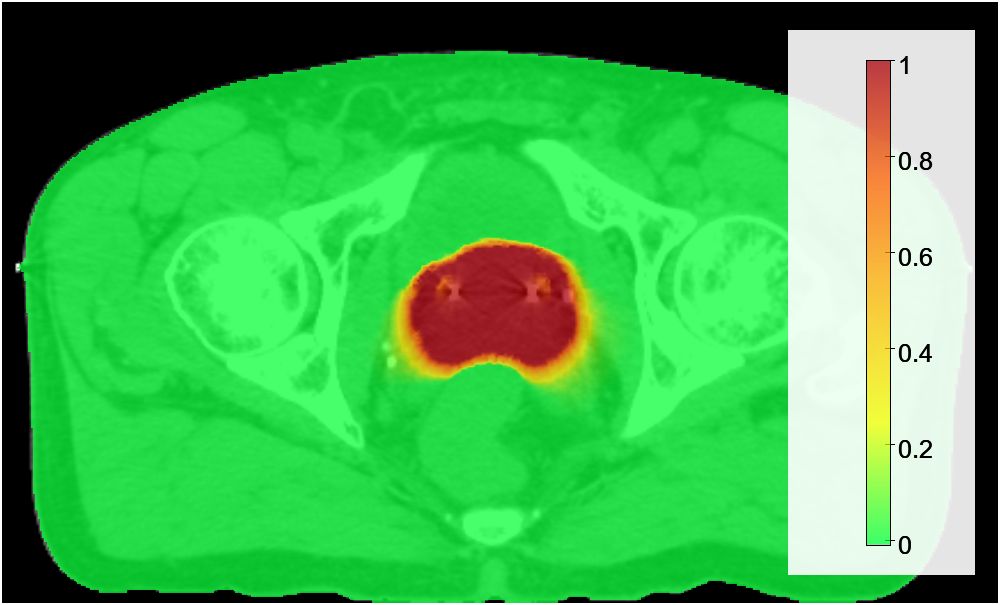} &
    \includegraphics[height=\imgheight,width=\imgwidth]{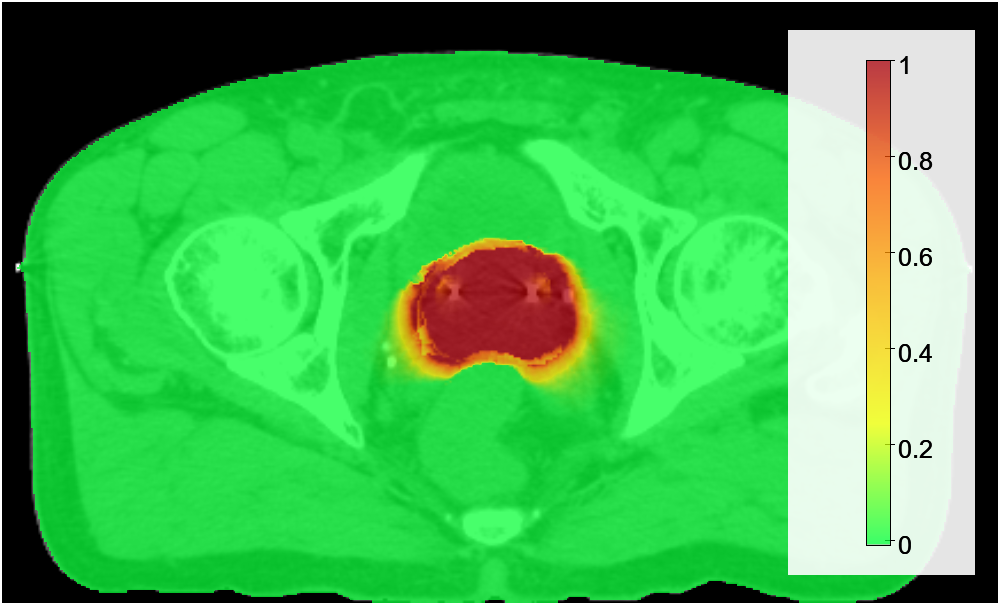} 
      \\
      \rotatebox{90}{\parbox{\imgheight}{\centering Upper 95\% Percentile}} &
    \includegraphics[height=\imgheight,width=\imgwidth]{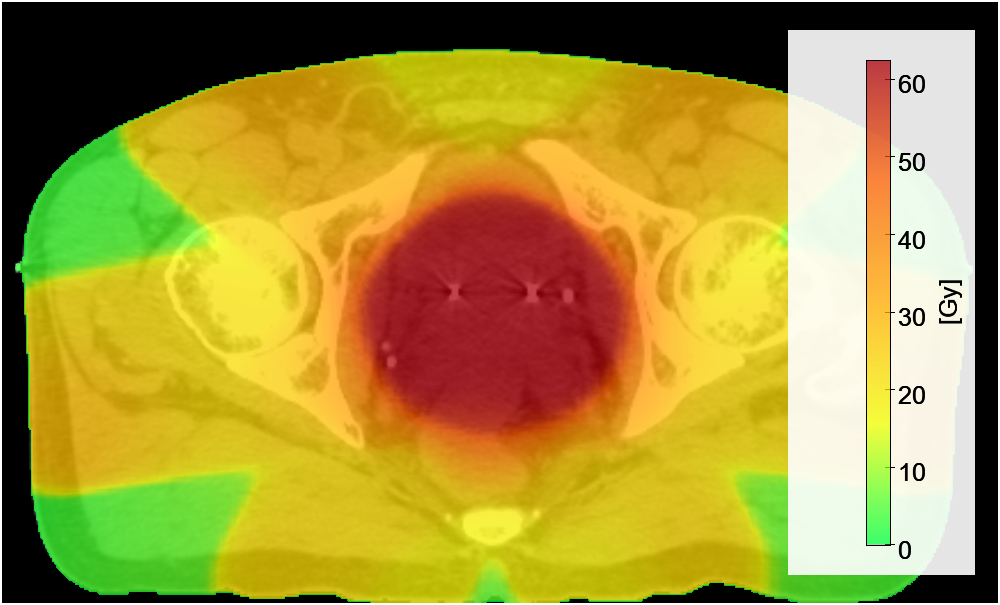} &
    \includegraphics[height=\imgheight,width=\imgwidth]{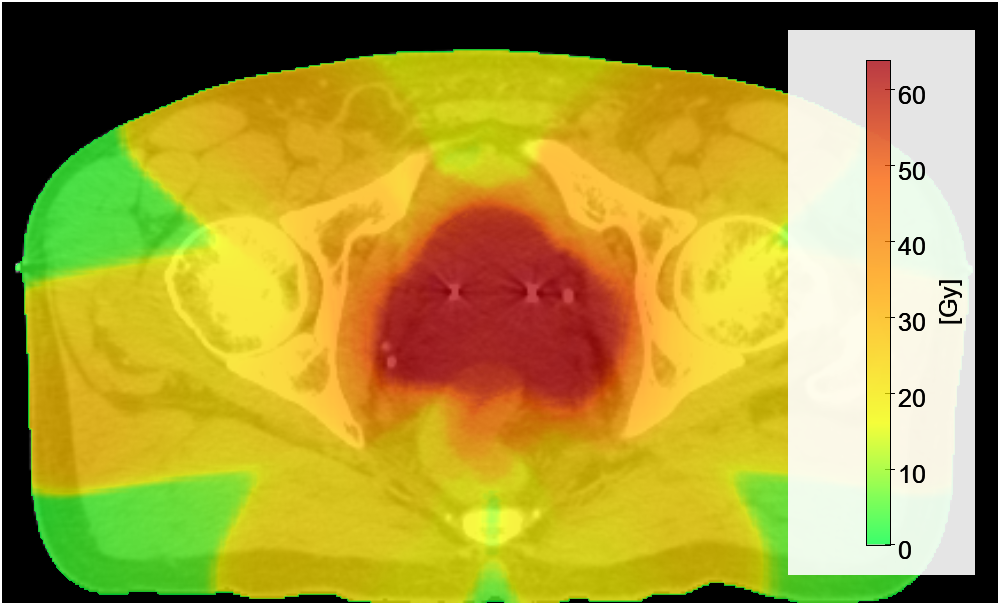} &
    \includegraphics[height=\imgheight,width=\imgwidth]{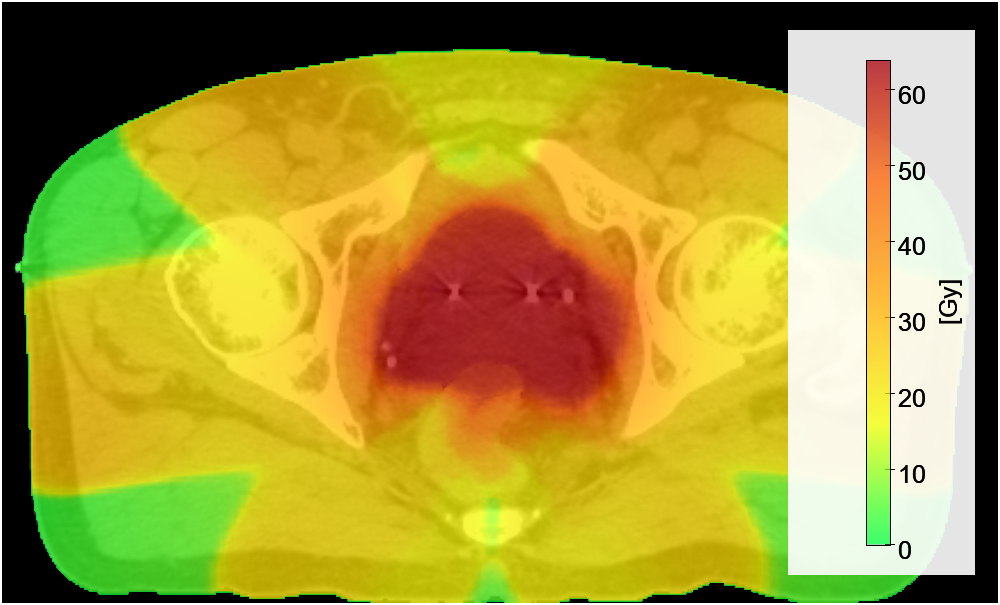} 
      \\
      \rotatebox{90}{\parbox{\imgheight}{\centering DVH Bladder}} &
    \includegraphics[height=\imgheight,width=\imgwidth,trim=10mm 5mm 40mm 30mm, clip]{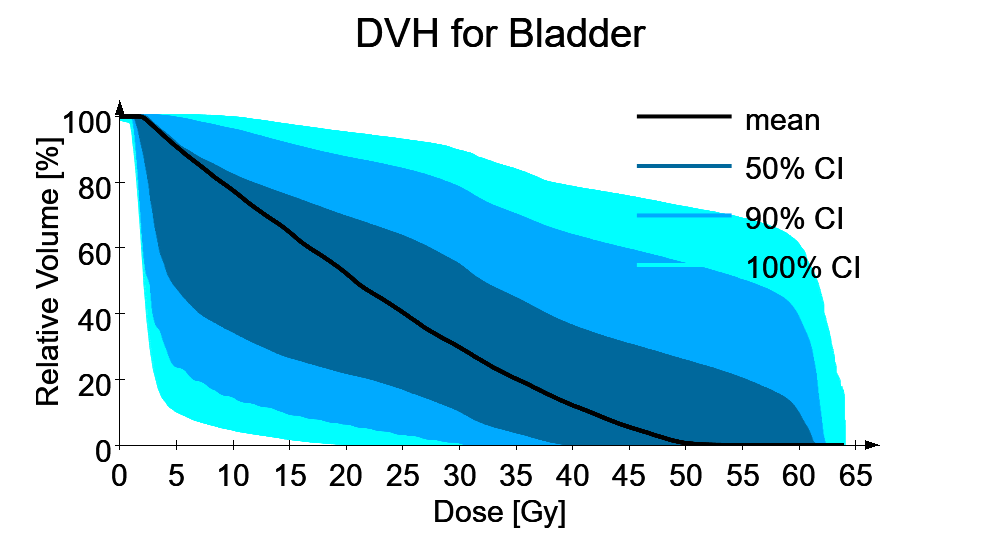} &
    \includegraphics[height=\imgheight,width=\imgwidth,trim=10mm 5mm 40mm 30mm, clip]{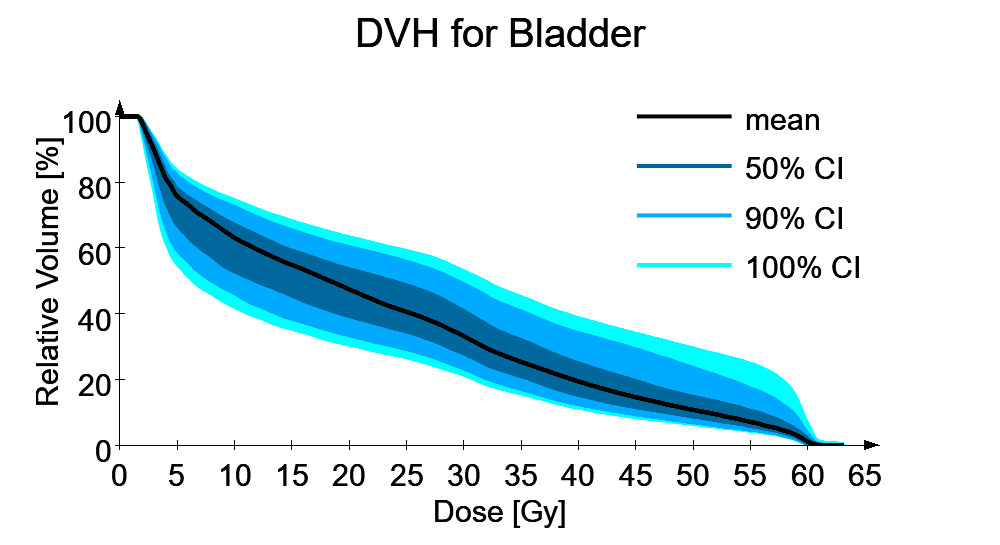} &
    \includegraphics[height=\imgheight,width=\imgwidth,trim=10mm 5mm 35mm 30mm, clip]{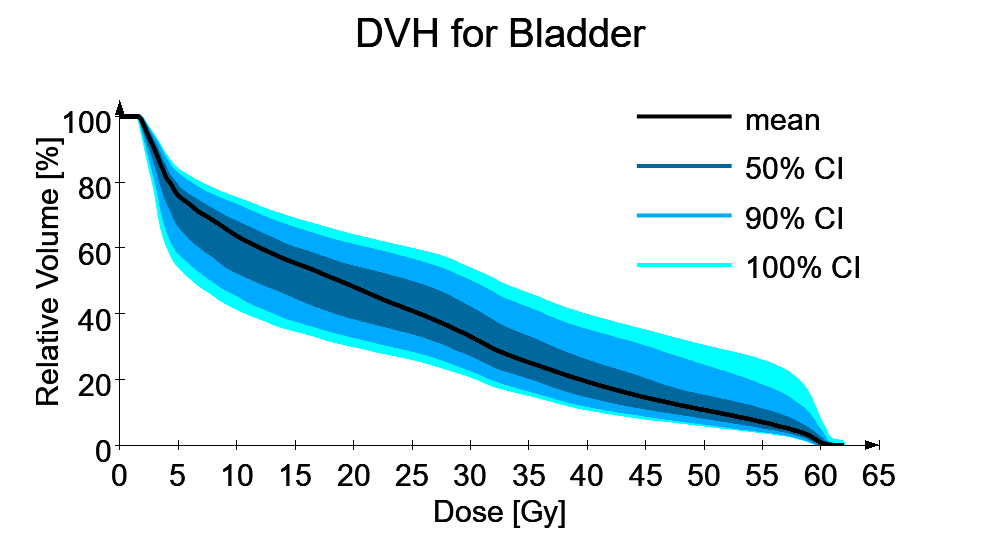} 
      \\
      \rotatebox{90}{\parbox{\imgheight}{\centering DVH Prostate}} &
    \includegraphics[height=\imgheight,width=\imgwidth,trim=10mm 5mm 40mm 30mm, clip]{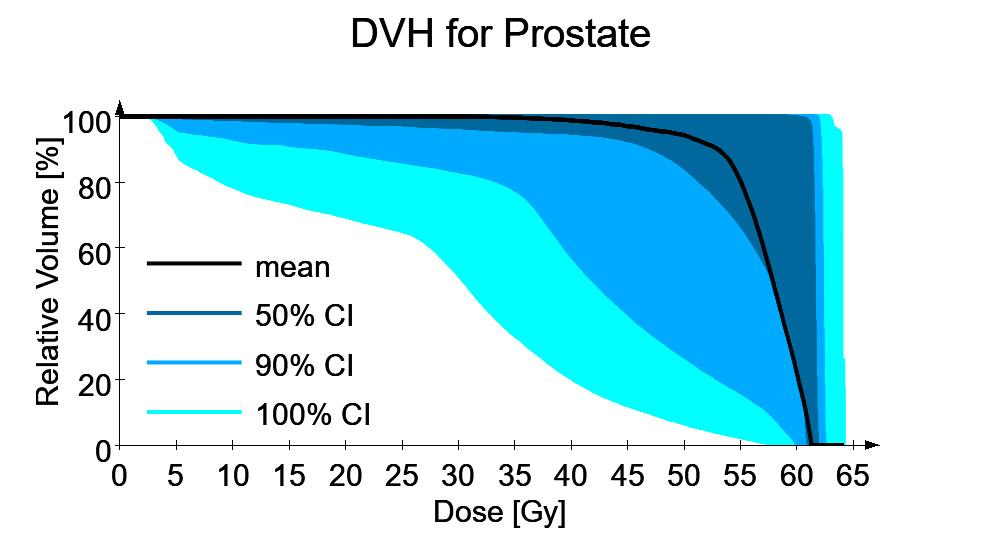} &
    \includegraphics[height=\imgheight,width=\imgwidth,trim=10mm 5mm 40mm 30mm, clip]{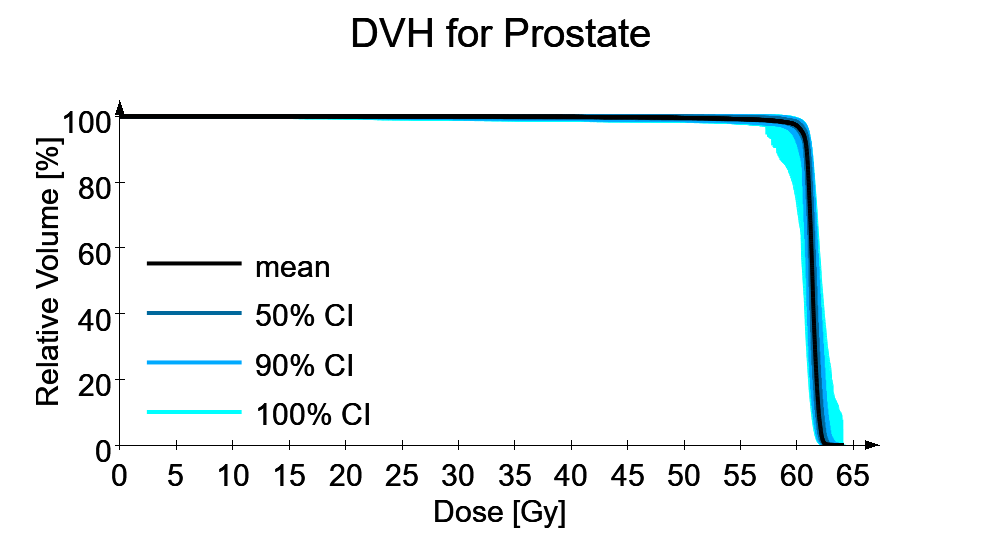} &
    \includegraphics[height=\imgheight,width=\imgwidth,trim=10mm 5mm 40mm 30mm, clip]{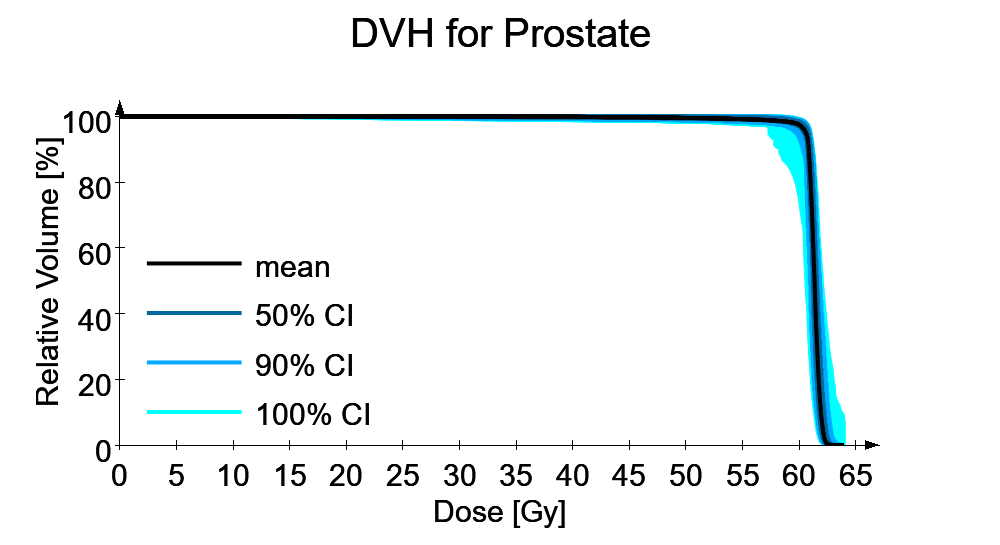} 
      \\
      \rotatebox{90}{\parbox{\imgheight}{\centering DVH Rectum}} &
    \includegraphics[height=\imgheight,width=\imgwidth,trim=10mm 5mm 40mm 30mm, clip]{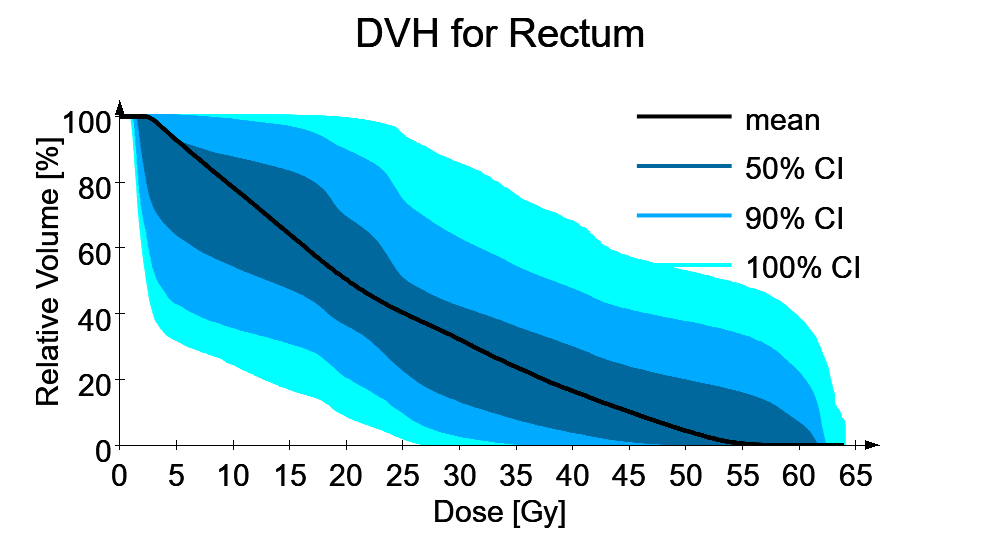} &
    \includegraphics[height=\imgheight,width=\imgwidth,trim=10mm 5mm 40mm 30mm, clip]{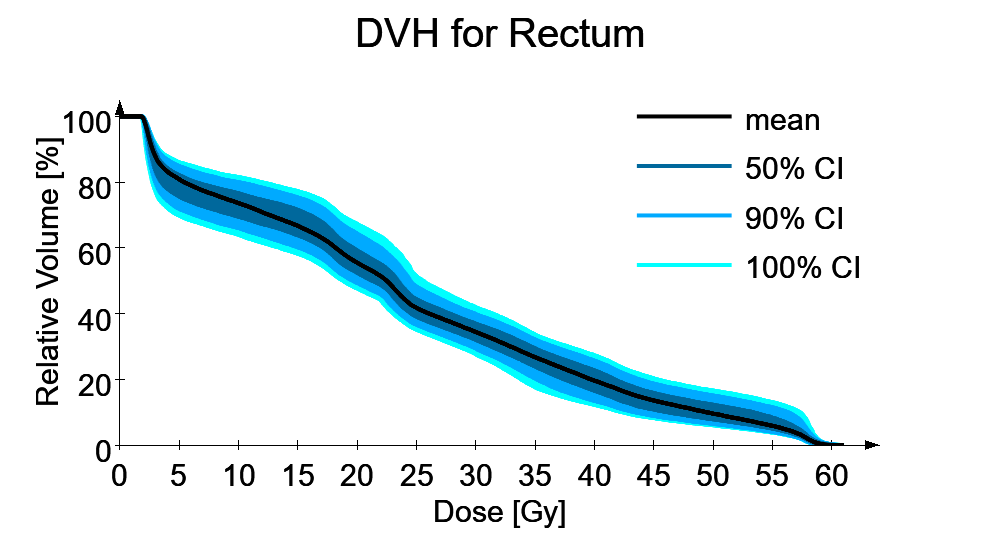} &
    \includegraphics[height=\imgheight,width=\imgwidth,trim=10mm 5mm 40mm 30mm, clip]{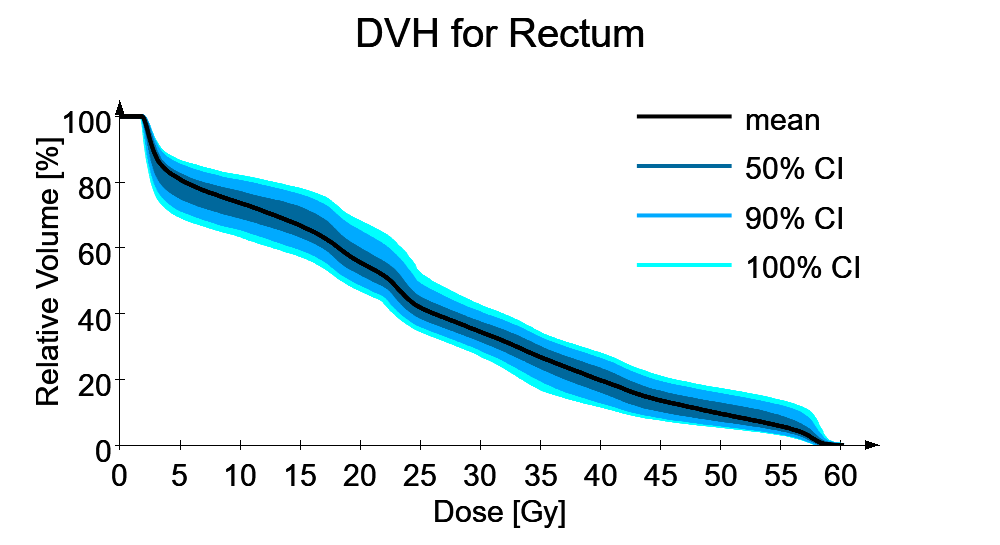} 
  \end{tabular}
  \caption{Comparison of uncertainty strategies with a max radius of 20mm, as shown in Figure~\ref{fig:compare_strategies_10mm} for the case of a max radius of 10mm. Again, the first column: strategy 1 with constant uncertainty radius; second + third column: strategy 2 with distance-based uncertainty radius without and with in/out consideration. In all examples the kernel is uniform and the max radius is 20mm.}  
  \label{fig:compare_strategies_20mm}
\end{figure}

The results are shown in Figures~\ref{fig:compare_strategies_10mm} and \ref{fig:compare_strategies_20mm}. The three columns correspond to Strategy 1, Strategy 2, and Strategy 2 + in/out. In all cases, a uniform kernel is used; the maximum radius is 10mm in Figure~\ref{fig:compare_strategies_10mm} and 20mm in Figure~\ref{fig:compare_strategies_20mm}. The rows show the certainty map, the probability of receiving at least 60Gy, the upper 95\% percentile map, and the DVHs for bladder, prostate, and rectum, including mean-dose DVHs and confidence-derived DVH envelopes. For the DVH, we report three increasing interval levels: $\mathrm{CI}_{50\%}:=[\operatorname{DVH}_{75\%}^{\min},\operatorname{DVH}_{75\%}^{\max}]$,
$\mathrm{CI}_{90\%}:=[\operatorname{DVH}_{95\%}^{\min},\operatorname{DVH}_{95\%}^{\max}]$,
and $\mathrm{CI}_{100\%}:=[\operatorname{DVH}_{100\%}^{\min},\operatorname{DVH}_{100\%}^{\max}]$.
Each interval is the envelope between the corresponding lower and upper confidence-bound DVH curves. Equivalently, these envelopes are induced by voxel-wise percentile bounds: 25\%/75\% for $\mathrm{CI}_{50\%}$, 5\%/95\% for $\mathrm{CI}_{90\%}$, and 0\%/100\% (support bounds) for $\mathrm{CI}_{100\%}$.

Note that, the certainty maps for the additional in/out strategy (top row, third column in Figures~\ref{fig:compare_strategies_10mm} and \ref{fig:compare_strategies_20mm}), have values that locally exceed the prescribed maximum radius because of large outside distances. In the present example, some voxels are approximately 26\,mm away from the target structure, which leads to certainty radii of up to approximately 54\,mm under the doubling rule. For improved visualization, we therefore use an additional bluish colormap for voxels that exceed the maximum radius due to the in/out strategy. In the displayed slice, the actual values are lower and reach only up to approximately 20\,mm. The largest deviations in this example occur in a region adjacent to the bladder. For better visibility, we additionally outline the outside-voxel region with a black contour.

Several observations emerge from these figures. First, over large parts of the domain, the certainty maps of all three strategies are nearly identical because the uncertainty radius reaches the prescribed maximum value of 10mm or 20mm. The relevant differences therefore arise mainly near the structure boundaries. Second, these boundary regions coincide with locations of high dose gradients. At such locations, changes in the registration have the strongest effect on the propagated dose, whereas even larger registration errors have only limited dosimetric impact in regions with flat dose gradients. This explains why local differences in the certainty map are reflected primarily in the probability and percentile maps near the organ and target interfaces.

At the DVH level, Strategy 1 yields clearly wider envelopes than the two distance-based variants, indicating that the globally constant model is more conservative in this example. By contrast, the distance-based approaches with and without in/out produce very similar DVHs and, more generally, very similar results. In this case, the main improvement therefore comes from the spatially varying certainty map itself, while the additional in/out restriction has only a minor effect.

This experiment is intended to illustrate the effect of the modeling choices rather than to identify a universally best strategy. The preferred setting depends on the clinical application, the available prior information, and the desired balance between simplicity and anatomical specificity.

%%
%%
%%  ------------------------------------------------------------------------------------
%%
%%

%%
%%
%%  ------------------------------------------------------------------------------------
%%
%%

%%
%%
%%  ------------------------------------------------------------------------------------
%%
%%

\begin{figure}[!htpb]
 \centering
  \def\imgheight{0.13\textheight}
   \scriptsize
  \begin{tabular}{ccc} 
      & Probality mapped dose $\geq 60$Gy & Confidence bound $\mathrm{Upper}(x)$. \\
      & $\Pr[\dose(Y(x)) \geq 60\,\mathrm{Gy}]$  & $\Pr[\dose(Y(x)) < \mathrm{Upper}(x)] \geq 0.95$ \\
      \\
      \rotatebox{90}{\parbox{\imgheight}{\centering Uniform}} &
    \includegraphics[height=\imgheight]{images/minRad_1-maxRad_20-slope_1_noInOut_prob_dose_ge_60gy_Uniform.png} &
    \includegraphics[height=\imgheight]{images/minRad_1-maxRad_20-slope_1_noInOut_prob_dose_lt_x_is_95gy_Uniform.png} 
      \\
      \rotatebox{90}{\parbox{\imgheight}{\centering Linear B-spline}} &
    \includegraphics[height=\imgheight]{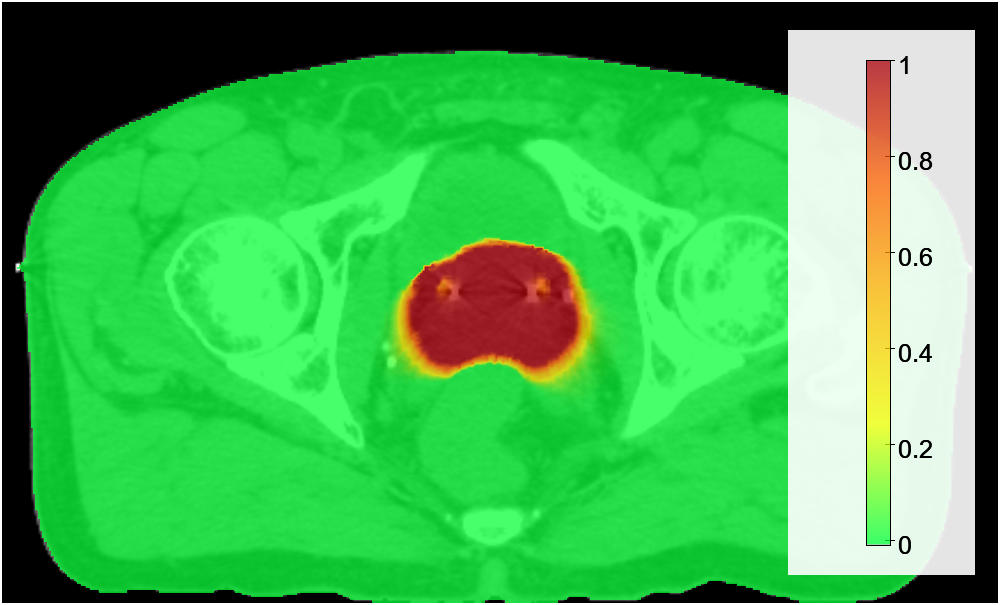} &
    \includegraphics[height=\imgheight]{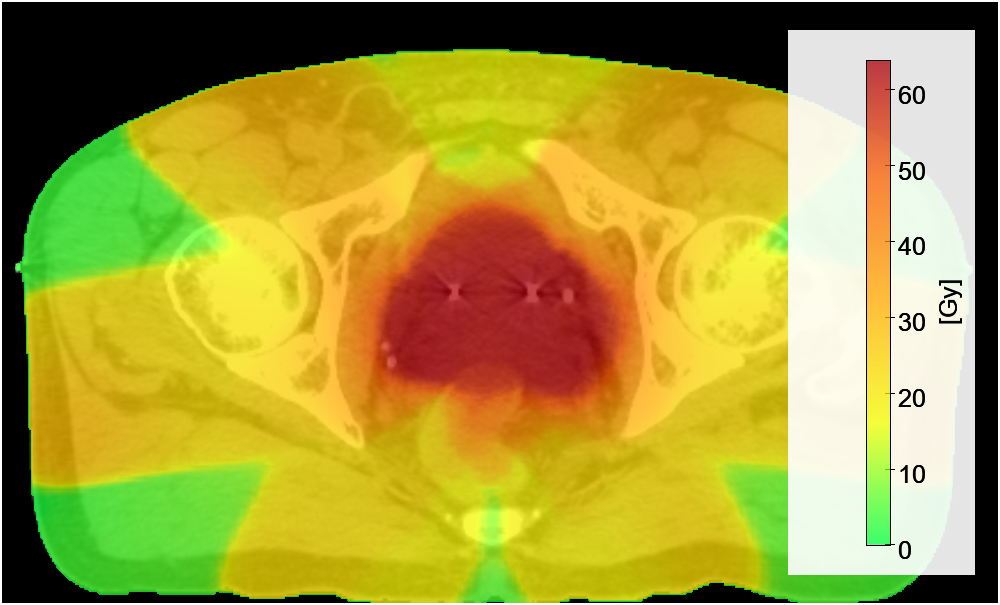} 
      \\
      \rotatebox{90}{\parbox{\imgheight}{\centering Quad B-spline}} &
    \includegraphics[height=\imgheight]{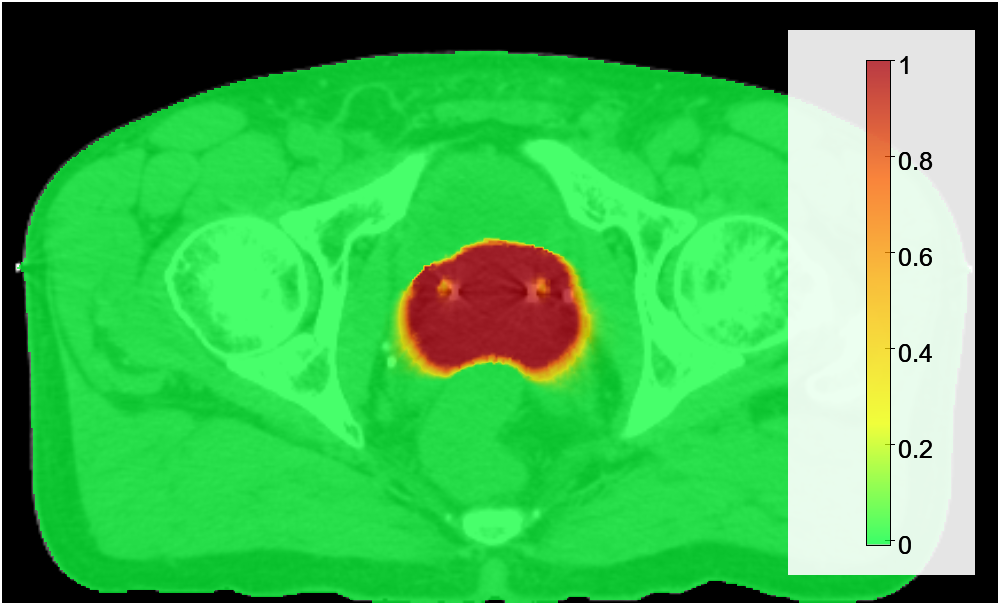} &
    \includegraphics[height=\imgheight]{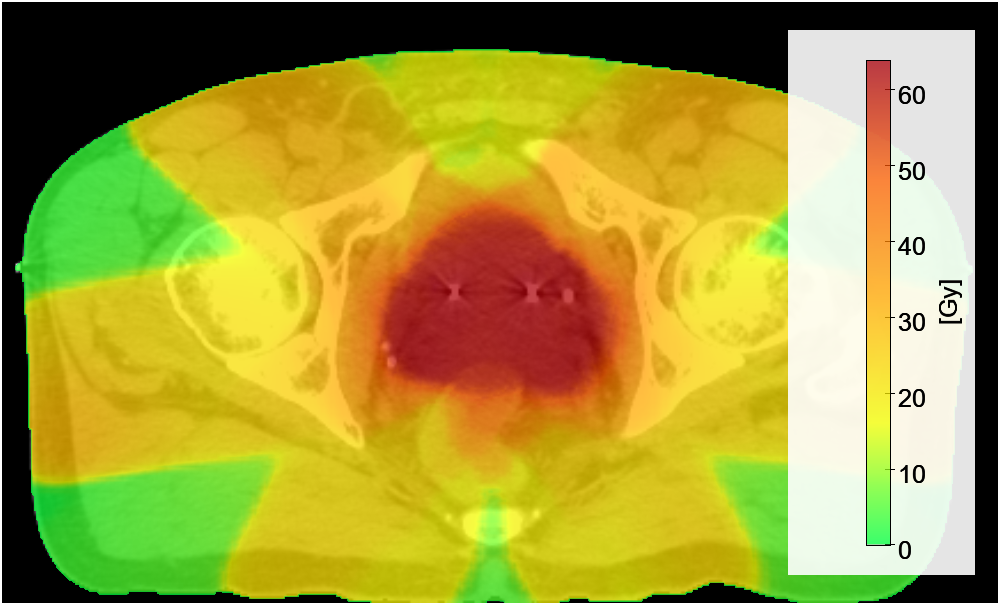} 
      \\
      \rotatebox{90}{\parbox{\imgheight}{\centering Cubic B-spline}} &
    \includegraphics[height=\imgheight]{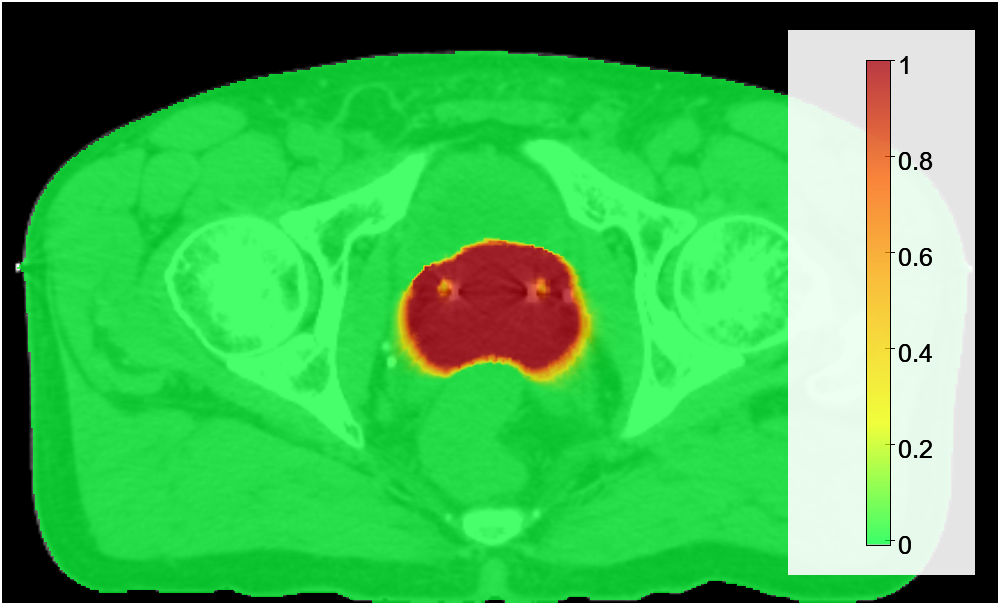} &
    \includegraphics[height=\imgheight]{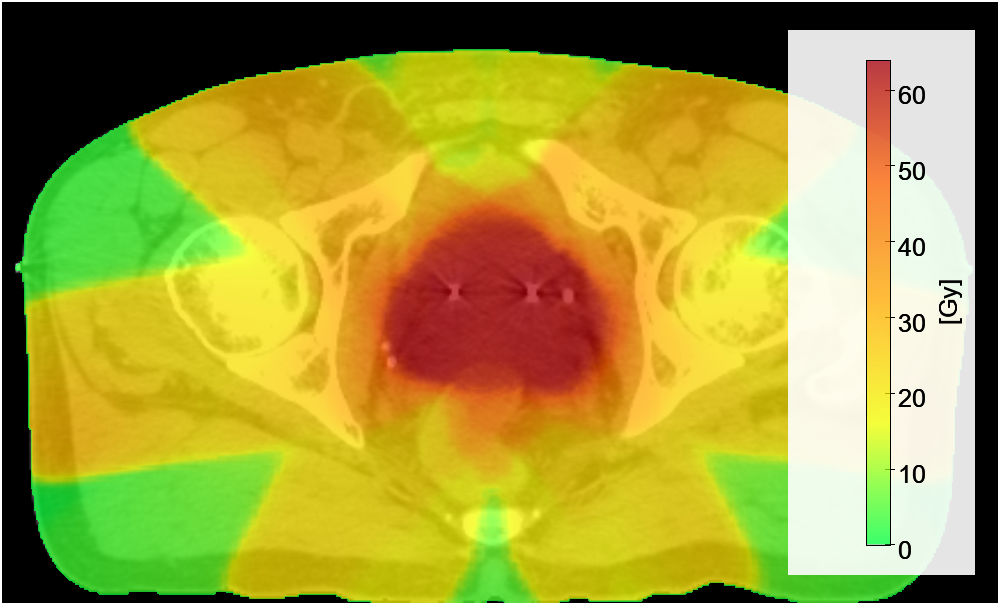} 
      \\
      \rotatebox{90}{\parbox{\imgheight}{\centering Gauss $\sigma=\frac13$}} &
    \includegraphics[height=\imgheight]{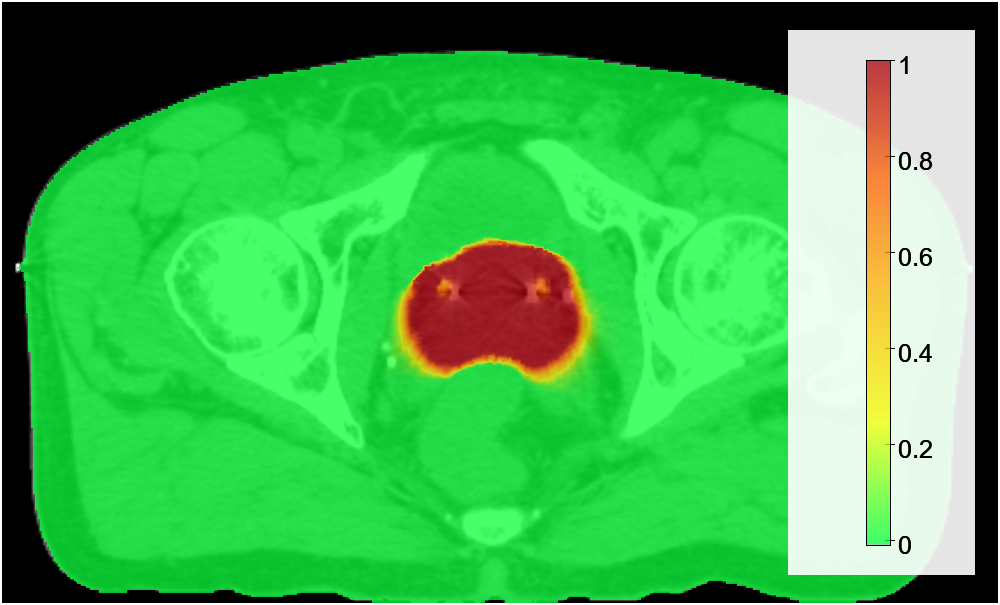} &
    \includegraphics[height=\imgheight]{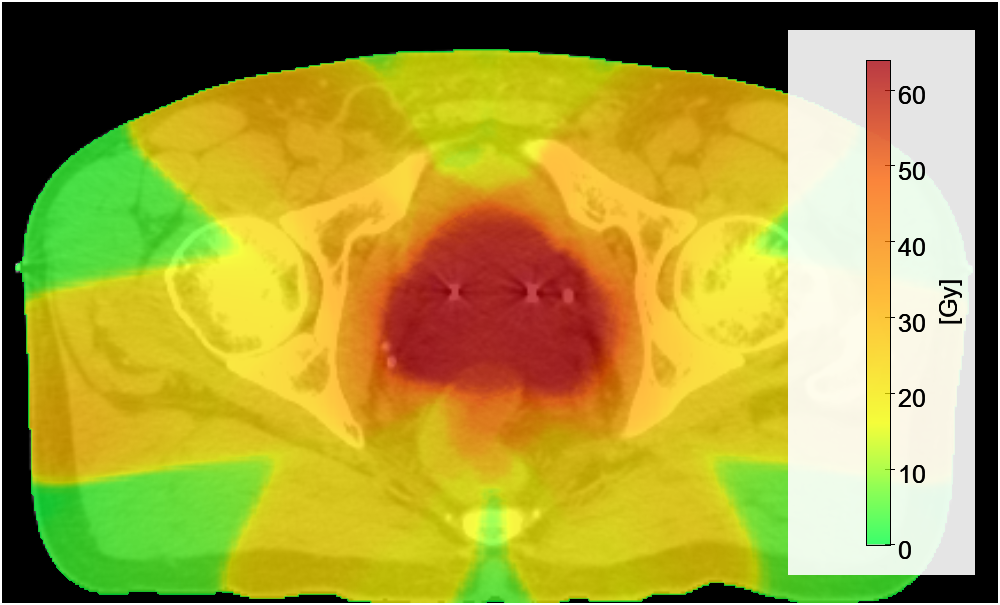} 
      \\
      \rotatebox{90}{\parbox{\imgheight}{\centering Gauss $\sigma=\frac14$}} &
    \includegraphics[height=\imgheight]{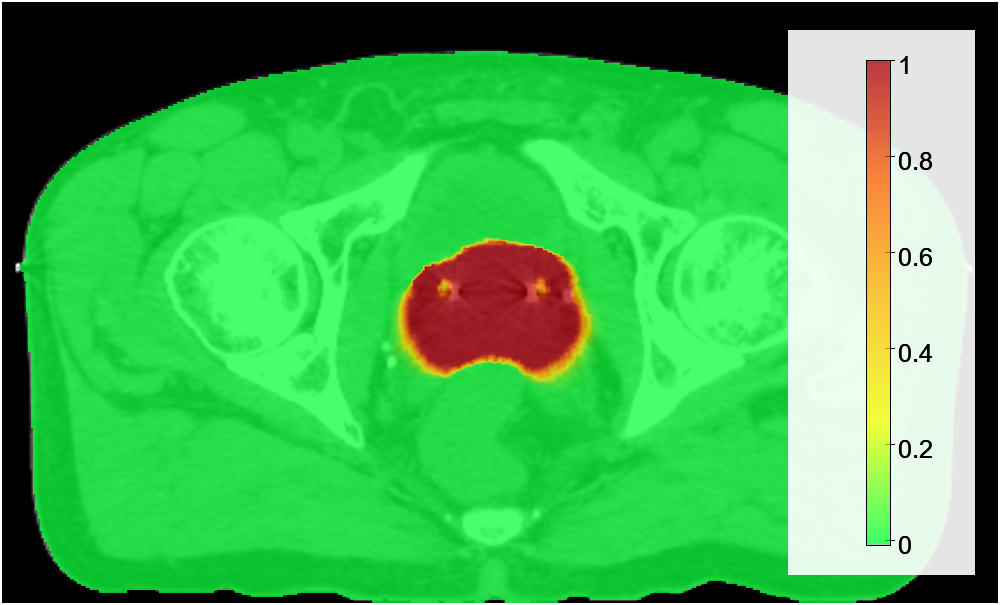} &
    \includegraphics[height=\imgheight]{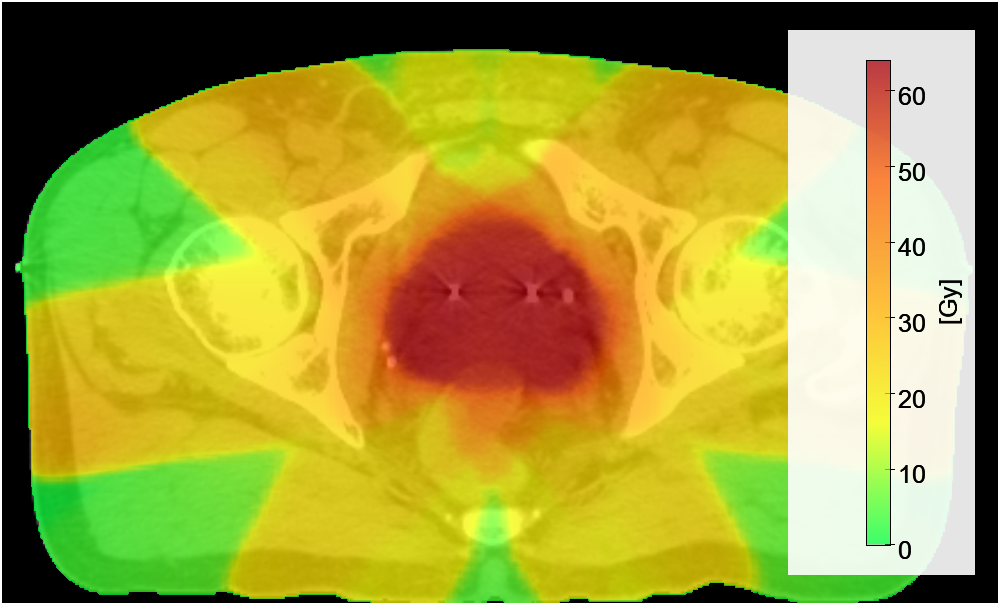} 
  \end{tabular}
  \caption{Comparison of different kernels for probability and confidence maps. In all examples, a distance-based uncertainty model is used with a minimum uncertainty of 1 mm and a maximum uncertainty radius of 20 mm. The uncertainty increases linearly with the distance from the structure boundaries, using a slope of 1.}
  \label{fig:compare_kernels_prob_conf}
\end{figure}

%%
%%
%%  ------------------------------------------------------------------------------------
%%
%%

\begin{figure}[!htpb]
  \setlength{\tabcolsep}{3pt}
  \renewcommand{\arraystretch}{1.0}
  \begin{tabular}{cccc} 
      & Bladder & Prostate & Rectum \\
      \scriptsize  \rotatebox{90}{\hspace{5mm}Uniform} &
    \includegraphics[width=0.3\textwidth, trim=12mm 5mm 40mm 20mm, clip]{images/minRad_1-maxRad_20-slope_1_noInOut_DVH_Bladder_Uniform.png} &
    \includegraphics[width=0.3\textwidth, trim=12mm 5mm 40mm 20mm, clip]{images/minRad_1-maxRad_20-slope_1_noInOut_DVH_Prostate_Uniform.png} &
    \includegraphics[width=0.3\textwidth, trim=12mm 5mm 40mm 20mm, clip]{images/minRad_1-maxRad_20-slope_1_noInOut_DVH_Rectum_Uniform.png} 
      \\
      \scriptsize \rotatebox{90}{\hspace{5mm}Linear B-spline} &
    \includegraphics[width=0.3\textwidth, trim=12mm 5mm 40mm 20mm, clip]{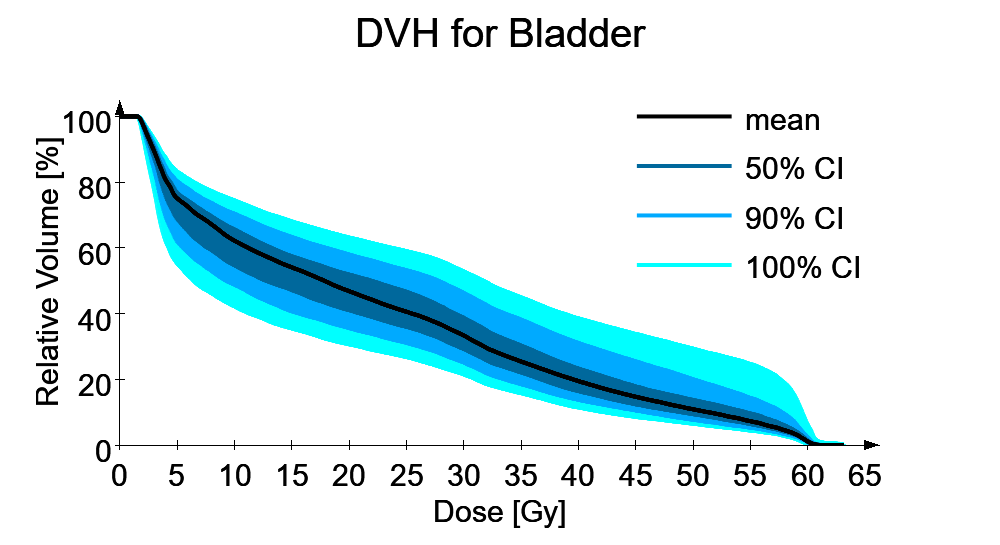}&
    \includegraphics[width=0.3\textwidth, trim=12mm 5mm 40mm 20mm, clip]{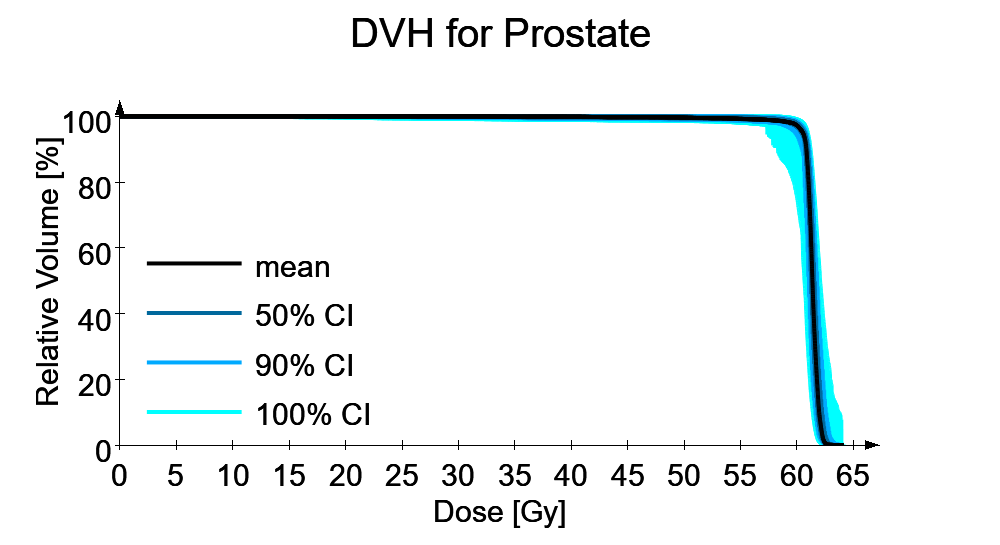} &
    \includegraphics[width=0.3\textwidth, trim=12mm 5mm 40mm 20mm, clip]{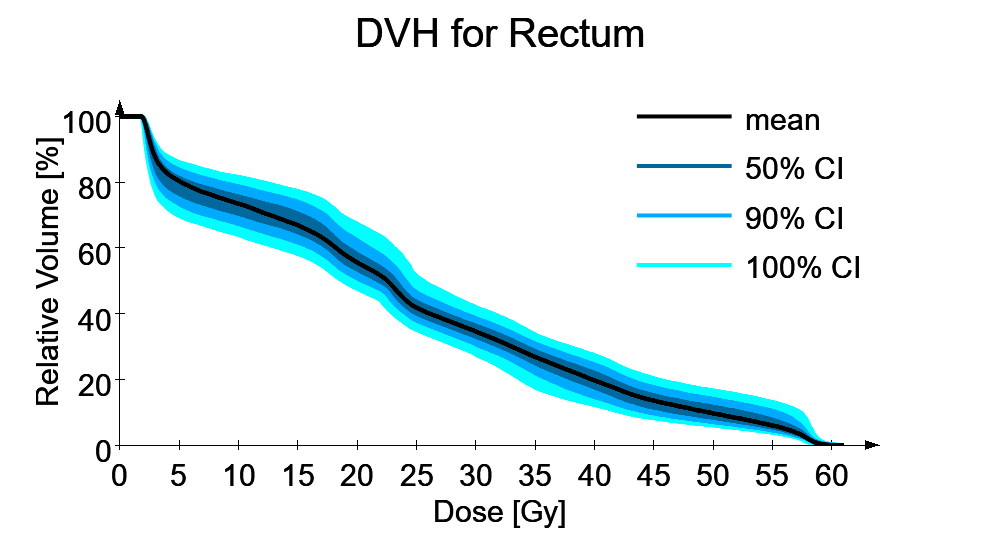}
      \\
      \scriptsize \rotatebox{90}{\hspace{5mm}Quad B-spline} &
    \includegraphics[width=0.3\textwidth, trim=12mm 5mm 40mm 20mm, clip]{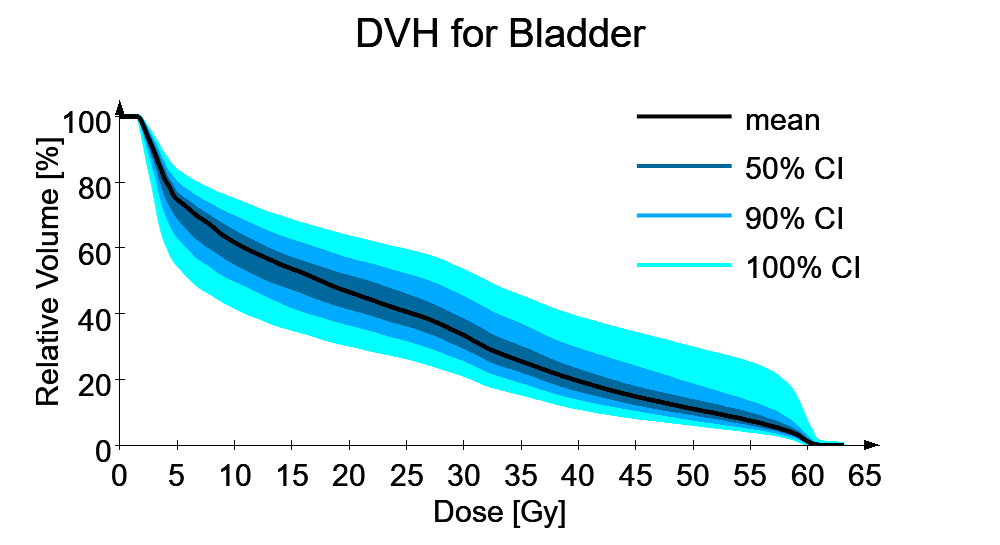}&
    \includegraphics[width=0.3\textwidth, trim=12mm 5mm 40mm 20mm, clip]{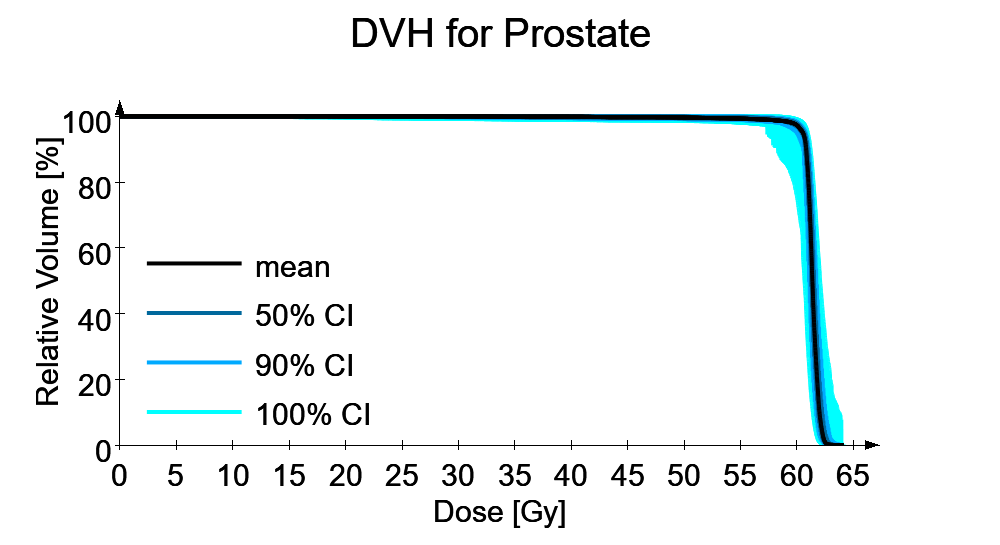}&
    \includegraphics[width=0.3\textwidth, trim=12mm 5mm 40mm 20mm, clip]{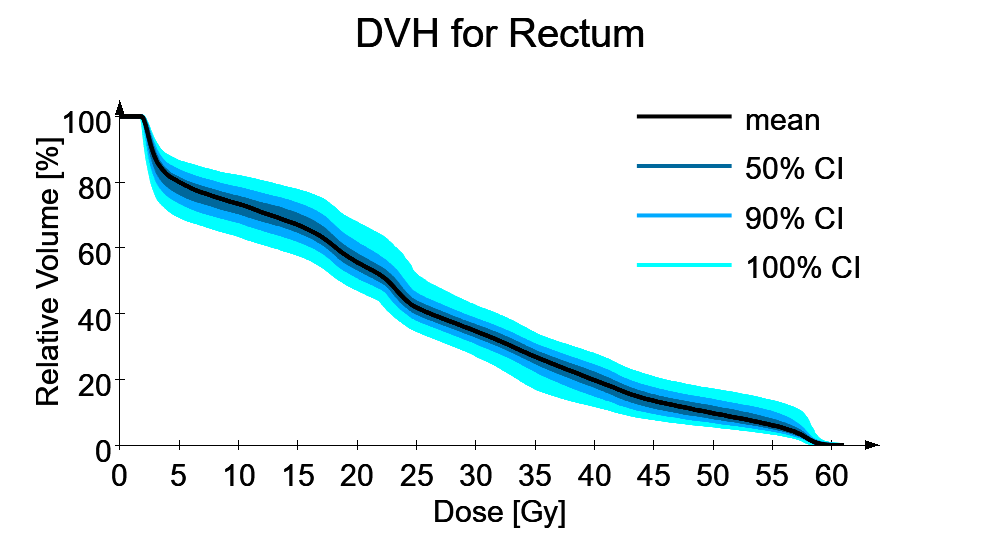}
      \\
      \scriptsize \rotatebox{90}{\hspace{5mm}Cubic B-spline} &
    \includegraphics[width=0.3\textwidth, trim=12mm 5mm 40mm 20mm, clip]{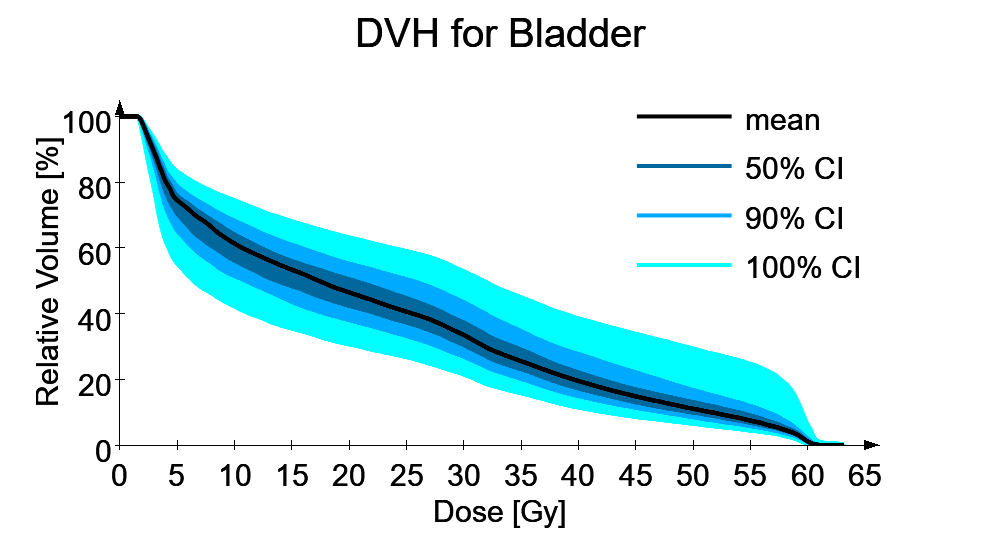}&
    \includegraphics[width=0.3\textwidth, trim=12mm 5mm 40mm 20mm, clip]{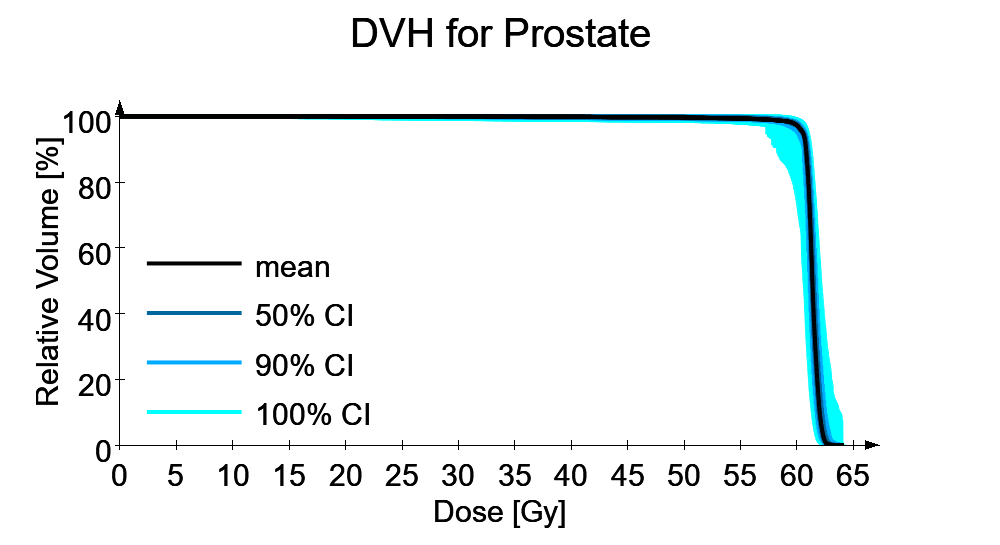} &
    \includegraphics[width=0.3\textwidth, trim=12mm 5mm 40mm 20mm, clip]{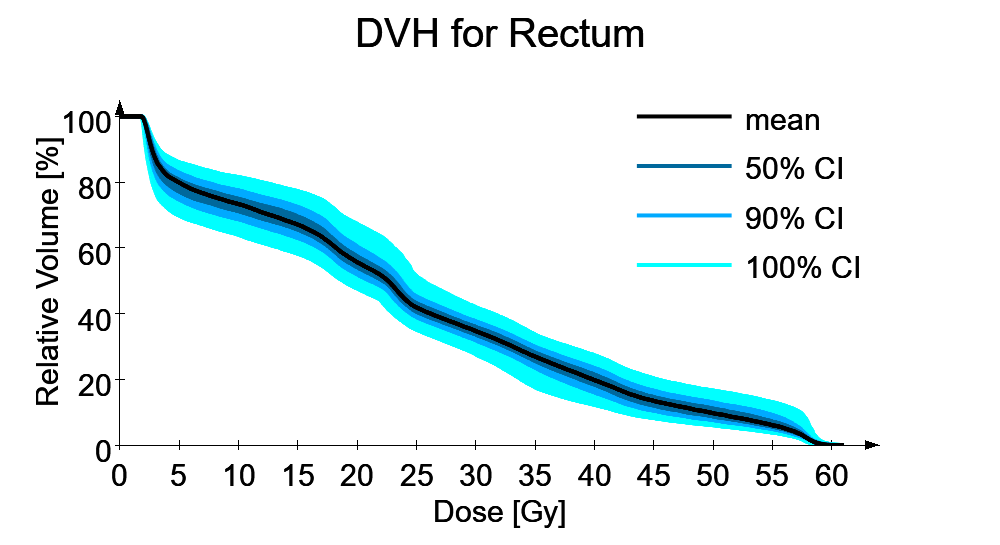} 
      \\
      \scriptsize \rotatebox{90}{\hspace{5mm}Gauss $\sigma=\frac13$} &
    \includegraphics[width=0.3\textwidth, trim=12mm 5mm 40mm 20mm, clip]{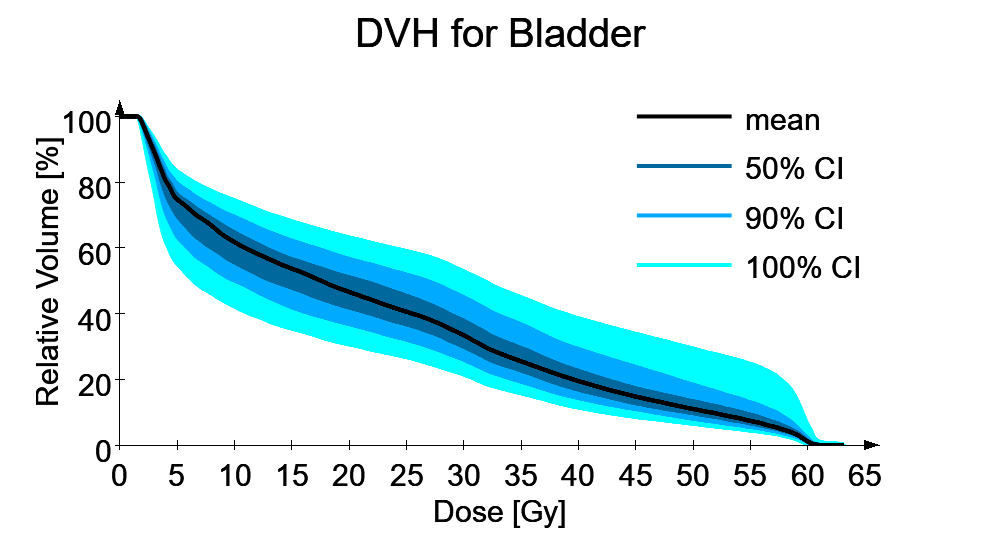} &
    \includegraphics[width=0.3\textwidth, trim=12mm 5mm 40mm 20mm, clip]{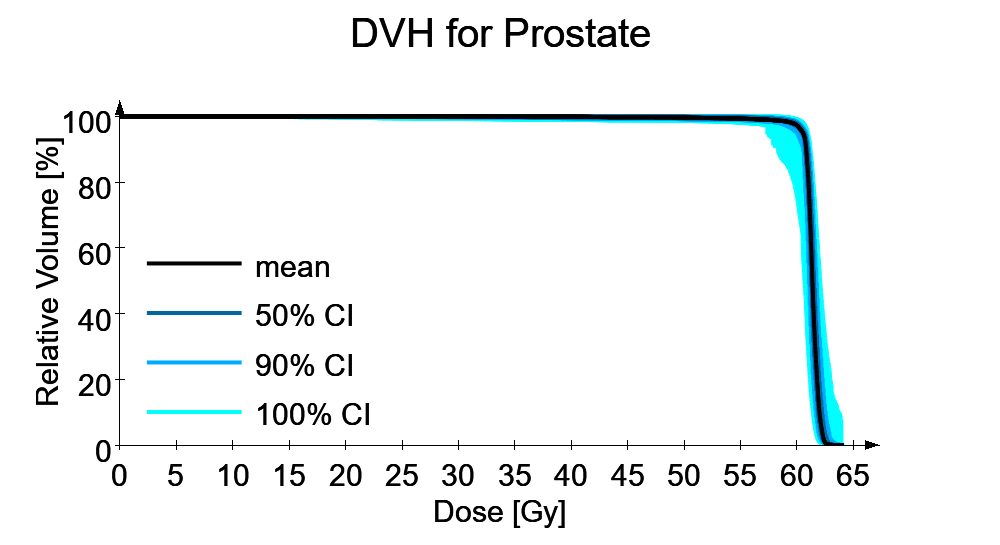} &
    \includegraphics[width=0.3\textwidth, trim=12mm 5mm 40mm 20mm, clip]{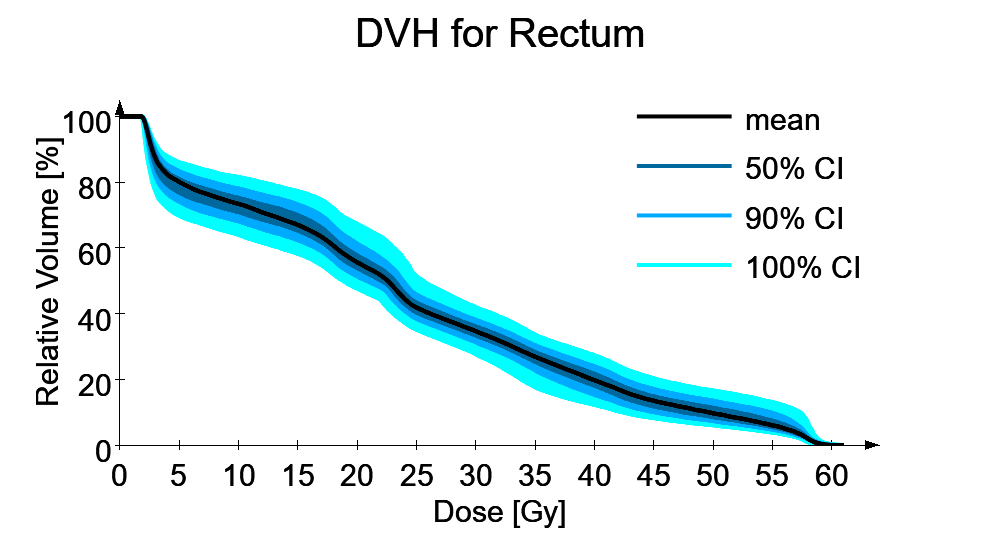} 
      \\
      \scriptsize \rotatebox{90}{\hspace{5mm}Gauss $\sigma=\frac14$} &
    \includegraphics[width=0.3\textwidth, trim=12mm 5mm 40mm 20mm, clip]{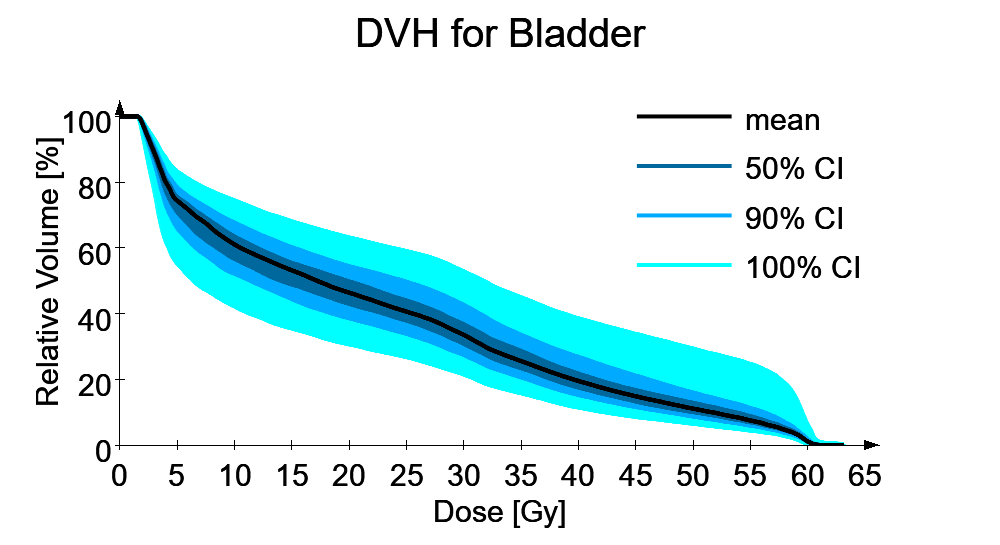} &
    \includegraphics[width=0.3\textwidth, trim=12mm 5mm 40mm 20mm, clip]{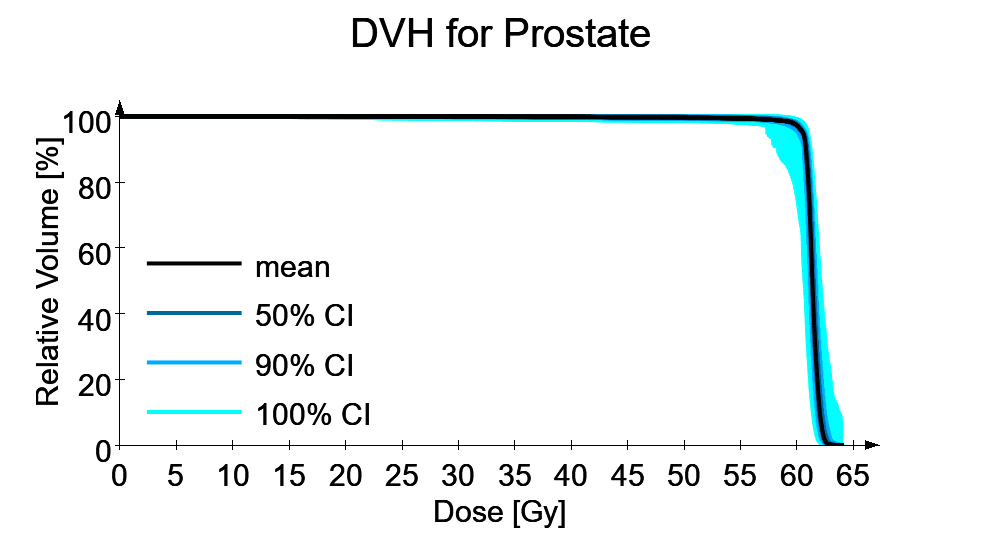} &
    \includegraphics[width=0.3\textwidth, trim=12mm 5mm 40mm 20mm, clip]{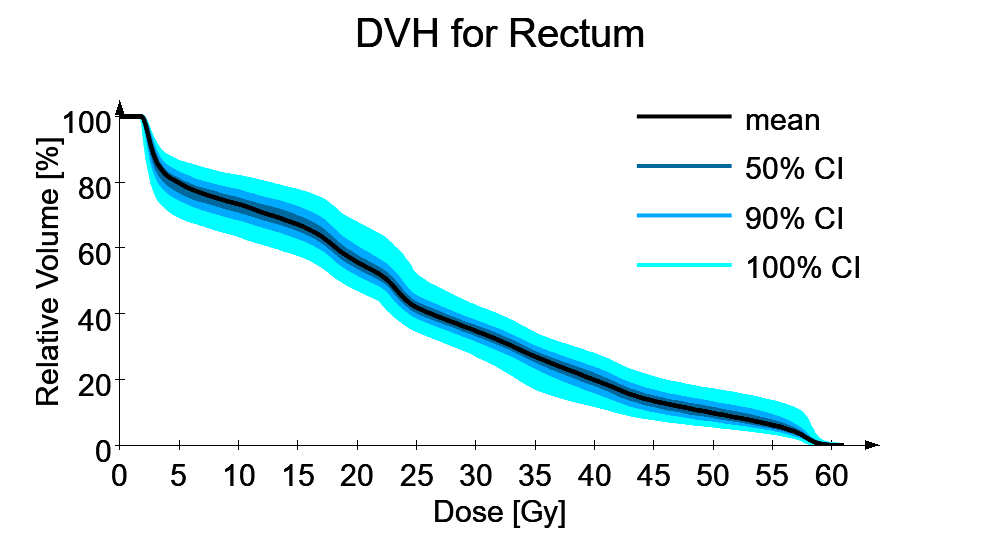} 
  \end{tabular}
  \caption{Comparison of the influence of different probability kernels on DVH and DVH bounds. In all examples, a distance-based uncertainty model is used with a minimum uncertainty of 1 mm and a maximum uncertainty radius of 20 mm. The uncertainty increases linearly with the distance from the structure boundaries, using a slope of 1.}
  \label{fig:compare_kernels_dvhs}
\end{figure}

\subsection{The choice of the probabilty kernel}
In the second experiment, we will compare different kernels for modeling the probability distribution of the mapped dose. We will compare the kernels described in figure \ref{fig:Probability-kernels} (uniform, linear, quadratic and cubic b-spline, Gausian), using two different standard deviations for Gaussian kernels,  $\sigma=\frac13$ and  $\sigma=\frac14$.

The results of this experiment are shown in Figures~\ref{fig:compare_kernels_prob_conf} and \ref{fig:compare_kernels_dvhs}. In Figure~\ref{fig:compare_kernels_prob_conf}, the first column shows the probability that the mapped dose is larger than 60 Gy, $\Pr[\dose(Y(x))\geq 60\,\mathrm{Gy}]$, while the second column shows the upper 95\% percentile map $\mathrm{Upper}(x)$, that is, the mapped dose is below $\mathrm{Upper}(x)$ with 95\% probability, i.e. $\Pr[\dose(Y(x))<\mathrm{Upper}(x)]\geq 0.95$. The rows correspond to the different kernels. Overall, the kernel choice has little visible effect on the probability and confidence maps. The only noticeable difference is observed for the uniform kernel in regions with steep dose gradients; the more localized kernels produce almost identical results. The same pattern is reflected in the DVHs shown in Figure~\ref{fig:compare_kernels_dvhs}. These curves are again very similar and differ mainly in the width of the confidence intervals, given by the DVHs of the lower and upper confidence bounds $D_\alpha^{\min}$ and $D_\alpha^{\max}$, cf. \eqref{eq:confidence-bounds:min} and \eqref{eq:confidence-bounds:max}, which become narrower for more localized kernels.
Note that the 100\% confidence intervals along each organ are the same for all kernels by design, since, as shown in \eqref{eq:min_max_dose}, these bounds depend only on the support of the probability distribution. The support is identical for all kernels here, as they are defined and scaled using the same certainty map.

\section{Discussion}
\label{sec:discussion}
In this work, we presented a simple, transparent, and practically applicable model for quantifying uncertainties caused by deformable image registration for dose propagation in radiotherapy. In contrast to complex and often hard-to-interpret approaches, our model focuses on clinical applicability and interpretability. The experiments demonstrate that even with straightforward uncertainty strategies and just a small number of parameters, robust and easy to understandable uncertainty margins for dose distributions and DVHs can be computed.

A key finding is that the choice of uncertainty strategy (e.g., globally constant vs. structure- or distance-based) has a much greater impact on the resulting uncertainty margins than the choice of kernel. While the kernel affects the probability distribution within the uncertainty radius, the uncertainty strategy primarily determines the size and spatial distribution of the uncertainty regions and thus the clinical relevance of the resulting dose estimates.

The first experiment also helps to explain why these differences arise. Over large parts of the anatomy, all three certainty maps are nearly identical because the uncertainty radius saturates at the prescribed maximum value. The practically relevant differences are therefore concentrated near the structure boundaries. This is particularly important because these boundary regions often coincide with high dose gradients. In such regions, even small changes in the mapped location can lead to substantial changes in propagated dose, whereas larger registration deviations have much less dosimetric effect in regions with flat dose gradients. Accordingly, local differences in the certainty-map design are amplified mainly where they matter most for dose evaluation.

This observation is also reflected at the DVH level. In our example, the globally constant strategy produces substantially wider DVH envelopes than the distance-based approaches, indicating a more conservative but also less specific uncertainty model. By contrast, the distance-based strategies yield tighter and more structure-adapted uncertainty bounds because they avoid assigning equally large uncertainty to regions where the dosimetric impact of registration error is limited.

The additional in/out strategy should be interpreted in this context as a structure-guided refinement rather than a universally strong effect. Conceptually, it is attractive because it removes anatomically implausible mappings and enforces consistency with matched source and target structures. In the present example, however, its effect on the DVHs is small and the results with and without in/out are very similar. This suggests that, for this dataset, the dominant gain already comes from using a spatially varying distance-based certainty map, whereas the additional structural restriction only yields a modest refinement. At the same time, this does not diminish the relevance of the in/out idea: its benefit can be expected to be larger in cases with stronger inter-structure ambiguity, more severe local misalignment, or more pronounced anatomical overlap.

Limitations of our approach include the assumption of independent errors and the lack of modeling of spatial correlations. This leads to conservative, potentially overestimated uncertainty margins. Consequently, the present work should be understood as a methodological and illustrative study rather than as a clinical validation. The model can, however, serve as a starting point for more advanced developments and for the separate commissioning and validation steps required before deployment in practice \cite[]{BosmaEtAl2024, KipritidisEtAl2025}.

\section{Conclusions}
\label{sec:conclusions}
We have introduced an efficient and transparent model to account for uncertainties in deformable image registration for dose propagation. The model enables the calculation of uncertainty margins and confidence intervals for dose distributions and DVHs with low computational effort and high interpretability.

We have presented a transparent and computationally efficient framework for incorporating deformable image registration uncertainty into dose propagation. It enables the computation of uncertainty margins and confidence intervals for dose distributions and DVHs with low computational overhead and clear interpretability. As the DVH is determined by the underlying dose map, pointwise statistics can be used to create corresponding maps and DVH curves. This provides a straightforward way of converting voxel-wise uncertainty into DVH margins and envelopes. The experiments show that even simple assumptions lead to practically relevant and comprehensible results.

In particular, the study indicates that the design of the certainty map is more influential than the precise choice of probability kernel. In our example, the decisive differences arise near structure boundaries, where high dose gradients make the propagated dose especially sensitive to registration uncertainty. This explains why the globally constant strategy produces markedly wider DVH envelopes, whereas the distance-based strategies provide tighter and more anatomically adapted uncertainty bounds. The additional in/out strategy offers a principled way to suppress anatomically implausible mappings; in the present case, however, its effect is comparatively small, so that most of the improvement is already achieved by the distance-based certainty map itself.

For clinical application, we therefore recommend that uncertainty strategies be chosen deliberately and transparently, with particular attention to boundary regions and high-gradient dose areas. Future work should address the integration of spatial correlations, broader validation on larger multicenter datasets, and a more systematic analysis of clinical scenarios in which structure-guided in/out constraints provide the greatest added value.

\bibliography{bibliography}
\bibliographystyle{abbrvnat}
\end{document}